\documentclass[journal]{IEEEtran}
\ifCLASSINFOpdf
\else
\fi
\usepackage{caption} 

\renewcommand{\thetable}{\arabic{table}}

\usepackage{amsmath,amssymb,amsfonts}

\usepackage{soul,color}
\usepackage{booktabs} 

\usepackage{graphicx}
\usepackage{textcomp}
\usepackage{float}
\usepackage{subfig}
\usepackage{amssymb}
\usepackage{enumitem}
\usepackage{pifont}
\usepackage[linesnumbered,ruled,vlined]{algorithm2e}

\usepackage{pifont}

\usepackage[utf8]{inputenc}
\usepackage[T1]{fontenc}
\usepackage{textcomp}
\usepackage{multirow}

\usepackage{pbox}
 
\usepackage{makecell}
\usepackage{tabularray}

\SetCommentSty{mycommfont}
\usepackage{bbding}
\SetKwInput{KwInput}{Input}                
\SetKwInput{KwOutput}{Output}              
\SetKwFunction{NewFunction}{NewFunction}   
\usepackage{makecell}
\begin{document}
%
\title{ A Novel Double Pruning method for Imbalanced Data using Information Entropy and Roulette Wheel Selection for Breast Cancer Diagnosis}

\author{
    \scriptsize
    \begin{minipage}{0.20\textwidth}
        \centering
        \IEEEauthorblockN{Bacha Soufiane $^{1}$} \\
        \IEEEauthorblockA{School of Computer and \\Communication Engineering,\\
        University of Science and Technology Beijing\\
        Beijing 100083, China. \\ 
        Email: d202361018@xs.ustb.edu.cn}
    \end{minipage}
    \hfill
    \begin{minipage}{0.20\textwidth}
        \centering
        \IEEEauthorblockN{Huansheng Ning $^{2}$} \\
        \IEEEauthorblockA{School of Computer and \\Communication Engineering,\\
        University of Science and Technology Beijing\\
        Beijing 100083, China. \\ 
        Email: ninghuansheng@ustb.edu.cn}
    \end{minipage}
    \hfill
    \begin{minipage}{0.19\textwidth}
        \centering
        \IEEEauthorblockN{Belarbi Mostefa $^{3}$} \\
        \IEEEauthorblockA{Laboratory of Informatics and Mathematics (LIM) of\\ University of Tiaret, \\University of Tiaret,\\Faculty of Mathematics and Informatics,\\ Department of Computer Science, BP 78 Za\^{a}roura, Tiaret 14000, Algeria \\ 
        Email: mbelarbi@univ-tiaret.dz}
    \end{minipage}
    \hfill
    \begin{minipage}{0.19\textwidth}
        \centering
        \IEEEauthorblockN{Doreen Sebastian Sarwatt $^{4}$} \\
        \IEEEauthorblockA{School of Computer and \\Communication Engineering,\\
        University of Science and Technology Beijing\\
        Beijing 100083, China. \\ 
        Email: dollysebastian@hotmail.com\\ d202161035@xs.ustb.edu.cn}
    \end{minipage}
    \hfill
    \begin{minipage}{0.20\textwidth}
        \centering
        \IEEEauthorblockN{Sahraoui Dhelim $^{5}$} \\
        \IEEEauthorblockA{School of Computing, \\ Dublin City University, Ireland \\ 
        Email: sahraoui.dhelim@dcu.ie}
    \end{minipage}
    
}


%


\maketitle

\begin{abstract}
Accurate illness diagnosis is vital for effective treatment and patient safety. Machine learning models are widely used for cancer diagnosis based on historical medical data. However, data imbalance remains a major challenge, leading to hindering classifier performance and reliability. The SMOTEBoost method addresses this issue by generating synthetic data to balance the dataset, but it may overlook crucial overlapping regions near the decision boundary and can produce noisy samples. This paper proposes RE-SMOTEBoost, an enhanced version of SMOTEBoost, designed to overcome these limitations. Firstly, RE-SMOTEBoost focuses on generating synthetic samples in overlapping regions to better capture the decision boundary using roulette wheel selection. Secondly, it incorporates a filtering mechanism based on information entropy to reduce noise, and borderline cases and improve the quality of generated data. Thirdly, we introduce a double regularization penalty to control the synthetic samples' proximity to the decision boundary and avoid class overlap. These enhancements enable higher-quality oversampling of the minority class, resulting in a more balanced and effective training dataset. The proposed method outperforms existing state-of-the-art techniques when evaluated on imbalanced datasets.   Compared to the top-performing sampling algorithms, RE-SMOTEBoost demonstrates a notable improvement of 3.22\% in accuracy and a variance reduction of 88.8\%. These results indicate that the proposed model offers a solid solution for medical settings, effectively overcoming data scarcity and severe imbalance caused by limited samples, data collection difficulties, and privacy constraints.

\end{abstract}


%
\IEEEpeerreviewmaketitle

\section{Introduction}
Nowadays, cancer outbreaks occur at a high rate, posing a significant threat to humanity's continuity. This disease has shown a rising incidence in various countries; for example, in the USA alone, the incidence rate has reached 300,000 new cancer cases per year, which is double the rate of the last 30 years. Although the majority of women survive, more than 40,000 women still die annually  \cite{introduction1}. This outbreak is associated with several factors, such as age, family history, and others \cite{introduction2}.

Early prevention and diagnosis are essential mechanisms for reducing incidence and mortality rates. These approaches include annual mammography, standard clinical practice, and more \cite{introduction3}.Moreover, with the increase in data dimensionality and the expansion of media and clinical data storage, many recent advancements have emerged in early healthcare prevention, particularly in the field of bioinformatics. Bioinformatics integrates various disciplines into the healthcare process, with machine learning tools playing a dominant role. These tools utilize various data-driven prediction and classification models, such as decision trees (DTs), random forests (RF), and support vector machines (SVMs).

In the machine learning domain, cancer prediction is typically formulated as a classification task to categorize tumors as either benign or malignant. The classification performance depends on various factors, including input features, parameter settings, and data quality such as imbalanced data, noise, and other challenges \cite{introduction4}\cite{introduction5}. In particular, cancer diagnosis is not only about achieving high accuracy but also about the robustness and reliability of the model, which can be highly affected by an imbalanced class ratio. Imbalanced data refers to situations where one class has more data than another, and this is common in several real-world tasks, such as medical diagnosis \cite{introduction6}. In an imbalanced data scenario, the classes with fewer sample data are called the minority class, which represents malignant tumors, while the class with more sample data is called the majority class, which represents benign tumors. The minority class is more important but contributes less to the final prediction. As a result, the trained model tends to favor the majority class and neglect the minority class, which negatively impacts patients' lives and safety. In cases where the data are extremely imbalanced, the model may classify all sample data as belonging to the majority class. Due to the extremely high imbalance ratio, the final accuracy might be misleading, as it would be based solely on the majority class \cite{introduction7}.

Imbalanced data classification tasks are more complex than balanced data \cite{introduction7}. In these tasks, the model should pay attention not only to overall performance but also to improving the minority class performance. The majority class is known as the negative class, $N_{neg}$, and the minority class is known as the positive class, $N_{pos}$. The proportion of samples and the ratio between the sizes of the $N_{neg}$ class and the $N_{pos}$ class is called the imbalanced ratio (IR), as shown in Eq \eqref{eq1} \cite{introduction8}. Many researchers have proposed various solutions in recent years to tackle imbalanced problems. These solutions can be categorized into three families: data-level approaches, ensemble-based approaches, and algorithgm-level \cite{introduction9}.

\begin{equation}
IR=\frac{N_{Neg}}{N_{pos}}. \label{eq1}
\end{equation}

The data-level approach is known as resampling, which deals with the data directly by removing or adding sample data. Several algorithms have been introduced in this direction, which can be divided into three groups: random under-sampling (RUS), random over-sampling (ROS), and the hybrid approach \cite{introduction10}. RUS aims to remove data from the majority class using different techniques, such as Condensed Nearest Neighbors, Near-Miss, One-Side Selection, and Edited Near-est-Neighbor \cite{introduction11}.\cite{introduction12}.  \cite{introduction13}. \cite{introduction14}, among others. For ROS, the main purpose is to generate new synthetic data to tackle class imbalance issues. Several popular methods have been introduced, such as the synthetic minority over-sampling technique (SMOTE) algorithm \cite{introduction15}, which creates artificial instances by linearly interpolating between a minority class sample and its K-nearest neighbors. The hybrid approach combines both RUS and ROS. However, due to RUS being based on random removal of data, it can discard useful information that could help the model in prediction \cite{introduction16}, while ROS can suffer from weak recognition due to sample duplication during the generation process (data are identical with low heterogeneity). Different from the data-level approaches, the ensemble-based approach combines the data-level approach with one of the ensemble strategies, e.g., boosting or bagging, where several base classifiers are combined to improve the final decision. In this article, we are concerned only with the ensemble boosting-based approaches that combines data-level methods with boosting methods. These methods were used to perform cancer classification, such as the SMOTEBoost method \cite{introduction17}. This method achieved better results compared to other methods. However, although it is an  effective method for reducing the impact of class imbalance, some other challenges still hinder ideal solutions. For instance, class overlapping and noisy data are strongly related to imbalanced data. Class imbalance, overlapping, and noisy data have been hot research topics in recent years in the medical field \cite{introduction18}.

Data overlapping is described as the situation where different classes share a common area in the data space, e.g., the similarity between their features is high but they belong to different classes. When the minority class data in the overlapping regions is highly sparse due to insufficient training data, the classification model tends to bias the decision boundary toward the majority class \cite{introduction19}. To tackle class overlapping, various methods have been introduced in the literature at different levels, e.g., data-level approaches. For instance, the Tomek link has been introduced to remove noise and overlapping by creating pairs of links \cite{introduction20}, Edited Nearest Neighbor (ENN), which is based on removing noisy data from the majority class instances  \cite{introduction14}, One-Side Selection, Neighborhood-cleaning rule \cite{introduction13}, and others. In recent years, other methods have been introduced, such as DBMIST-US \cite{introduction22}, DB\_US, DB\_HS \cite{introduction23}, and others. However, as reported in [66], the main drawbacks of these methods include: (1) a lack of consideration for the potential information content of the majority instances during the elimination phase of minority class samples; (2) highly imbalanced datasets with minimal overlap, the method struggles to balance the data distribution due to insufficient elimination; (3) the KNN search rule is employed to identify the overlap region, with the k-value estimated based on the training data size and imbalance ratio. However, this approach assumes a uniform probability distribution, which is often not practical in real-world applications, such as medical diagnostics and bioinformatics. For boosting-based learning, the SMOTEBoost method uses the SMOTE over-sampling strategy, but it still suffers from overlapping issues \cite{introduction24}. Consequently, this leads to generating noisy data, which reduces the quality of synthetic samples. Additionally, to the best of our knowledge, no previous studies incorporate roulette wheel selection and information entropy to solve the aforementioned issues or use them for double pruning the majority and minority classes simultaneously in the SMOTEBoost algorithm.

To tackle the issues of imbalanced data, class overlapping, noisy data, and low-quality synthetic data in the SMOTEBoost algorithm, we propose a novel method, RE-SMOTEBoost, based on double pruning of the majority and minority classes, respectively. This method aims to reduce the size of the majority class by eliminating low-quality class instances and retaining only high-quality data using information entropy measurement while increasing the minority class by generating synthetic data in overlapping regions using roulette wheel selection to better capture the decision boundary. Secondly, a filtering mechanism based on information entropy computation is applied to reduce noise and borderline cases, thus improving the quality of generated data. Furthermore, a double regularization penalty has been proposed to reduce classification difficulty in inter-class overlapping regions and create a clear separation boundary by pushing synthetic samples farther away from the majority class and closer to the minority class.  The experimentation results demonstrated the effectiveness of the proposed methods, showing a significant improvement in accuracy of 3.22\% and an 88.8\% reduction in variance compared to the best-performing sampling methods.
Our major contributions to this study are as follows:
 
\begin{itemize}

 \item First, the overlapping region plays a critical role in the decision-making process. To improve the generated data in this region, we propose incorporating the roulette wheel selection technique into the SMOTEBoost procedure. This approach enhances the quality of synthetic minority samples, particularly in areas where the classes overlap.

\item Second, we propose an enhanced SMOTEBoost model, RE-SMOTEBoost, which reduces the generated noisy data by introducing a noise sample filtering strategy that can effectively prevent the generation of noise samples based on information entropy.

\item The proposed method aims to improve the reliability (confidence) and performance results by maintaining the structural distribution of data during the pruning process based on information entropy.	

\item The proposed method, RE-SMOTEBoost, reduces class overlapping by incorporating a similarity and difference double regularization penalty. The similarity loss ensures that generated samples are closer to the minority class distribution, while the difference loss pushes the generated samples away from the majority class distribution. 

\item No class overlap problem: Clear boundaries are defined between the majority and minority classes using legalization (Clear separation), guaranteeing that there is no convergence or resemblance between the classes.

\item Our proposed method prevents high bias toward the majority class (over-generalization) by relying on roulette wheel selection to generate minority synthetic data. This approach ensures that the essential features and distinct characteristics of the minority class are accurately preserved.

\item Our proposed method can be beneficial in large dataset scenarios by reducing the both majority and minority class sizes to tackle volume constraints and improve processing speed while enhancing dataset quality by minimizing noise, and class overlap, ultimately improving accuracy and reliability.

\item We carried out relevant experimental verification on nine real-world imbalanced datasets.

\end{itemize}

The rest of this paper is structured as follows. Section II summarizes the existing classification methods for imbalanced data and overlapping. Section III describes the methodology proposed in this study in detail. In Section  IV, detailed experimental verification is carried out, and the results are discussed and analyzed. Finally, the conclusion is drawn in Section V. 

\section{Imbalanced data background}

Imbalanced data, noisy data, and overlapping issues are recent challenges that hinder the model from achieving higher performance. Due to the nature of medical data, these challenges are common and occur frequently \cite{RelatedWork24}. Several solutions have been introduced in the literature in recent years, and each solution has its advantages and drawbacks depending on the case study. These solutions can be categorized into two main levels: data-level and ensemble-based approaches, as summarized in the following subheadings.
\subsection{Data-level approach}

This approach is known as the resampling strategy, which deals with the data directly rather than the model itself by adding or removing data from the minority or majority class, respectively. These approaches include many algorithms to tackle class imbalance and overlapping and may be categorized into three strategies: random under-sampling (RUS), random over-sampling (ROS), and the hybrid approach. RUS aims to modify data by removing instances from the majority class using different methods. For instance, All k-Nearest Neighbors (All k-NN) uses k-NN to classify each instance, where an instance is eliminated if most neighbors classify it erroneously \cite{RelatedWork24}. The Cluster Centroids method divides the majority class into clusters to reduce the size using k-means \cite{RelatedWork24}. Another reference \cite{RelatedWork26} mentioned the Edited Nearest Neighbor (ENN) method, where the idea is to perform k-nearest neighbors, and if all instances have been misclassified by the k-NN rule, they will be discarded. According to the same reference, the Condensed Nearest Neighbor (CNN) method has been proposed. CNN is an under-sampling technique that iteratively builds a subset T by adding instances from dataset D whose nearest neighbor in T does not match their label, continuing until no more instances can be added. However, this method is sensitive to noise, which leads to preserving noisy instances \cite{RelatedWork27}. Another method, called Tomek Links (TL), has been proposed, where TL is a pair of nearest-neighbor instances from different classes, and the majority class instance is removed to address boundary or noisy data \cite{introduction20}. However, TL might lead to over-pruning. Also, depending on the data distribution, certain original data may be considered noise data \cite{RelatedWork29}\cite{RelatedWork30}. Other extensions of TL have been introduced, such as SMOTE-Tomek Links, where the process starts by performing SMOTE synthetic data generation, followed by TL. However, this method requires high computation and results in information loss in the majority class \cite{RelatedWork30}. Near Miss is another method that has been introduced to select majority samples that are close to minority samples. This method has been extended to three different versions, e.g., NearMiss-1, NearMiss-2, and NearMiss-3, where each one has its core improvement \cite{RelatedWork31}. The One-Sided Selection (OSS) method has been proposed, which retains all minority class samples and selects misclassified majority class samples using 1-NN, as they are informative and near the decision boundary. It then removes the majority of class samples involved in Tomek Links, which indicate noise or redundancy \cite{RelatedWork32}. Although the RUS solution can balance the data and reduce class overlapping problems by deleting samples, its main drawback is that it can easily remove useful data that may contain important information \cite{RelatedWork28}. Recently, authors in \cite{RelatedWork33} proposed active learning based on information entropy to reduce the loss of informative data during the RUS procedure. However, applying this technique is limited to binary classification, and the number of minority samples should not be too small, which limits the effectiveness of the proposed method. In addition, all RUS methods focus on majority class reduction without considering minority class expansion. ROS is an over-sampling approach that aims to add data to the minority class. This direction includes several methods. For instance, the most common method is the synthetic minority over-sampling (SMOTE) algorithm \cite{introduction15}, which creates artificial instances by linearly interpolating between a minority class sample and its K-nearest neighbor. However, this algorithm blindly selects neighborhoods and overlooks boundary samples, leading to class overlapping near the boundary \cite{introduction24}.According to the same reference, other extensions of SMOTE have been developed to tackle this issue, such as the Borderline-SMOTE algorithm, which focuses on generating samples near the decision boundary region. However, this method works by expanding minority samples and ignoring the distributional characteristics of minority data \cite{RelatedWork35}. Another version, the ADASYN algorithm, has been proposed. It generates data from minority classes near the decision boundary, which is harder to classify \cite{RelatedWork36}. Wang et al. \cite{RelatedWork37} developed a combination of SMOTE with EE (SMOTE-ENN) to predict adverse outcomes using different classifiers, such as logistic regression (LR), k-nearest neighbor (KNN), support vector machine (SVM), random forest (RF), and extreme gradient boosting (XGBoost). Another enhancement has been proposed, named the MICE SMOTE-ENN algorithm, where Multivariate Imputation by Chained Equations (MICE) is combined with SMOTE-ENN \cite{RelatedWork38}. This 
method performs better than other methods, such as SMOTE-ENN, and outperforms AND\_SMOTE, DBSMOTE, GASMOTE, NRAS, NRSBoundary-SMOTE, SMOTE-OUT, SMOTE-PSO, SOICJ, and VIS\_RST in AUC and G-Mean metrics \cite{RelatedWork38}. The same result indicated that MICE-SMOTE-ENN works well in the presence of missing values. Recently, Zhai et al. \cite{RelatedWork39} proposed weighted SMOTE (WSMOTE) based on K-means and Intuitionistic Fuzzy Set theory to design the weight of existing samples and generate new synthetic data from them, performing the classification task using various methods such as kernel-free fuzzy quadratic surface support vector machine (QSSVM). To increase the performance of SMOTE, other words were proposed by Sowjanya and 
Mrudula \cite{RelatedWork40}, where they used Distance-based SMOTE (D-SMOTE) and Bi-phasic SMOTE (BP-SMOTE) techniques for breast cancer prediction. The results of experimentation indicated that these techniques increase accuracy compared to basic SMOTE. However, these techniques focus on generating data within the minority class without considering the majority class, which may impact overall performance. In addition, D-SMOTE used the parameter $\gamma$ to control the overlap between majority and minority classes, which can introduce extra noise in the form of unimportant variables while generating new synthetic data. Another SMOTE enhancement, M-SMOTE, has been proposed in \cite{RelatedWork41}, where the authors replace the Euclidean distance in the basic SMOTE with the Mahalanobis distance, as traditional SMOTE does not consider the coupling relationship between features. In addition, the kernel local Fisher discriminant analysis (KLFDA) algorithm is used to capture the global and local information of instances and for feature extraction. The classification task of fault diagnosis is built on AdaBoost.M2 and a decision tree with the Tennessee Eastman process. The results indicate that M-SMOTE achieved higher accuracy and F1-score compared to other methods. However, this method focuses on the minority class only while ignoring the impact of the majority class, which can affect the overall result or lead to overfitting and overlapping. Kumari et al. \cite{RelatedWork42} proposed a combination of SMOTE-Stacked hybrid methods for the early diagnosis of PCOS using six basic classifiers, e.g., LR, SVM, DT, RF, NB, and AdaB, on a PCOS imbalanced dataset. The results showed that Stack-AdaB achieved higher performance. However, the drawback associated with over-sampling is the overfitting of the classifier due to elongated training time duration, as duplicate instances are added to the minority class. Also, SMOTE has a higher computational cost than other undersampling methods.

\subsection{Ensemble-based approach}

This approach combines several methods to build strong classifiers to boost the final decision. It integrates the data-level approach with one of the ensemble strategies (e.g., ensemble bagging or ensemble boosting). The combination of the data level with ensemble bagging is called the bag-ging-based approach, while the combination of the data level with ensemble boosting is called the boosting-based approach. In this research, we focus on a boosting-based approach that com-bines a popular ensemble algorithm, the AdaBoost classifier, with one of the previous data-level algorithms. This direction includes various methods, such as the SMOTEBoost algorithm. This method combines the basic SMOTE algorithm with the AdaBoost classifier, where SMOTE generates new synthetic samples for the minority class and AdaBoost combines several base classifiers to increase the overall performance with many iterations \cite{RelatedWork43}. However, it has high computational complexity due to performing SMOTE in each iteration. Also, it does not consider the impact of synthetic minority samples on the model's performance \cite{RelatedWork44}. Moniz et al. \cite{RelatedWork45} proposed four variants of SMOTEBoost, e.g., (SMOTEBoost.RT, SMOTEBoost+, SMOTEBoost.R2, and SMOTEBoost.BEM) for imbalanced regression problems, where the goal was the prediction of extreme values. The results showed that the variants of SMOTEBoost outperformed baseline boosting methods for extreme target values. Jhamat et al. \cite{RelatedWork46} proposed a SMOTEMultiBoost extension that leverages the MultiBoost ensemble and SMOTE algorithm. The proposed method outperforms other methods such as CART, MultiBoost, SMOTE, BalanceCascade, EasyEnsem-ble, RUSBoost, and SMOTEBoost. However, this method may suffer from overlapping because it uses the same basic SMOTE to generate new samples. Recently, another boosting-based approach has been developed based on the SMOTEBoost idea. For instance, the RUSBoost method has been introduced to address the class imbalance by combining the data-level random under-sampling method with the AdaBoost algorithm \cite{introduction16}. It is a simpler and faster method than SMOTEBoost and prevents the classifier from being biased \cite{RelatedWork48}\cite{RelatedWork49}. However, this method can lead to the loss of useful information and class overlapping. Recently, new works have been introduced to solve the loss of useful information; for instance, the MPSUBoost method, where the authors used modified PSU and AdaBoost algorithms. Although this method achieved good results, it became weak when the error of false negatives was greater than that of false positives. Also, the likelihood of misclassifying
minority points can be high when the degree of complexity is high (e.g., majority and minority points are randomly distributed). Additionally, median sampling points in each partition are expected to achieve high precision. Other works in \cite{RelatedWork50} proposed the LIUBoost algorithm using under-sampling to balance data. This algorithm leverages the K-nearest neighbor (KNN) method and weight calculations to extract significant information about the local characteristics of each instance. It incorporates this information as cost terms in AdaBoost's weight update equation. However, these methods heavily depend on randomness, which can lead to poor classification performance for the underrepresented majority class. Moreover, they may randomly overlook selecting samples near the minority region (overlapping region). Another modification of SMOTEBoost is the CUSBoost method \cite{RelatedWork51}. To mitigate the loss of majority class information, the majority class instances are clustered into k groups using the k-means algorithm. Random under-sampling is then applied within each cluster by selecting instances randomly and discarding the rest. However, this approach requires the number of clusters to be predetermined, which is both arbitrary and unsuitable for datasets that do not lend themselves to clustering. Furthermore, the use of random under-sampling can result in the loss of valuable information. Rayhan et al. \cite{RelatedWork51} proposed a clustering-based under-sampling approach with boosting. However, this method requires a predetermined (arbitrary) 
number of clusters. Also, there is the possibility of obtaining false positives. Kumar et al. \cite{RelatedWork53} proposed the TlusBoost algorithm, which combines Tomek-Link with redundancy-based under-sampling and boosting. However, applying Tomek Link (TL) only between the two nearest neighbors (majority and minority class). Also, computing the distance between all examples is expensive. Ahmed et al. \cite{RelatedWork50} integrated a cost with an under-sampling and boosting approach into the Liuboost method. However, this method is based on randomness, which decreases reliability and fails to capture samples near the minority region (overlapping). Wang et al. \cite{RelatedWork55} proposed ECUBoost, based on the confidence of random forest and information entropy integrated into the boosting-based ensemble. However, this method requires slightly higher time complexity. Also, the entropy of the majority class was calculated once at the beginning and remained constant, with a weighting coefficient ($\Lambda $) balancing confidence and entropy during ranking. When $\Lambda$  = 0, it can result in low-quality samples and a reduced model performance. Conversely, when $\Lambda $ = 1, the ranking and selection depend entirely on static entropy, increasing the risk of including low-quality samples. Furthermore, while the method emphasizes harder-to-classify majority samples, it may overlook valuable ones, adversely affecting prediction accuracy. Although these various boosting-based approaches attempt to address class imbalance, all are based on one class (majority or minority) without taking the impact of both classes into account, which may lead to noise, overlapping data, and inter-class issues.

\subsection{Algorithm-level}

The algorithmic approach offers an alternative to data preprocessing for handling imbalanced datasets. Instead of altering the training set, this technique directly modifies the existing learning procedure or develops specialized algorithms to adjust the learning cost, remove the bias, and improve accuracy \cite{RelatedWork56}. This approach requires a good understanding of the procedure and its mathematical idea to determine the causes of failure in mining skewed distributions. A popular example of an algorithmic approach to address class imbalance is with a decision tree (DT) algorithm, where the splitting criteria of DT (entropy and Gini) are replaced by the Hellinger Dis-tance Decision Tree (HDDT) for creating the splitting of features \cite{RelatedWork58}. According to the same reference, the traditional support vector machine (SVM) has been modified using instance weight. Generally, this approach is divided into two main branches: one-class learning (OCC) and cost-sensitive learning (CSL). OCC aims to learn and capture the characteristics of training samples from one class (target class) due to the lack of non-target samples and ignores the remaining classes (outliers) \cite{RelatedWork57}. CSL is a widely used approach to address class imbalance by assigning different penalties (costs) to misclassification errors. Misclassifications of greater importance are given higher costs, while less critical errors are assigned lower costs. CSL uses a cost matrix to assess different costs instead of a traditional confusion matrix. Several CSL methods have been proposed in the literature. For instance, Zhang \cite{RelatedWork59} designed two CSL methods called Direct-CS-KNN and Distance-CS-KNN to reduce the misclassification cost. Other methods have been developed with boosting methods such as AdaCost, CSB1, CSB2, RareBoost, and AdaC1 \cite{RelatedWork60}. However, this approach suffers from the challenges of selecting the costs.

\subsection{Summary}

From the overview of the three solutions to imbalanced learning problems, it is evident that data-level, algorithm-level, and ensemble learning approaches have inherent challenges and weaknesses. In this article, we focus on the solutions related to a boosting-based approach, which particularly emphasizes the SMOTEBoost method to address class imbalance. Although scholars have applied this method to generate synthetic data and improve performance, they still face some difficulties: (1) The original SMOTEBoost model struggles to effectively train minority samples in overlapping regions. (2) While SMOTEBoost can generate samples with high generalization capability, they are also susceptible to overgeneralization due to failing to capture the data distribution (e.g., due to class overlap or noise). (3) Previous methods directly used the generated data for classifier training, without addressing the issue of noise within the generated data.

Based on this, this study proposes an improvement of the SMOTEBoost method called RE-SMOTEBoost to solve such problems. First, we employed the roulette wheel selection method using Mahalanobis distance as a fitness function to emphasize generating synthetic minority samples in overlapping regions to more effectively capture the decision boundary. Second, it incorporates a filtering mechanism based on information entropy to minimize noise, handle borderline cases, and enhance the quality of the generated data. Third, we designed a double regularization penalty to control the synthetic samples' proximity to the decision boundary and avoid class overlap using the Euclidean distance norm. Table \ref{tab1} Compares our proposed methods with state-of-the-art boosting-based methods across various criteria: Addressing class overlap refers to methods that handle the issue of overlap between classes. Existing noise/outlier filters refer to any strategies used to reduce the presence of noise. The type of sampling refers to the balancing strategy used. 'Losing useful information' refers to methods that use random under-sampling to balance the data, potentially leading to the loss of important information due to the inherent randomness. Class pruning refers to which class the method applies balancing techniques to, either the majority or minority class. The identification of the boosting iteration refers to the strategy for determining its value.

\begin{table*}[h]

\caption{A comparative summary of relevant existing boosting-based approaches.}

\label{tab1}
\setlength{\tabcolsep}{5pt}
\centering

\begin{tabular}{p{80pt}|p{50pt}|p{55pt}|p{60pt}|p{60pt}|p{70pt}|p{70pt}}
\hline
\textbf{Methods/ References} & \textbf{Tackling class overlapping} &\textbf{Existing noise / outliers filter } & \textbf{Type of  sampling} & \textbf{Lossing useful information due to randomness} & \textbf{identification of boosting iteration}& \textbf{Classes pruning}\\
\hline

RUSBoost \cite{introduction16}& \ding{56}& \ding{56}& Under-sampling& \checkmark& End-user &Majority\\

\pbox{20cm}{SMOTEBoost\\ \cite{introduction16},\\ \cite{introduction24} }&\ding{56}& \ding{56}& Over-sampling & \ding{56}& End-user&Minority \\

EUSBoost  \cite{Methododology62}&\ding{56}&\ding{56}& Under-sampling&\ding{56}& End-user&Majority\\

Cusboost \cite{Methododology62} &\ding{56}& \ding{56}&Under-sampling&\checkmark & End-user&Majority\\

TLUSBoost \cite{RelatedWork53} &\checkmark & \checkmark& Under-sampling&\checkmark & End-user&Majority\\

\pbox{20cm}{LIUBoost\\ \cite{RelatedWork50},\\  \cite{introduction16} }&  \checkmark & \ding{56}& Under-sampling& \ding{56}& End-user& Majority\\

ECUBoost \cite{RelatedWork55}&\ding{56}& \ding{56} &Under-sampling& \ding{56}& End-user&Majority\\

MPSUBoost  \cite{introduction16}&\ding{56}& \checkmark & Under-sampling& \ding{56}& End-user&Majority\\

\hline
Our work &\checkmark & \checkmark & Under and Over sampling& \ding{56}& Herustic formula&Minority \& Minority\\
\hline
\end{tabular}
\end{table*}

\section{Methodology}

Imbalanced data presents a significant challenge in medical diagnosis, as models often prioritize the negative class while overlooking the positive class, which can represent critical cases, such as cancer patients. To mitigate this issue, boosting-based approaches have been introduced in the literature as effective solutions for addressing data imbalance and improving overall performance. Among these methods, SMOTEBoost has achieved notable success and is widely regarded as a commonly used technique. This method is a combination of the SMOTE over-sampling technique and the AdaBoost.

\subsection{Basic concept of SMOTE over-sampling method}

SMOTE [15] It is a well-known over-sampling algorithm. It is built upon the k-nearest neighbor methodology. For each minority class instance,$x_{min}^i$, a sample $x_{min}^j$  is selected from its k-nearest neighbors, and a new minority class instance $ x_{min}^{'} $ is synthesized along linearly interpolating  using Eq \eqref{eq2} as follows \cite{introduction15}:

\begin{equation}
 x_{min}^{'}=x_{min}^i + rand(0,1)\cdot(x_{min}^j-x_{min}^i). \label{eq2}
\end{equation}

where rand $(\cdot)$ indicate a random number between intervals (0,1),  $x_{min}^{'}$ is a new sample synthesized.

The SMOTE algorithm uses over-sampling of minority classes by duplicating samples to balance the data, which may lead to duplicating low-quality samples while ignoring high-quality ones. In addition, it mainly focuses on the minority class without considering the impact of the majority class on overall performance. It also suffers from class overlap \cite{introduction24} and noise during the synthesis of new samples.

\subsection{Boosting approach}

Adaboost is a sequential algorithm that aims to increase global performance by focusing on misclassification errors. According to the full explanation mentioned in \cite{Methododology62}, Adaboost starts by assigning a standard weight to all dataset samples. Then, a base classifier is fitted to the sample data and their weights at each iteration $T_i$. Next, the error of the base classifier $g(x)$ is computed using  Eq \eqref{eq3} \cite{Methododology80}:

\begin{equation}
g(x)^{error}=\sum_{1}^{m} D(x_i),x_i \in S, g(x_i) \neq y_i. \label{eq3}
\end{equation}

Then, the weight of the base classifier is computed in Eq \eqref{eq4} as follows [62]:

\begin{equation}
g(x)^{error}=\log(\frac{1-g(x)^{error}}{g(x)^{error}}). \label{eq4}
\end{equation}

Finally, after Tmax iterations, the final boosting model for testing new samples is constructed using Eq \eqref{eq5} [80]:

\begin{equation}
H(x)=\sin(\sum_{1}^{T_{max}} \alpha_i\cdot g(x)). \label{eq5}
\end{equation}

\subsection{Key definition and notations}
A comprehensive overview of key definitions, symbols, and abbreviations used throughout the paper is provided in Table \ref{keydefintion}.

\begin{table}[H] 
    \centering
    \caption{List of symbols abbreviation.}
     \label{keydefintion}
    \begin{tabular}{p{60pt}|p{185pt}}
        \hline
        \textbf{Abbreviation} & \textbf{Meaning }\\
        \hline

      $\sigma $& The Variance\\

       $\mu$ & The mean\\

       Pre. & Precision metric\\

      Rec.  & Recall metric\\

 \multirow{3}{*}{RE-SMOTEBoost } &\textbf{R}: Roulette wheel selection \\
 & \textbf{E}: Entropy \\
  & \textbf{SMOTEBoost}: boosting method \\

   $Dist$& Euclidean distance \\

 $L_{Maj}$ & Euclidean distance norm  of majority class\\

  $L_{Min}$ & Euclidean distance norm  of minority class\\

  \ding{56} & No\\

 \checkmark  & Yes\\

   RMSE & The root mean square error\\
       
    RUS & Random under-sampling method \\
   ROS& Random over-sampling method \\

        \hline
    \end{tabular}
\end{table}

\subsection{Proposed RE-SMOTEBoost method}

\begin{figure*}[h]
	\centering
	\includegraphics[width=1\textwidth]{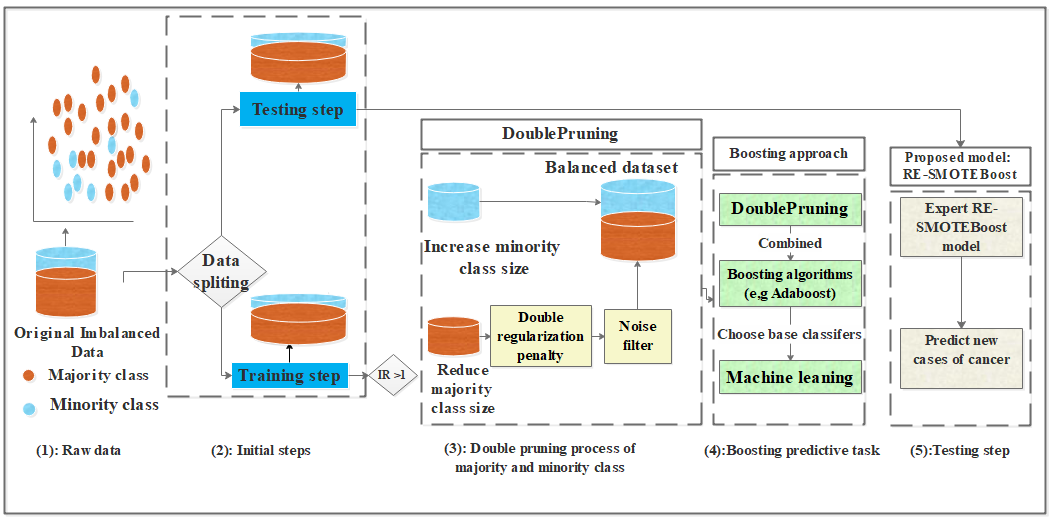}
	\caption{Overall structural flowchart of the proposed method.}
    \label{fig:fig1}
\end{figure*}

This section introduces the proposed algorithm, RE-SMOTEBoost, with its flow illustrated in Fig.\ref{fig:fig1}. The methodology consists of four main steps: (1) Starting with imbalanced raw patient data, (2) The data is split into a training set and a test set, with the test set retaining the original distribution for evaluation purposes. (3) The training set is balanced using the DoublePruning algorithm, which combines over-sampling, under-sampling, information entropy, double regularization penalties, and noise filtering. (4) Finally, predictive tasks are performed using an AdaBoost classifier.

\subsubsection{\textbf{Double pruning process based on DoublePruning procedure}}

\begin{figure*}[h]
	\centering
	\includegraphics[width=1\textwidth]{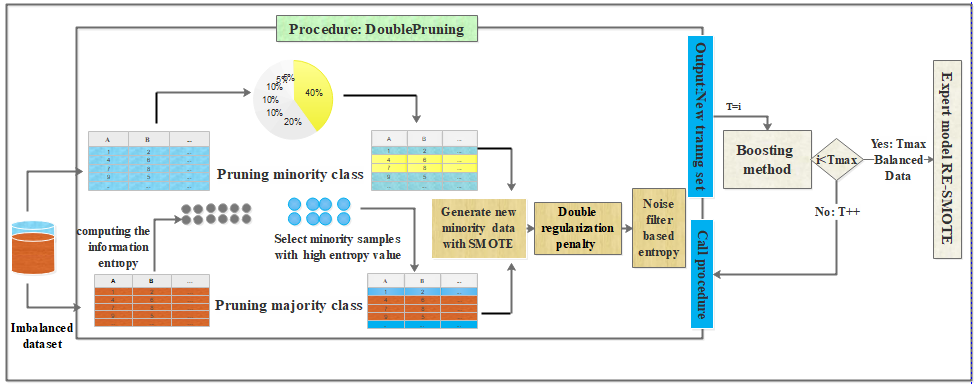}
	\caption{Double pruning process for majority and minority classes.}
    \label{fig:fig2}
\end{figure*}

The DoublePruning model architecture proposed in this study is shown in Fig.\ref{fig:fig2}. Unlike basic SMOTE, which generates synthetic samples only for the minority class, DoublePruning introduces a double-pruning procedure for both classes, as described in \textbf{Algorithm 1}. This involves under-sampling majority class samples using information entropy and over-sampling minority class samples through a roulette wheel selection strategy based Manhattan distance, with a focus on generating data in overlapping areas. To improve performance, two regularization penalties based on the Euclidean distance norm $(L_{maj}, L_{min})$ are applied to prevent synthetic samples from being too close to the majority class, which could negatively affect classification and decision boundaries. Additionally, a noise filter based on information entropy is used to eliminate noisy, and borderline samples. Finally, the pruned majority and minority samples are combined to construct incrementally balanced data, which optimizes the predictive model during each iteration.

\begin{algorithm}[!ht]
\DontPrintSemicolon

  \KwInput{
$S^{*}_{Maj}= \{\}=\varnothing $ : Newly pruned subset of the majority class  \\
$S^{*}_{Min}= \{\}=\varnothing $ : Newly pruned subset of the minority class }
   
   \vspace{0.2cm}
  
  \KwOutput{$S_{BalancedSet} $  \tcc{Return a balanced dataset (pruned dataset) }}

   \vspace{0.2cm}
  
  \SetKwFunction{FMain}{DoublePruning} 

  \SetKwProg{Fn}{Function}{:}{}
  \Fn{\FMain{$S_{Maj},S_{Min}$, k}}{
  
\vspace{0.2cm}

 $S^{*}_{Maj} \longleftarrow $ \textbf{MajorityClassPruning($S_{Maj}$, k)}\\
 $S^{*}_{Min} \longleftarrow $ \textbf{MinorityClassPruning($S_{Maj},S_{Min}$, k)}\vspace{0.3cm}\\
$S^{*}_{BalancedSet} \longleftarrow S^{*}_{Maj} \cup S^{*}_{Min}$
       
       \textbf{Return} $S^{*}_{BalancedSet}$    
  }
  
\caption{Double pruning}
\end{algorithm}

\subsubsection{\textbf{Majority class pruning}}

\begin{figure*}[h]
	\centering
	\includegraphics[width=1\textwidth]{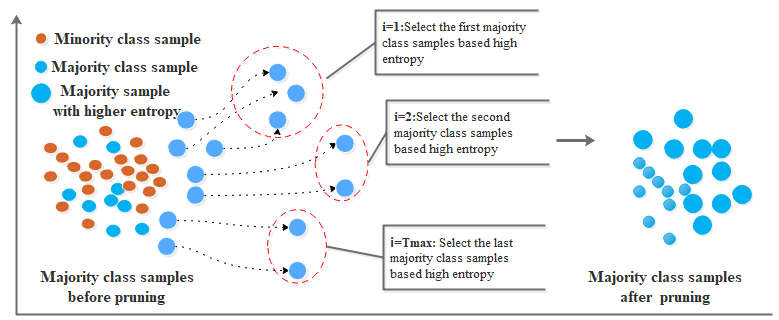}
	\caption{Prodecure of majority class pruning.}
    \label{fig:fig3}
\end{figure*}

The majority class refers to the class with the most data, which can cause overfitting during training. This overfitting may lead the model to prioritize the majority class, perceiving the minority class as noise and failing to recognize it during testing. In medical datasets, however, the minority class often contains more critical information. For instance, identifying whether the patient has cancer refers to a more important outcome than identifying a patient without cancer. To mitigate the dominance of the majority class while retaining its informative samples, a strategy based on information entropy is proposed. This approach involves computing entropy for all majority class samples to reduce the size by keeping samples with high entropy and ignoring those with low entropy, as outlined in  Fig.\ref{fig:fig3}. and \textbf{Algorithm 2}. This phase consists of four steps:

\begin{enumerate}

\item \textbf{ Step(1): Calculating the probability of  majority class samples (Algorithm 2, \# Line 3)}
Gaussian Naïve Bayes (GNB) is a widely used supervised model that employs a probabilistic approach based on the Gaussian distribution. It outperforms various algorithms, including linear discriminant analysis, decision trees, random forests, and quadratic discriminant analysis, among others \cite{RelatedWork33}. It requires minimal computational resources and training data to estimate the essential classification parameters efficiently \cite{Experimentation96}. This trait makes it particularly valuable in scenarios with limited data, such as the medical domain, where acquiring features is expensive. Recent studies \cite{Methododology85} have demonstrated that GNB is among the most effective algorithms for assessing cancer prognosis. Based on the aforementioned characteristics, GNB is applied in our study to calculate the probability of each sample of the majority class.  Let assume  $S_{Maj}=\{x_1,x_2,...,x_n\}$ represent the the features of majority class with correspondence target class $Y=\{ y_1 , y_2,…, y_n \}$. GNB used  Bayes’ theorem in   Eq \eqref{eq6} to calculate the posterior probability of samples.

\begin{equation}
P(Y |\{x_1,x_2,...,x_n\})=\frac{P(Y |\{x_1,x_2,...,x_n\})P(Y)}{P(\{x_1,x_2,...,x_n\})}. \label{eq6}
\end{equation}

GNB  use Gaussian probability density function to compute the posterior probability $P(x_i |y)$ for a given 
sample $x_i$ belonging to class majority class $y_i$ \cite{RelatedWork33}, as defined in Eq \eqref{eq7}. Here, $\mu_y $ denotes the mean, and  $\sigma_y^2$ represents the variance.

\begin{equation}
P(x_i|y)=\frac{1}{\sqrt{2\pi\sigma_y^2}}\cdot \exp(-\frac{(x_i-u_y)^2}{2\sigma_y^2}). \label{eq7}
\end{equation}

\item \textbf{Step(2): Computing the information entropy based on the posterior probability of majority samples (Algorithm 2, \# Line 4)}
Shannon introduced the concept of information entropy (H) as a measure of uncertainty or randomness within a system \cite{Experimentation92}. This mathematical tool helps quantify the information content (Q) associated with a particular event. The Shannon entropy is expressed by the formula in Eq \eqref{eq8}.

\begin{equation}
H(x_i)=-\sum_{y=2}^{y=1}P_y(x_i)\cdot\log(P_y(x_i)). \label{eq8}
\end{equation}

Where  $p(xi)$  represent probability of  the sample $x_i, (x_i \in S_{Maj})$ regarding y classes and the amount of data, $\log_2(p(x_i)) $. Samples with higher entropy H(X) indicate a greater amount of information, which significantly impacts the model's predictions.

\item \textbf{Step(3): Sorting the information entropy in ascending order (Algorithm 2, \# Line 5)}
In this step, the entropy values are ranked in ascending order.

\item \textbf{Step(4): Selecting K majority class samples based on the index of   high entropy values (Algorithm 2, \# Line 6-7)}
This step involves removing the majority of class samples with low entropy values and retaining those with high entropy values.

\end{enumerate}

\begin{algorithm}[!ht]
\DontPrintSemicolon

 \KwInput{$S^{*}_{Maj}= \{\}=\varnothing $ \tcc{Initialization of the majority class selection set (Empty)}}
 
  \KwOutput{$S^{*}_{Maj}$ \tcc{Return a new subset of high-quality majority class samples }} 

  \SetKwFunction{FMain}{MajorityClassPruning} 

  \SetKwProg{Fn}{Function}{:}{}
  \Fn{\FMain{$S_{Maj}$,k}}{
  
\vspace{0.2cm}

\ForEach{$x_i \in S_{Maj}$}{

 Calculate the probabilities P( $x_i \mid C_i$)  using Eq(7).\\
 Calculate the entropy H($x_i$) using Eq(8).\\
 Ranking the entropy values ascendant : $H_{ranked}(x_i$).\tcc{From lowest to highest value}
\vspace{0.1cm}
- The selected proportion of majority class samples,k, satisfied the condition as follow:

 $S^{*}_{Maj} \longleftarrow  S_{Maj} \cap  S_{Maj}[indice(H_{ranked}(x_i)), where \quad i  \le k] $
 \tcc{ Remove a proportion, K, of samples from the majority class based on their low entropy values}
 }
 \vspace{0.3cm}
     
       \textbf{Return} $S^{*}_{Maj}$
  }
  
\caption{MajorityClassPruning}
\end{algorithm}

\subsubsection{\textbf{Minority class pruning}}

\begin{figure*}[h]
	\centering
	\includegraphics[width=1\textwidth]{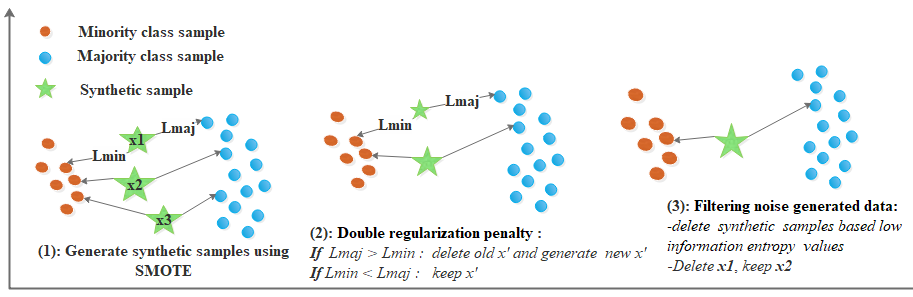}
	\caption{Prodecure of minority class pruning.}
    \label{fig:fig4}
\end{figure*}

The minority class has a low number of samples. However, the traditional synthetic method, SMOTEBoost, does not account for the area of class overlap where the informative data is. Also, it produces noise. To address this issue, we proposed pruning the minority class using roulette wheel selection based on Manhattan distance, double regularization penalties using Euclidean distance norm, and filtering noise-generated data based on information entropy, as described below in the subheading and  Fig.\ref{fig:fig4}.

\begin{enumerate}

 \item \textbf{Roulette wheel selection for  minority class reduction}
 Imbalanced data often leads to a sparse representation of minority samples in boundary regions, which are frequently treated as noise \cite{Methododology67}. However, boundary regions play a critical role in classification performance, particularly when classes overlap. Class overlap has the most significant negative impact among various factors affecting classification tasks, such as small disjuncts and dataset size \cite{Methododology68}. Furthermore, the SMOTE synthetic algorithm suffers from class overlapping \cite{introduction24}. Therefore, when applying the SMOTE technique to generate new sample data combined with the boosting approach, we aim for it to focus more on the effective learning of minority samples in the overlapping region so that minority class samples can receive more training opportunities. Based on this, we apply the roulette wheel selection-based Manhattan distance algorithm to assign varying probabilities to minority class samples and send them to the SMOTE algorithm for training and the generation of synthetic new samples. This allows minority samples in the boundary region to receive more learning, resulting in the creation of more effective minority synthetic samples in this region, which improves the classifier’s decision-making performance. Furthermore, we introduced a double regularization penalty to adjust the positioning of new synthetic data and a noise filter to reduce noise and outliers, contributing to the production of high-quality synthetic samples. These steps are described in \textbf{Algorithm 3} for minority class pruning.


\begin{algorithm}[!ht]
\DontPrintSemicolon

\KwInput{
\vspace{0.2cm}

$S^{}_{Syn}  = \{\}=\varnothing  $ \\

$S^{*}_{Min}  = \{\}=\varnothing $\\

\vspace{0.2cm}

$Dist^{Maj}_{i}:$ \textit{ A variable of the closest distance between  the synthetic sample and  real majority  sample}

\vspace{0.1cm}

$Dist^{Min}_{i}:$  \textit{ A variable of the closest distance between  the synthetic sample and real minority  sample }\\
 
}
\vspace{0.2cm}

  \KwOutput{$S^{*}_{Min}$ \tcc{Return a new subset of the minority class}} 

  \vspace{0.2cm}

  \SetKwFunction{FMain}{MinorityClassPruning} 

  \SetKwProg{Fn}{Function}{:}{}
  \Fn{\FMain{$S_{Maj},S_{Min}$,k}}{

$S^{'}_{min} \longleftarrow $   \textbf{Procedure RouletteWheelSelection($S_{Min}$)}

\ForEach{ $x_i \in S^{'}_{min}$}{
 $\alpha=rand (0,1) $.   \tcc{Select random number $\alpha$ from the range}
\vspace{0.1cm}

neighbors=find\_k\_nearest\_neighbors($x_i$)\\
$\bar{x}_i$=random\_choice(neighbors)\\

$\hat{x}_{i}=x_i +\alpha\cdot(\bar{x}_i-x_i$) \tcc{Syntthetic new  minority class samples \^x based on  roullete whelle selection filter}
\vspace{0.1cm}

\textbf{Double regularization penalty :}

\vspace{0.2cm}

$Dist^{Maj}_{i}\longleftarrow  L(\hat{x}_{i},x_y)$ , $ \forall x_y \in S_{Maj}$ using Eq(13).
\tcc{Apply  majority class regularization penalty filter}

\vspace{0.1cm}
$Dist^{Min}_{i}\longleftarrow  L(\hat{x}_{i},x_i)$ using Eq(14).

\tcc{Apply  minority class regularization penalty filter}

\If{$Dist^{Min}_{i} \leq Dist^{Maj}_{i}$}{}

$S^{}_{Syn} \longleftarrow S^{}_{Syn} \cup \{ \hat{x}_{i}\}$
}
\textbf{Noise filter:}
    
\ForEach{$ \hat{x}_{z} \in S^{}_{Syn}$}{

 Calculate the probabilities P( $\hat{x}_{z} \mid C_i$)  using Eq(7).\\
Calculate the entropy H($\hat{x}_{z}$) using Eq(8).\\
 Ranking the entropy values descendent: $H_{ranked}(\hat{x}_{z}$).\tcc{From highest to lowest value}
\vspace{0.1cm}
 Selecting a proportion of minority class samples,$S^{*}_{Min}$, that meet the following condition:

 $S^{*}_{Min} \longleftarrow  S_{Min}\cup \hat{x}_{z}[index(H_{ranked}(\bar{x}_i)), where \quad i \le k] $\vspace{0.3cm}

}      
       \textbf{Return} $S^{*}_{Min}$
  }
  
\caption{MinorityClassPruning}
\end{algorithm}

The roulette wheel method \cite{introduction8} is a stochastic selection method based on proportions. In the roulette wheel method, the wheel is divided into several sections, each representing a solution. The size of each sector is proportional to the selection probability assigned to its corresponding solution (minority class sample). A pointer, which remains fixed, points to a specific section. The wheel spins, and when it stops, the solution corresponding to the sector in front of the pointer is selected for the next generation. This process is repeated N times, with one solution chosen in each iteration.

This section concentrates on generating minority class samples in the overlapping area between classes. Specifically, it targets minority class samples that are near the majority class in the overlapping region as training data for the SMOTEBoost model, as described in Fig.\ref{fig:fig5}. Accordingly, the minority samples training strategy using the roulette wheel selection algorithm is outlined as follows.

\begin{figure*}[h]
	\centering
	\includegraphics[width=1\textwidth]{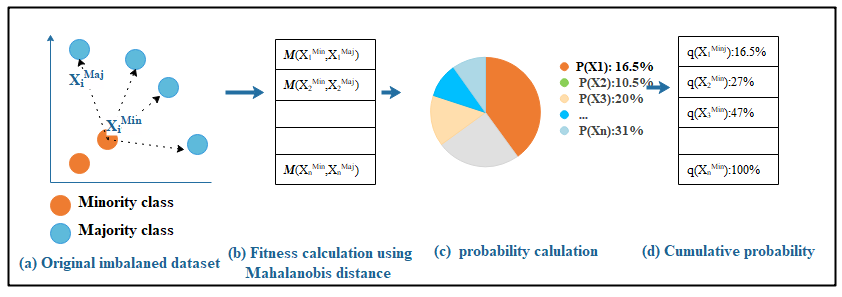}
	\caption{illustrates the flowchart of the roulette wheel selection algorithm for selecting minority class samples. (a) represents the original data with two classes (majority and minority), (b) shows the fitness function between minority and majority class samples based on the Mahalanobis distance D, (c) presents the probability values corresponding to the minority class samples derived from the fitness function, and (d) illustrates the cumulative probability values corresponding to the minority class samples, computed from the previous probability values.}
    \label{fig:fig5}
\end{figure*}

Let‘s give a dataset S   containing two classes, a majority class samples  $x_i^{maj}$ with n samples and a minority class with m samples $x_i^{min} $, where   $x_j^{maj},x_i^{min} \in S $.  The distance between  each minority sample $x_i^{min} $and  all majority class sample  is computed by Manhattan distance M as in Eq \eqref{eq9} \cite{Methododology81}:

\begin{equation}
  M(x_i^{Min},x_j^{Maj})=\sum_{1}^{n} \| x_i^{Min} - x_j^{Maj} \| . \label{eq9}
\end{equation}

Based on \cite{overalpping}, the minority samples that are closer to the majority class distribution are more likely to fall within the class overlap region. According to this definition, we can define the fitness function f(x) of the roulette wheel selection for selecting minority class samples as the reciprocal of the Mahalanobis distance $M_i$, as shown in Eq \eqref{eq10}. \cite{Methododology82}: 

\begin{equation}
 f(x_i)=\frac{1}{ M_i}. \label{eq10}
\end{equation}

Based on the definition of the fitness function f(x) for each minority class sample, the closer a minority class sample is to the majority class sample, the higher the fitness value of the minority sample. Based on this, the probability p(xi) and cumulative probability q(xi)of each selected minority sample can be computed as in  Eq \eqref{eq11} and  Eq \eqref{eq12} \cite{Methododology82}: 

\begin{equation}
 P(x_i)=\frac{f(x_i)}{\sum_{i=1}^{n}f(x_i) } ,i  \in\{1,2,...,m\}. \label{eq11}
\end{equation}

\begin{equation}
 q(x_i)=\sum_{j=1}^{m}P(x_i) . \label{eq12}
\end{equation}

In the iterative training process of  DoublePruning, when selecting minority samples for oversampling minority 
class, random numbers $r (where r\in (0,1))$ are generated. If  $q(x_{i-1}) \leq r < q(x_i)$, Then, the selected i-th minority sample is assigned to SMOTE for generating new synthetic samples in the boundary region. The complete procedure of roulette wheel selection is detailed in \textbf{Algorithm 4}.

\begin{algorithm}[!ht]
\DontPrintSemicolon

 \KwInput{$S^{'}_{Min}= \{\}=\varnothing $ \tcc{Initialization of the minority class selection set (Empty)}} 
 
  \KwOutput{$S^{'}_{Min}$ \tcc{Return the high-potential minority subset for generating synthetic new data}} 
  
  \SetKwFunction{FMain}{RouletteWheelSelection} 
  
  \SetKwProg{Fn}{Function}{:}{}
  \Fn{\FMain{$S_{Min}$}}{
  
\vspace{0.2cm}

- Evaluate the fitness function, f,  for each minority class samples, $x_{i}$, where $x_i \in S_{Min}$, using Eq(10)\\
-Calculate the corresponding probabilities p($x_{i}$) of each minority class samples using Eq(11).\\
- Computing the cumulative probabilities,q($x_{i}$),  of each minority class samples using Eq(12).\\ 

\ForEach{ $j \in Range (S_{Min})$}{

- Generate a uniformly distributed random number $r \in(0,1))$.\\

\uIf{q($x_{i-1}) < r_j < q(x_{i}$)}{

   $S^{'}_{Min} \longleftarrow  S^{'}_{Min} \cup \{x_{i} \} $ \tcc{Select the minority class sample,$x_{i}$}
   
  }
  \uElseIf{$ r_j< q(x_{1}$) }{
   Select the first minority sample,$x_{1}$, \;

   $S^{'}_{Min} \longleftarrow  S^{'}_{Min} \cup \{x_{1} \} $\tcc{ Select the first minority sample,$x_{1}$}
 
  }
}
  \textbf{Return} $S^{'}_{Min}$
  }
\caption{Roulette wheel selection}
\end{algorithm}

\item \textbf{Synxthetic minority  class-based regularization penalty}

The decision boundary plays a crucial role in the classification task. Indeed, the decision boundary of the trained classification model is often biased towards the majority class, making minority samples near the boundary more prone to misclassification \cite{Methododology70}. To ensure that synthetic samples are generated far from the majority class and closer to the minority class, creating a clear separation boundary and decreasing overlap in feature space, we designed a double regularization penalty filter. The double regularization penalty filter pushes synthetic samples far from the majority class and closer to the minority class. This can be described as the distance between synthetic samples and the two classes. Based on this, the distance between each synthetic sample and the majority class is calculated using the Euclidean distance norm $ L_{Maj}(x_y^{Min},x^{'})$.   Similarly, the distance between each synthetic sample minority class sample is calculated using  Euclidean distance norm $ L_{Min}(x_i^{Min},x^{'})$. the both regularization penalty filter $L_{maj}, L_{min}$ are defined as in Eq \eqref{eq13} and Eq \eqref{eq14}:

\begin{equation}
 L_{Maj}(x_y^{Maj},x^{'})= \| x_{Syn}^{'}-x_{y}^{Maj}\| . \label{eq13}
\end{equation}

\begin{equation}
 L_{Min}(x_i^{Min},x^{'})= \| x_{Syn}^{'}-x_{i}^{Min}\| . \label{eq14}
\end{equation}

Where $x_y^{maj}$,$ x_i^{min}$,and 	$ x_{Syn}^{'}$  represent the majority class, minority class, and synthetic samples, respectively. Based on the definition of $L_{min}$ and $L_{maj}$, it is evident that the feature distance between synthetic data and the majority and minority classes defines an optimization problem. The goal is to improve the position of synthetic data in the overlapping region by maximizing the distance from the majority class while minimizing the distance from the minority class. The mathematical formulation is given  Eq \eqref{eq15} and \eqref{eq16} below:

\begin{equation}
 L(x_y^{Maj},x^{'})= Max(Dist(x_y^{Maj},x^{'})) . \label{eq15}
\end{equation}

\begin{equation}
 L(x_i^{Min},x^{'})= Min(Dist(x_i^{Min},x^{'})) . \label{eq16}
\end{equation}

Where $Dist$ represents the Euclidean distance. From Eq \eqref{eq17}, the  optimization problem can be written as follows:

\begin{equation}
  L(x_i^{Min},x^{'}) < L(x_y^{Maj},x^{'})  . \label{eq17}
\end{equation}

\item \textbf{Filtering noise-generated data to improve synthetic data quality}

SMOTE algorithm generates new synthetic samples to over-sample minority classes, so it uses roulette wheel selection to choose a subset of minority samples to generate high-quality samples, aiming to improve results by maintaining the structural distribution of data. However, although this procedure increases the number of synthetic high-quality samples, it may generate a small amount of noisy data. Noisy data is defined as synthetic samples that have likely been generated in overlapping regions, samples generated near class boundaries, or blindly in majority class regions \cite{Methododology83}. These may not align well with the true data distribution, resulting in noisy or ambiguous data points. For this purpose, we use information entropy to evaluate synthetic samples, where those with high entropy are selected while those with low entropy are discarded.

\item \textbf{Boosting predictive task}

At this stage, the proposed method, DoublePruning, is integrated with a boosting algorithm (e.g., AdaBoost) to enhance the final results. DoublePruning serves as a balancing mechanism, applied at each iteration of AdaBoost, to simultaneously reduce the majority class size and increase the minority class size. The objective of this stage is to improve classification performance by increasing the weights of misclassified samples and decreasing the weights of correctly classified ones, thereby encouraging the classifier to focus on the misclassified samples. The final result is a combination of the results of multiple base classifiers using the majority weight average. The combination of DoublePruning and the AdaBoost classifier forms our proposed method, RE-SMOTEBoost, as outlined in the following key steps and detailed in the pseudocode provided in \textbf{Algorithm 5}.

\begin{itemize}
    \item  \textbf{Step (1):} The DoublePruning method is used as a balancing procedure to increase minority class and decrease majority class samples. The result of this procedure is a new subset S\_balancedSet (Algorithm 5, \#Line 3).

    \item  \textbf{Step (2):}  Train a base classifier I on subset  S with its correspondent weight w in (Algorithm 5, \#Line 5.

    \item   \textbf{Step (3):} Calculate the weight of misclassified samples.
    
    \item   \textbf{Step (4):} Update the weight of each trained sample according to the base classifier's error. Increase the weight of misclassified samples and decrease the weight of correctly classified ones (Algorithm 5, \#Line 8).
    
    \item  \textbf{Step (4):} Normalize the weight of the sample weights to maintain a consistent scale (Algorithm 5, \#Line 9).
    
    \item  \textbf{Step (5):} Repeat the boosting process and its steps (1) to (4) for Tmax iterations until the data becomes balanced.

    \item  \textbf{Final steps:} utilizing the boosted classifier to make predictions on new, unseen samples (Algorithm 5, Output).

\end{itemize}

\begin{algorithm}[!ht]
\DontPrintSemicolon

  \KwInput{\\
  Training set:$S=(x_i,y_i)$, where $y_i \in \{yes, no \} $\\
 
  I: Boosting base classifer\\
  $\omega(x_i)=\frac{1}{N}$ , where $ \forall x_i \in \{1,...,N \}$ \\
  
  $S_{Maj}$:Majority class samples\\
   $S_{Min}$:Minority class samples\\
  
  $T_{Max}=\frac{(S_{Maj}-S_{Min})}{2\cdot K}$ (\textbf{Herustic formula})\\

  $T_{Max}$: number of boosting iteration \\
  k: required samples for balancing dataset \\
  }

  \KwOutput{$F(x_i) =\sin\sum_{t=1}^{T_{Max}} (\alpha\cdot h_{t} (x_i))$ \tcc{ Final booted  cancer  model}}

  \SetKwFunction{FMain}{RE-SMOTEBoost} 

  \SetKwProg{Fn}{Function}{:}{}
  \Fn{\FMain{$S_{Maj},S_{Min},k $}}{
  
\vspace{0.2cm}

 \For  {i  to $T_{Max}$} {

$S_{BalancedSet}=\textbf{DoublePruning ($S_{Maj},S_{Min}$, k)}$\\

$\omega_t^{'}=\{d_i \in \omega_i, (x_i,y_i) \in S_{BalancedSet} \} $\\

$h_i=I(\omega^{'},S^{'})$

$E_t=\sum_{i,y \ne h(x_i)}^{N} \omega_i^{t+1}$

$\alpha_t=\frac{1}{2}\ln (\frac{1-E_t}{E_t})$

$\omega_i^{t+1}\longleftarrow \omega_i \cdot\exp(-\alpha_t\cdot y_i\cdot h(x_i)) \forall x_i \in \{1,...,N \}$

$\omega_i^{t+1} \longleftarrow \frac{\omega_i^{t+1}}{\sum_{i=1}^{N} \omega_i^{t+1}}$

 }

 \textbf{Return} $S^{*}_{Maj}$
  }
  
\caption{RE-SMOTEBoost}
\end{algorithm}

\end{enumerate}

\section{Experimentation}
 
To evaluate the proposed method and its effectiveness, the improvement in effectiveness in addressing class imbalance, improvement in reliability, overlapping problems, and noisy data have been evaluated. The following subsections cover key elements of our research study, including assessment metrics, baseline models, benchmark datasets, and parameter configuration. It concludes with the main findings and discussion.
    
\subsection{Assessment metrics}

To evaluate the improvement in effectiveness of the proposed method, RE-SMOTEBoost, on imbalanced medical data, we used several evaluation metrics that have been recommended based on the state of the art. The classification accuracy metric is defined as the ratio of correctly classified samples to the total number of samples, as shown in Eq \eqref{eq18} \cite{RelatedWork33}. The recall represents the ratio of correctly classified positive samples to the total number of actual positive samples, as shown in Eq \eqref{eq19} \cite{Methododology70}. According to the same references, additional metrics such as precision, G-means, AUC curve, and F1-score are also used for imbalanced data. Precision is the proportion of correctly identified positive samples among all examples predicted as positive, as shown in Eq \eqref{eq20}. F1-score is the weighted harmonic mean of precision and recall, as described in Eq \eqref{eq21}. The AUC (area under the curve) represents the area beneath the ROC curve and serves as a numerical indicator of classifier quality. A higher AUC value (close to the upper right) corresponds to better classification performance. G-means evaluates the result based on both classes at the same time; it represents the geometric mean of recall and precision, as shown in Eq \eqref{eq_means}.

\begin{equation}
 Accurcay= \frac{TP +TN}{TP + FP + FN +TN}. \label{eq18}
\end{equation}

\begin{equation}
 Recall= \frac{TP}{TP + FN}. \label{eq19}
\end{equation}

\begin{equation}
Precision= \frac{TP}{TP + FP}. \label{eq20}
\end{equation}

\begin{equation}
F_1-score= \frac{2\cdot Recall \cdot Precision}{Recall+ Precision}. \label{eq21}
\end{equation}

\begin{equation}
G-means= \sqrt{Precision\cdot Recall}. \label{eq_means}
\end{equation}

Where TP represents the number of true positive instances, TN represents the number of true negative instances, FP denotes the number of false positive instances, and FN denotes the num-ber of false negative instances.

To evaluate the reliability improvement, the variance metrics have been used. The variance measurement used standard deviation $(\sigma)$ to assess the degree of spread or stability in the prediction results. Low variance indicates greater stability \cite{Methododology84}, as defined in Eq \eqref{eq22}.

\begin{equation}
\sigma (P) =\sqrt{\frac{1}{(L-1)}\cdot\sum_{1}^{L}(E-\bar{E}}).\label{eq22}
\end{equation}

Where L indicates the number of run-time executions (replication), E indicates the result of one of the previous metrics (e,g accuracy, recall) and $\bar{E}$ indicates the average of  L execution.

\subsection{Imbalanced benchmark dataset }

To evaluate the effectiveness of the proposed method for cancer diagnosis, a set of imbalanced datasets is utilized. Table \ref{DATASET} describes each dataset along with its characteristics. For every dataset, we provide the dataset name, total number of instances, number of features (attributes),  and the imbalance ratio. In this study, the imbalance ratio (IR) is defined as the percentage of majority instances to minority instances, i.e., $IR=\frac{N_{Maj}}{N_{Min}}$. All datasets are standard real-world data from the cancer domain, specifically designed for binary classification tasks, and are publicly available on the UCI Machine Learning Repository and Kaggle website. The dataset features range from a small number of attributes to a large number.

\begin{table*}[h]
\caption{Description of cancer dataset.}

\label{DATASET}
\setlength{\tabcolsep}{5pt}
\centering

\begin{tabular}{p{150pt}|p{80pt}|p{80pt}|p{60pt}}
\hline
\textbf{Cancer dataset} & \textbf{Number of features} &\textbf{Number of instances} & \textbf{IR} \\
\hline

Breast cancer wisconsin diagnostic (WDBC)& 32&569& 1.68\\

Breast cancer wisconsin original (WOBC)& 9&699& 1.90\\

Mammographic Mass& 5&961& 1.16\\

Breast Cancer Coimbra & 9& 116 & 1.23\\

Breast cancer & 9& 286 &2.36\\

Real Breast Cancer Data & 16& 334 & 4.06\\

Haberman's Survival breast  & 4&306&2.78\\

Breast Cancer Survival-preproceed  & 11&1207&14.08\\

21239\_breast\_cancer\_survival & 10&2000&1.04\\
\hline
\end{tabular}
\end{table*}

\subsection{Baseline  methods and parameters settings}

Through this study, we aim to provide extensive experimentation to prove the effectiveness of the proposed methods. To do this, we compare the proposed method with a set of state-of-art methods.  These methods are similar to our proposed method in many characteristics (e,g; integrated boosting method, solving class imbalance), making them very suitable for this comparison. The complete comparison consists of approximately 14 algorithms, categorized into several groups: ensemble boosting-based methods (e.g., RUSBoost, CUTBoost) and data-level methods (e.g., SMOTE), bagging methods.

Since the proper selection of hyperparameters can significantly affect the performance of models, we set the parameters for the base classifiers, and, boosting-based approach according to the values recommended in the literature. For the boosting iteration, $T_{max}$, and the number of over/ under samples, k we applied the proposed heuristic formula as defined in (\textbf{Algorithm 5, Input)} for both the proposed method and the state-of-the-art methods that require boosting iterations. The parameters and their corresponding values are specified in Table \ref{tab2}.


\begin{table*}[h]

\caption{Overview of the configuration settings.}

\label{tab2}
\setlength{\tabcolsep}{5pt}
\centering

\begin{tabular}{p{100pt}|p{110pt}|p{180pt}}
\hline
\multicolumn{2}{c|}{Methods} &\textbf{Values/ references} \\
\hline

\multirow{4}{*}{Base classifers}&Decision tree (DT)&Number of trees: 1 (decision stumps) \cite{Experimentation103}\\ \cline{2-3}

&Support vector machine (SVM)&Probability:True,
C:1.0, gamma: 1 \\\cline{2-3}

& Random forest (RF)&n\_estimators=300,   max\_depth=10,    min\_samples\_split=10,         min\_samples\_leaf=5 [101]\\\cline{2-3}
 &  k-nearest neighbors (KNN)&n\_neighbors=10 \cite{Experimentation102} \\\cline{2-3}
 & Naïve Bayes(NB)&var\_smoothing=1.0 \cite{Experimentation100}\\
\hline
Boosting-based approaches& 
\makecell{Proposed method,\\
RUSboost,\\
SMOTEBoost}
& \makecell{Boosting iteration: $T_{max}$\\
Number of  over/ under samples: k\\
Learning rate = 1.0\\
Random state: 42}
 \\

\hline
Run-time executions&All methods  &L: 100 replications, 20 CV \\

\hline
\multicolumn{2}{c|}{LightGBM} & num\_leaves=25,learning\_rate=0.1, n\_estimators=100, 
    max\_depth=8
 \\
\hline
\multicolumn{2}{c|}{EasyEnsemle}& n\_estimators=9 \cite{Experimentation100} \\
\hline
\multicolumn{2}{c|}{Bagging} & n\_estimators=40\cite{Experimentation99} \\
\hline
\multicolumn{2}{c|}{Booting}& n\_estimators=10\cite{Experimentation99}\\

\hline
\multicolumn{2}{c|}{CUTBoost}& n\_neighbors=5 \cite{introduction16} \\

\hline
\multicolumn{2}{c|}{MPSUBoost}& n\_estimators=100 \cite{introduction16} \\
\hline
\end{tabular}
\end{table*}

\subsection{Experimentation design and configuration }
This section explains the dataset preparation process for our experiments. We applied a train-test split to ensure reliable evaluation, allocating 80\% of the data for training and 20\% for testing. To validate the stability of the results, the experiments were repeated L=100 times, and the outcomes were computed as the average performance across all replications for the relevant metrics. This approach ensures consistency and robustness when evaluating the models. To do this, we conducted the experiments using Python software and the Windows operating system, as described in Table \ref{tab4}.

\begin{table*}[h]
\caption{Software and libraries  of experiment.}

\label{tab4}
\setlength{\tabcolsep}{5pt}
\centering

\begin{tabular}{p{130pt}|p{100pt}|p{100pt}}
\hline
\textbf{Tools} & \textbf{Types} &\textbf{Version} \\
\hline

Python& Programming language& 3.8.5\\

Jupyter NoteBook&web-based application & 6.1.4\\
Imblearn, NumPy, sklearn& Python libraries& 0.6.0 ,1.19.2, 0.22, respectively\\
Windows& Operating system & Windows 10, 64-bit\\
\hline
\end{tabular}
\end{table*}

\subsection{Experimental  results}

In clinical practice, cancer data often experiences class imbalance due to limited samples, data collection difficulties, and privacy concerns, which impact research and model accuracy. To address this, in this section, we assess the proposed imbalanced learning method from several aspects: analysis of the effectiveness of the improvement, analysis of the reliability improvement, reducing noise, and minimizing class overlap. The evaluation results are discussed in relation to the research questions: 

\begin{itemize}
    
    \item \textbf{RQ1:} How effective is the proposed imbalanced learning method in improving model performance for cancer data with class imbalance? 

 \item  \textbf{RQ2:} To what extent does the proposed method enhance the reliability of diagnostic models in the presence of imbalanced cancer data? 

 \item  \textbf{RQ3:} How does the proposed method reduce noise and minimize class overlap in imbalanced cancer datasets? 

 \item  \textbf{RQ4:} How does the roulette wheel selection method compare to other selection techniques in terms of efficiency in sample selection?
\end{itemize}

This study aims to offer valuable insights into the comparative performance of the proposed method with various state-of-the-art imbalanced data approaches. To achieve this, the proposed model, RE-SMOTEBoost, is evaluated using the set of imbalanced datasets. The proposed method is designed to address the stated research objectives and questions through performance metrics such as accuracy, precision, F1-score, G-means, recall, and AUC/ROC values. The evaluation employs splitting data into training (80\%) and testing (20\%) and a cross-validation approach with L=100 executions. The findings are presented under the following subheadings.

\subsubsection{\textbf{Evaluating the effectiveness of diagnostic improvement}}

In this section, we compare the performance of the proposed RE-SMOTEBoost method with popular data augmentation and imbalanced learning techniques. These methods are divided into several categories e,g; data-level approaches, bagging methods, and ensemble boosting approaches. we use several base classifers such as Decision Trees and Support Vector Machines with relevant evaluation metrics such as precision and recall. The data-level methods were chosen for their focus on addressing class overlap and noise, making them suitable for experimental comparison with our approach. Meanwhile, the selected boosting methods aim to enhance prediction performance. Both selection criteria align with the primary objective of this study. Table 6 and Table 12 summarize the comparative experimental results with different datasets. The bold values represent the highest performance.

Table \ref{tab5} shows the comparative experimental results of the proposed method against a boosting-based approach using DT and SVM base classifiers for four relevant metrics. Based on the findings cited in the table, the proposed method outperforms others across different datasets for both base classifiers. It can be seen that DT-RE-SMOTEBoost has achieved competitive results compared to the model without sampling methods (None model), with the corresponding difference values, e.g., precision (3.94\%), recall (2.71\%), f-score (3.68\%), and g-means (2.71\%) for WOBC, and precision: 3.94\%, recall: 2.71\%, F-score: 3.68\%, and G-means: 2.71\%. It is worth noting that DT-RE-SMOTEBoost achieved better performance on the WOBC dataset compared to the other datasets.

While applying the WDBC dataset, the results showed that the proposed DT-RE-SMOTEBoost algorithm ranked first compared to the None model regarding recall, F-score, and G-means metrics, with the corresponding difference values (6.54\%), (3.86\%), and (6.54\%), respectively. However, it is worth noting that DT-RE-SMOTEBoost ranked third in terms of precision, with a very slight difference compared to the second model (None), with a difference value of 0.3\%. For the same dataset with the SVM classifier, the proposed method achieved similar results to another model, except for the precision metric, where SVM-SMOTEBoost achieved a higher ranking with a corresponding value of 34.44\%. For the Coimbra and Breast cancer dataset,  DT-RE-SMOTEBoost and  SVM-RE-SMOTEBoost outperforms other methods.

\begin{table*}[h]
\renewcommand{\arraystretch}{0.6} 
\setlength{\tabcolsep}{5pt} 

\caption{Diagnosis testing efficiency of the proposed method vs. state-of-the-art boosting methods.}

\label{tab5}
\setlength{\tabcolsep}{8pt}
\centering
\resizebox{\textwidth}{!}{ 
\begin{tabular}{l c c c c c c c c c}

\hline
\multirow{2}{*}{Cancer dataset} & \multirow{2}{*}{Boosting methods}  & \multicolumn{4}{c}{DT classifier} & \multicolumn{4}{c}{SVM classifier} \\ 
\cmidrule(lr){3-6} \cmidrule(lr){7-10}  
&  & Precision($\%$) & Recall(\%) & $F_1$-score($\%$) & G-means($\%$) & Precision($\%$) & Recall($\%$) & $F_1$-score($\%$) & G-means($\%$)\\ 
\midrule 

 \multirow{4}{*}{WOBC} 
        & None & 89.15 & 91.94 & 90.10 & 91.94 & 94.06 & 91.62 & 92.67 & 91.62 \\
        & SMOTEBoost & 89.18 & 91.91 & 90.10 & 91.91 & 93.80 & 87.30 & 89.53 & 87.31 \\
        & RUSBoost & 89.15 & 91.94 & 90.10 & 91.94 & 93.54 & 90.58 & 91.81 & 90.58 \\
        
        & RE-SMOTEBoost & \textbf{93.09} &\textbf{ 94.65 }& \textbf{93.78 }& \textbf{94.65 }& \textbf{94.27} & \textbf{94.70} &\textbf{ 94.48} &\textbf{ 94.70 }\\\\
       
        \multirow{4}{*}{WDBC}
        & None & \textbf{89.55 }& 84.33 & 86.00 & 84.33 & 31.42 & 49.31 & 38.38 & 49.31 \\
        & SMOTEBoost & \textbf{90.38 }& 86.30 & 87.65 & 86.30 &\textbf{ 34.44} & 49.61 & 37.68 & 49.61 \\
        & RUSBoost & 86.84 & 89.58 & 86.61 & 89.58 & 31.42 & 49.31 & 38.38 & 49.31 \\
        
        & RE-SMOTEBoost & \textbf{89.25} & \textbf{90.87 }& \textbf{89.86 }&\textbf{ 90.87} &\textbf{ 31.42} &\textbf{ 49.31} & \textbf{38.38 }&\textbf{ 49.31} \\\\
      
        \multirow{4}{*}{Mammographic} 
        & None & 81.60 & 81.72 & 81.34 & 81.72 & 30.68 & 30.95 & 30.79 & 30.95 \\
        & SMOTEBoost & 83.00 & 82.92 & 82.38 & 82.92 & 64.52 & 67.01 & 64.69 & 67.01 \\
        & RUSBoost & 82.51 & 82.68 & 82.37 & 82.68 & 70.47 & 69.84 & 69.92 & 69.84 \\
        & RE-SMOTEBoost& \textbf{83.61} &\textbf{ 83.49} & \textbf{82.90} & \textbf{83.49} & \textbf{71.65 }& \textbf{70.80} & \textbf{70.89} & \textbf{70.80} \\\\

        \multirow{4}{*}{Breast Cancer Coimbra} 
        & None & 62.50 & 62.59 & 62.43 & 62.50 & 27.08 & 50.00 & 35.13 & 50.00 \\
        & SMOTEBoost & 61.52 & 61.51 & 61.34 & 61.51 & 27.08 & 50.00 & 35.13 & 50.00 \\
        & RUSBoost & 53.57 & 53.50 & 53.44 & 53.50 & 22.92 & 50.00 & 31.43 & 50.00 \\
        & RE-SMOTEBoost&\textbf{ 62.50} &\textbf{ 62.59 }& \textbf{62.43 }& \textbf{62.50 }& \textbf{27.08 }& \textbf{50.00} &\textbf{ 35.13} &\textbf{ 50.00 }\\\\
        
        \multirow{4}{*}{Breast cancer} 
        & None & 69.02 &\textbf{ 69.66} & 69.31 & \textbf{69.66 }& 35.34 & 50.00 & 41.41 & 50.00 \\
        & SMOTEBoost & 65.46 & 66.93 & 65.90 & 66.93 & 27.90 & 50.00 & 34.67 & 50.00 \\
        & RUSBoost & 63.10 & 65.78 & 61.88 & 65.78 & 35.34 & 50.00 & 41.41 & 50.00 \\
        & RE-SMOTEBoost & \textbf{78.68} &\textbf{ 68.15} & \textbf{70.26} &\textbf{ 68.15} & \textbf{35.34 }& \textbf{50.00 }& \textbf{41.41 }& \textbf{50.00} \\

\midrule
\end{tabular}

}
\end{table*}

Table \ref{tab6} represents the comparison results of the proposed method against well-known data-level methods. These methods are designed to address class imbalance, minimize overlap, and reduce noise and outliers, all of which are closely related to our objectives. The methods are categorized into over-sampling methods (e.g., SMOTE) and under-sampling methods (e.g., Tomek Link). As can be seen from the table, RE-SMOTEBoost outperforms all other methods for both classifiers (DT and SVM) across all datasets. For example, DT-RE-SMOTEBoost demonstrated superior performance compared to the baseline model (None), with a notable difference in precision values (3.94\%), (2.71\%), (3.68\%), (2.71\%). However, it’s worth noting that some methods achieved similar performance to that of the proposed method with the SVM classifier.

\begin{table*}[h]
\renewcommand{\arraystretch}{0.6} 
\setlength{\tabcolsep}{5pt} 

\caption{Diagnosis testing efficiency of the proposed method vs. data-level state-of-the-art methods. }

\label{tab6}
\setlength{\tabcolsep}{5pt}
\centering
\resizebox{\textwidth}{!}{ 
\begin{tabular}{l c c c c c c c c c}

\hline
\multirow{2}{*}{Cancer dataset} & \multirow{2}{*}{Data-level methods }  & \multicolumn{4}{c}{DT classifier} & \multicolumn{4}{c}{SVM classifier} \\ 
\cmidrule(lr){3-6} \cmidrule(lr){7-10}  
&  & Precision($\%$) & Recall(\%) & $F_1$-score($\%$) & G-means($\%$) & Precision($\%$) & Recall($\%$) & $F_1$-score($\%$) & G-means($\%$)\\ \\ 
\midrule 

\multirow{5}{*}{WOBC} 
        & SMOTE & 89.15 & 91.94 & 90.10 & 91.94 & 92.11 & 89.14 & 90.35 & 89.14 \\
        & ADASYN & 89.52 & 92.21 & 90.46 & 92.21 & 91.46 & 87.58 & 89.08 & 87.57 \\
        & Tomek Link & 89.15 & 91.94 & 90.10 & 91.94 & 94.27 & 94.70 & 94.48 & 94.70 \\
        & Borderline SMOTE & 89.45 & 92.16 & 90.39 & 92.16 & 92.00 & 87.94 & 89.50 & 87.94 \\
        & RE-SMOTEBoost & \textbf{ 93.09} & \textbf{ 94.65 }& \textbf{ 93.78} &  \textbf{94.65 }&  \textbf{94.27} &  \textbf{94.70} &  \textbf{94.48} & \textbf{ 94.70} \\\\

        \multirow{5}{*}{WDBC} 
        & SMOTE & 89.95 & 85.66 & 87.00 & 85.66 & 31.42 & 49.31 & 38.38 & 49.31 \\
        & ADASYN & 89.68 & 88.66 & 88.77 & 88.66 & 31.42 & 49.31 & 38.38 & 49.31 \\
        & Tomek Link & 92.35 & 84.52 & 86.70 & 84.52 & 31.42 & 49.31 & 38.38 & 49.31 \\
        & Borderline SMOTE & 89.67 & 85.30 & 86.66 & 85.30 & 31.42 & 49.31 & 38.38 & 49.31 \\
        & RE-SMOTEBoost &\textbf{ 89.25} & \textbf{90.87 }&\textbf{ 89.86} &\textbf{ 90.87} &\textbf{ 31.42} & \textbf{49.31} &\textbf{ 38.38} &\textbf{ 49.31} \\\\

        \multirow{5}{*}{Mammographic} 
        & SMOTE & 83.00 & 82.93 & 82.42 & 82.93 & 34.38 & 34.34 & 34.33 & 34.34 \\
        & ADASYN & 83.41 & 83.41 & 82.90 & 83.41 & 68.75 & 68.90 & 68.74 & 68.90 \\
        & Tomek Link & 82.87 & 82.65 & 82.51 & 82.65 & 32.49 & 32.49 & 32.49 & 32.49 \\
        & Borderline SMOTE & \textbf{84.15 }& 83.29 & \textbf{83.27 }& 83.29 & 58.24 & 58.25 & 58.25 & 58.25 \\
       & RE-SMOTEBoost & \textbf{83.61} & \textbf{83.49} & \textbf{82.90} & \textbf{83.49} & \textbf{71.65} & \textbf{70.80} & \textbf{70.89} & \textbf{70.80} \\\\

        \multirow{5}{*}{Breast Cancer Coimbra} 
        & SMOTE & 60.18 & 60.22 & 60.10 & 60.22 & 22.92 & 50.00 & 31.43 & 50.00 \\
        & ADASYN & 62.50 & 62.43 & 62.21 & 62.43 & 22.92 & 50.00 & 31.43 & 50.00 \\
        & Tomek Link & 62.50 & 62.59 & 62.43 & 62.50 & 27.08 & 50.00 & 35.13 & 50.00 \\
        & Borderline SMOTE & 59.65 & 59.45 & 59.23 & 59.45 & 22.92 & 50.00 & 31.43 & 50.00 \\
      & RE-SMOTEBoost & \textbf{62.50} & \textbf{62.59} & \textbf{62.43} & \textbf{62.50} & \textbf{27.08} & \textbf{50.00} & \textbf{35.13} & \textbf{50.00} \\\\

        \multirow{5}{*}{Breast Cancer} 
        & SMOTE & 69.10 & 64.17 & 65.30 & 64.17 & 14.66 & 50.00 & 22.67 & 50.00 \\
        & ADASYN & 71.01 & 65.58 & 66.90 & 65.58 & 14.66 & 50.00 & 22.67 & 50.00 \\
        & Tomek Link & 65.57 & 61.55 & 62.34 & 61.55 & 35.34 & 50.00 & 41.41 & 50.00 \\
        & Borderline SMOTE & 69.50 & 64.46 & 65.63 & 64.46 & 14.66 & 50.00 & 22.67 & 50.00 \\
& RE-SMOTEBoost & \textbf{78.68} & \textbf{68.15} & \textbf{70.26} & \textbf{68.15} & \textbf{35.34} & \textbf{50.00} & \textbf{41.41} & \textbf{50.00} \\

\midrule
\end{tabular}

}
\end{table*}

Table \ref{tab7} presents the comparative experimental results of the proposed method with the RF classifier. The last rows of the table display the average values obtained by each method across nine datasets. The experimental results indicate that RF-RE-SMOTEBoost achieved competitive performance, obtaining the highest mean accuracy of (80.42\%). Additionally, compared to other evaluation metrics, the proposed method outperformed other approaches across all relevant metrics. The LightGBM method ranked second in terms of accuracy, recall, and F-score, with the lowest corresponding values of (78.07\%), (67.16\%), and (67.23\%), respectively, except for the G-Mean metric, where the EasyEnsemble classifier achieved a better average result of (69.50\%) than LightGBM. Compared to the last-ranked model (EasyEnsemble), RF-RE-SMOTEBoost achieves an improvement of  accuracy (10.65\%), recall(18.67\%), f-score(3.73\%)

\begin{table*}[h]
\renewcommand{\arraystretch}{1.2} 
\setlength{\tabcolsep}{4pt} 

\caption{Diagnostic testing efficiency of the proposed method vs. state-of-the-art ensemble methods using a random forest classifier.}
\label{tab7}

\centering
\resizebox{\textwidth}{!}{ 
\begin{tabular}{l c c c c c c c c c c c c c c}
\hline
\multirow{2}{*}{Cancer Dataset}  
& \multicolumn{4}{c}{LightGBM} 
& \multicolumn{4}{c}{EasyEnsembleClassifer} 
& \multicolumn{4}{c}{RF-RE-SMOTEBoost} \\

\cmidrule(lr){2-5} \cmidrule(lr){6-9} \cmidrule(lr){10-13}  
 & Accuracy($\%$) & Recall(\%)&  $F_1$-score($\%$) &  G-means($\%$)  &  Accuracy($\%$) & Recall(\%)&  $F_1$-score($\%$) &  G-means($\%$)  &  Accuracy($\%$) & Recall(\%)&  $F_1$-score($\%$) &  G-means($\%$) \\  
\midrule
        WOBC & 95.00 & 94.70 & 94.48 & 94.27 & 96.43 & 96.29 & 96.06 & 96.29 & 96.43 & 96.78 & 96.09 & 96.78 \\
        WDBC & 95.61 & 94.05 & 95.16 & 94.05 & 98.25 & 97.62 & 98.10 & 97.62 & 97.36 & 96.42 & 97.12 & 96.42 \\
        
        Mammographic & 83.42 & 83.72 & 83.41 & 83.72 & 84.46 & 84.52 & 84.41 & 84.52 & 84.97 & 85.24 & 84.95 & 85.24 \\
        
        Breast Cancer Coimbra & 66.67 & 67.13 & 66.67 & 67.13 & 70.83 & 70.98 & 70.78 & 70.98 & 79.17 & 79.37 & 79.13 & 79.37 \\
        
        Breast Cancer & 75.86 & 65.71 & 67.05 & 65.71 & 60.34 & 61.62 & 58.25 & 61.62 & 75.86 & 65.71 & 67.05 & 65.71 \\
        
        Real Breast Cancer Data & 79.10 & 54.63 & 55.08 & 54.63 & 47.76 & 54.14 & 43.70 & 54.14 & 77.55 & 57.31 & 57.57 & 57.31 \\
        
        Haberman's Survival Breast & 62.90 & 46.47 & 45.55 & 46.47 & 59.68 & 54.48 & 53.24 & 54.48 & 62.90 & 46.47 & 45.55 & 46.47 \\
        
        Breast Cancer Survival (Preprocessed) & 95.04 & 49.15 & 48.73 & 49.15 & 58.68 & 54.48 & 40.40 & 54.48 & 96.27 & 49.84 & 49.13 & 49.84 \\
        
        21239 Breast Cancer Survival & 49.00 & 48.90 & 48.90 & 48.90 & 51.50 & 51.37 & 51.35 & 51.37 & 53.25 & 53.24 & 53.23 & 53.24 \\
        \midrule
        Average results & 78.07 & 67.16 & 67.23 & 67.11 & 69.77 & 51.37 & 66.25 & 69.50 &\textbf{ 80.42} & \textbf{70.04 }& \textbf{69.98 }& \textbf{70.04} \\
        \bottomrule
\end{tabular}
}
\end{table*}

Tables \ref{tab8} and \ref{tab9} provide the comparative results of the proposed RF-RE-SMOTEBoost method based on random forest, evaluated against the boosting approach (AdaBoost classifier) using different base classifiers, including DT, NB, KNN, and RF. The average result is calculated for each evaluation metric among the nine datasets. Table \ref{tab8} shows that RF-RE-SMOTEBoost achieved the highest average result for all metrics, with corresponding values of precision (70.42\%), recall (70.04\%), F-score (69.98\%), and G-means (70.04\%), while the NB-Boosting classifier achieved the lowest mean value among other methods, with corresponding values of precision (54.37\%), recall (57.60\%), F-score (51.66\%), and G-means (57.60\%). From Table \ref{tab9}, it can be seen that the KNN-Boosting classifier achieved the lowest average result for all relevant metrics with corresponding values of accuracy (58.24\%), recall (59.38\%), G-means (56.67\%), and F-score (59.38\%), while RF-RE-SMOTEBoost achieved the highest result.

\begin{table*}[h]
\renewcommand{\arraystretch}{1.2} 
\setlength{\tabcolsep}{5pt} 

\caption{Diagnostic testing efficiency of the proposed method vs. state-of-the-art boosting methods using a random forest classifier.}
\label{tab8}

\centering
\resizebox{\textwidth}{!}{ 
\begin{tabular}{l c c c c c c c c c c c c c c}
\hline
\multirow{2}{*}{Cancer Dataset}  
& \multicolumn{4}{c}{DT-Boosting classifer} 
& \multicolumn{4}{c}{NB-Boosting classifer} 
& \multicolumn{4}{c}{RF-RE-SMOTEBoost} \\  

\cmidrule(lr){2-5} \cmidrule(lr){6-9} \cmidrule(lr){10-13}  
 & Precision($\%$) & Recall(\%)&  $F_1$-score($\%$) &  G-means($\%$)  & Precision($\%$) & Recall(\%)&  $F_1$-score($\%$) &  G-means($\%$)  & Precision($\%$) & Recall(\%)&  $F_1$-score($\%$) &  G-means($\%$) \\    
\midrule
        WOBC & 94.08 & 93.16 & 93.59 & 93.16 & 95.15 & 93.70 & 94.36 & 93.70 & 95.52 & 96.78 & 96.09 & 96.78 \\
        WDBC & 96.05 & 94.54 & 95.21 & 94.54 & 68.58 & 50.69 & 28.47 & 50.69 & 97.99 & 96.42 & 97.12 & 96.42 \\
        Mammographic & 84.87 & 85.00 & 84.92 & 85.00 & 76.33 & 75.62 & 75.76 & 75.62 & 85.04 & 85.24 & 84.95 & 85.24 \\
        Breast Cancer Coimbra & 66.43 & 66.43 & 66.43 & 66.43 & 27.08 & 50.00 & 35.14 & 50.00 & 79.16 & 79.37 & 79.13 & 79.37 \\
        Breast Cancer & 65.81 & 63.27 & 64.03 & 63.27 & 35.09 & 48.78 & 40.82 & 48.78 & 71.18 & 65.71 & 67.05 & 65.71 \\
        Real Breast Cancer Data & 41.67 & 49.11 & 45.08 & 49.11 & 41.79 & 50.00 & 45.53 & 50.00 & 57.90 & 57.31 & 57.57 & 57.31 \\
        Haberman's Survival Breast & 41.40 & 44.43 & 42.38 & 44.43 & 44.94 & 47.69 & 44.74 & 47.69 & 45.36 & 46.47 & 45.55 & 46.47 \\
        Breast Cancer Survival-preproceed & 48.34 & 49.79 & 49.05 & 49.79 & 48.35 & 50.00 & 49.16 & 50.00 & 48.44 & 49.84 & 49.13 & 49.84 \\
        21239 Breast Cancer Survival & 54.85 & 54.49 & 53.33 & 54.49 & 52.05 & 51.93 & 50.96 & 51.93 & 53.23 & 53.24 & 53.23 & 53.24 \\
        \midrule
        Average results & 65.94 & 66.69 & 66.00 & 66.69 & 54.37 & 57.60 & 51.66 & 57.60 &\textbf{ 70.42} & \textbf{70.04} &\textbf{ 69.98} & \textbf{70.04} \\
        \bottomrule
\end{tabular}
}
\end{table*}

\begin{table*}[h]
\renewcommand{\arraystretch}{1.2} 
\setlength{\tabcolsep}{5pt} 

\caption{Diagnostic testing efficiency of the proposed method vs. state-of-the-art boosting methods using a random forest classifier.}
\label{tab9}

\centering
\resizebox{\textwidth}{!}{ 
\begin{tabular}{l c c c c c c c c c c c c c c}
\hline
\multirow{2}{*}{Cancer Dataset}  
& \multicolumn{4}{c}{KNN-Boosting classifer} 
& \multicolumn{4}{c}{RF-Boosting classifer} 
& \multicolumn{4}{c}{RF-RE-SMOTEBoost} \\  

\cmidrule(lr){2-5} \cmidrule(lr){6-9} \cmidrule(lr){10-13}  
& Precision($\%$) & Recall(\%)&  $F_1$-score($\%$) &  G-means($\%$)  & Precision($\%$) & Recall(\%)&  $F_1$-score($\%$) &  G-means($\%$)  & Precision($\%$) & Recall(\%)&  $F_1$-score($\%$) &  G-means($\%$) \\  
\midrule
        WOBC & 94.66 & 94.20 & 94.42 & 94.20 & 95.84 & 96.29 & 96.06 & 96.29 & 95.52 & 96.78 & 96.09 & 96.78 \\
        WDBC & 75.98 & 60.52 & 58.75 & 60.52 & 97.99 & 96.42 & 97.12 & 96.42 & 97.99 & 96.42 & 97.12 & 96.42 \\
        Mammographic & 81.85 & 82.04 & 81.83 & 82.04 & 82.19 & 82.28 & 81.86 & 82.28 & 85.04 & 85.24 & 84.95 & 85.24 \\
        Breast Cancer Coimbra & 46.32 & 47.55 & 43.75 & 47.55 & 79.16 & 79.37 & 79.13 & 79.37 & 79.16 & 79.37 & 79.13 & 79.37 \\
        Breast Cancer & 35.34 & 50.00 & 41.41 & 50.00 & 65.81 & 63.27 & 64.03 & 63.27 & 71.18 & 65.71 & 67.05 & 65.71 \\
        Real Breast Cancer Data & 41.79 & 50.00 & 45.53 & 50.00 & 41.79 & 50.00 & 45.53 & 50.00 & 57.90 & 57.31 & 57.57 & 57.31 \\
        Haberman's Survival Breast & 49.57 & 49.86 & 46.35 & 49.86 & 40.58 & 43.34 & 41.60 & 43.34 & 45.36 & 46.47 & 45.55 & 46.47 \\
        Breast Cancer Survival-preproceed & 48.35 & 50.00 & 49.16 & 50.00 & 48.35 & 50.00 & 49.16 & 50.00 & 48.44 & 49.84 & 49.13 & 49.84 \\
        21239 Breast Cancer Survival & 50.30 & 50.27 & 48.85 & 50.28 & 54.03 & 54.03 & 53.10 & 54.03 & 53.23 & 53.24 & 53.23 & 53.24 \\
        \midrule
        Average Results & 58.24 & 59.38 & 56.67 & 59.38 & 67.30 & 68.33 & 67.51 & 68.33 &\textbf{ 70.42 }&\textbf{ 70.04 }& \textbf{69.98} & \textbf{70.04} \\
        \bottomrule
\end{tabular}
}
\end{table*}

Table \ref{tab10} illustrates the comparative results of the bagging approach with RF-RE-SMOTEBoost using different classifiers (e.g., KNN, DT). The last row shows the average result of each metric across nine datasets. The results show that RF-RE-SMOTEBoost achieved the optimal average results for all relevant metrics with corresponding values: avg- precision (70.42\%), avg-recall (70.04\%), avg-f-score (69.98\%), and avg-g-means (70.04\%). DT-Bagging ranked second in terms of avg-accuracy (66.1\%), avg-f-score (65.2\%), and avg-g-means (62.49\%). It can be concluded from the average results that RF-RE-SMOTEBoost is higher than DT-Bagging by precision (4.32\%), f-score (4.78\%), and g-means (7.55\%).

\begin{table*}[h]
\renewcommand{\arraystretch}{1.2} 
\setlength{\tabcolsep}{5pt} 

\caption{Diagnostic testing efficiency of the proposed method vs. state-of-the-art bagging methods using a random forest classifier.}
\label{tab10}

\centering
\resizebox{\textwidth}{!}{ 
\begin{tabular}{l c c c c c c c c c c c c c c}
\hline
\multirow{2}{*}{Cancer Dataset}  
& \multicolumn{4}{c}{KNN-Bagging classifer} 
& \multicolumn{4}{c}{DT-Bagging classifer} 
& \multicolumn{4}{c}{RF-RE-SMOTEBoost} \\  

\cmidrule(lr){2-5} \cmidrule(lr){6-9} \cmidrule(lr){10-13}  
& Precision($\%$) & Recall(\%)&  $F_1$-score($\%$) &  G-means($\%$)  & Precision($\%$) & Recall(\%)&  $F_1$-score($\%$) &  G-means($\%$)  & Precision($\%$) & Recall(\%)&  $F_1$-score($\%$) &  G-means($\%$) \\ 
 \midrule
        WOBC & 94.66 & 94.20 & 94.42 & 94.20 & 89.15 & 91.94 & 90.10 & 91.94 & 95.52 & 96.78 & 96.09 & 96.78 \\
        WDBC & 72.11 & 63.19 & 63.07 & 63.19 & 92.31 & 87.40 & 89.07 & 87.40 & 97.99 & 96.42 & 97.12 & 96.42 \\
        Mammographic & 80.43 & 80.60 & 80.30 & 80.60 & 83.38 & 82.35 & 82.59 & 82.35 & 85.04 & 85.24 & 84.95 & 85.24 \\
        Breast Cancer Coimbra & 62.14 & 61.89 & 61.90 & 61.89 & 62.50 & 62.59 & 62.43 & 62.59 & 79.16 & 79.37 & 79.13 & 79.37 \\
        Breast Cancer & 86.61 & 55.88 & 52.79 & 55.88 & 78.68 & 68.15 & 70.26 & 68.15 & 71.18 & 65.71 & 67.05 & 65.71 \\
        Real Breast Cancer Data & 41.79 & 50.00 & 45.53 & 50.00 & 41.79 & 50.00 & 45.53 & 50.00 & 57.90 & 57.31 & 57.57 & 57.31 \\
        Haberman's Survival Breast & 42.36 & 45.52 & 43.17 & 45.52 & 47.90 & 48.64 & 47.39 & 48.64 & 45.36 & 46.47 & 45.55 & 46.47 \\
        Breast Cancer Survival-preproceed & 48.35 & 50.00 & 49.16 & 50.00 & 48.35 & 50.00 & 49.15 & 50.00 & 48.44 & 49.84 & 49.13 & 49.84 \\
        21239 Breast Cancer Survival & 53.44 & 53.41 & 53.21 & 53.41 & 50.83 & 50.81 & 50.26 & 50.81 & 53.23 & 53.24 & 53.23 & 53.24 \\
        \midrule
        Average results & 64.65 & 61.63 & 60.39 & 61.63 & 66.10 & 50.81 & 65.20 & 62.49 &\textbf{ 70.42} &\textbf{ 70.04} &\textbf{ 69.98} & \textbf{70.04} \\
        \bottomrule
\end{tabular}
}
\end{table*}

Table \ref{tab11} reveals the comparative result of DT-RE-SMOTEBoost across different boosting approaches using a DT classifier. Based on the average result in the last row, DT-RE-SMOTEBoost achieved the highest average values across nine datasets for relevant metrics. DT-CUTBoost achieved the second lowest values with corresponding average recall (58.21\%), average F-score (47.78\%), and average G-means (58.21\%), except in average accuracy, where DT-MPSUBoost achieved the lowest value (55.57\%). Based on these results, DT-RE-SMOTEBoost achieved an improvement of average recall (8.84\%), average F-score (14.97\%), and average G-means (8.84\%), compared to the DT-CUTBoost method and a precision improvement of (10.57\%) compared to DT-MPSUBoost.

\begin{table*}[h]
\renewcommand{\arraystretch}{1.2} 
\setlength{\tabcolsep}{7pt} 

\caption{Diagnostic testing efficiency of the proposed method vs. state-of-the-art boosting methods using a random forest classifier.}
\label{tab11}

\centering
\resizebox{\textwidth}{!}{ 
\begin{tabular}{l c c c c c c c c c c c c c c c c c c}
\hline
\multirow{2}{*}{Cancer Dataset}  
& \multicolumn{4}{c}{DT-CUTBoost} 
& \multicolumn{4}{c}{DT-MPSUBoost} 
& \multicolumn{4}{c}{DT-Stack-Adaoost} 
& \multicolumn{4}{c}{RF-RE-SMOTEBoost} \\

        \cmidrule(lr){2-5} \cmidrule(lr){6-9} \cmidrule(lr){10-13} \cmidrule(lr){14-17}
         &  Prec.($\%$) & Rec.(\%) & $F_1$-score($\%$)&  G-means($\%$) &   Prec.($\%$) & Rec.(\%) & $F_1$-score($\%$)&  G-means($\%$) &   Prec.($\%$) &  Rec.(\%) & $F_1$-score($\%$)&  G-means($\%$) &   Prec.($\%$) & Rec.(\%) & $F_1$-score($\%$)&  G-means($\%$) \\
        \hline
        WOBC & 89.15 & 91.94 & 90.10 & 91.94 & 85.49 & 89.18 & 85.90 & 89.18 & 93.06 & 92.62 & 92.83 & 92.62 & 93.05 & 94.59 & 93.72 & 94.59 \\
        WDBC & 88.95 & 80.26 & 82.34 & 80.26 & 89.28 & 91.37 & 89.94 & 91.37 & 92.31 & 87.40 & 89.07 & 87.40 & 89.25 & 90.87 & 89.86 & 90.87 \\
        Mammographic & 53.14 & 50.32 & 34.16 & 50.32 & 23.06 & 50.00 & 31.56 & 50.00 & 83.16 & 81.71 & 81.98 & 81.71 & 83.61 & 83.49 & 82.90 & 83.49 \\
        Breast Cancer Coimbra & 22.92 & 50.00 & 31.42 & 50.00 & 62.50 & 62.59 & 62.43 & 62.59 & 50.35 & 50.35 & 50.00 & 50.35 & 62.50 & 62.59 & 62.43 & 62.59 \\
        Breast Cancer & 35.34 & 50.00 & 41.41 & 50.00 & 57.21 & 56.67 & 56.85 & 56.67 & 35.34 & 50.00 & 41.41 & 50.00 & 69.03 & 69.66 & 69.31 & 69.66 \\
        Real Breast Cancer Data & 51.19 & 50.81 & 23.61 & 50.81 & 49.86 & 49.92 & 21.95 & 49.92 & 41.79 & 50.00 & 45.53 & 50.00 & 55.43 & 56.20 & 33.03 & 56.20 \\
        Haberman's Survival & 37.10 & 50.00 & 42.59 & 50.00 & 56.14 & 57.61 & 55.71 & 57.61 & 41.40 & 44.43 & 42.38 & 44.43 & 45.37 & 46.47 & 45.55 & 46.47 \\
        Breast Cancer Survival-prep. & 48.35 & 50.00 & 49.16 & 50.00 & 50.74 & 55.56 & 41.42 & 55.56 & 48.35 & 50.00 & 49.16 & 50.00 & 48.33 & 49.36 & 48.84 & 49.36 \\
        21239 Breast Cancer Survival & 76.01 & 50.52 & 35.24 & 50.52 & 25.88 & 50.00 & 34.10 & 50.00 & 57.23 & 52.43 & 42.21 & 52.43 & 50.65 & 50.20 & 39.13 & 50.20 \\
       \midrule
        Average results & 55.79 & 58.21 & 47.78 & 58.21 & 55.57 & 62.54 & 53.32 & 62.54 & 60.33 & 62.10 & 59.40 & 62.10 & \textbf{ 66.36} &\textbf{ 67.05} & \textbf{ 62.75} &  \textbf{67.05} \\
        \hline
\end{tabular}
}
\end{table*}

\begin{figure*}
    \centering
    
    \subfloat[\centering WOBC dataset]{{\includegraphics[width=5.9cm,height=7cm]{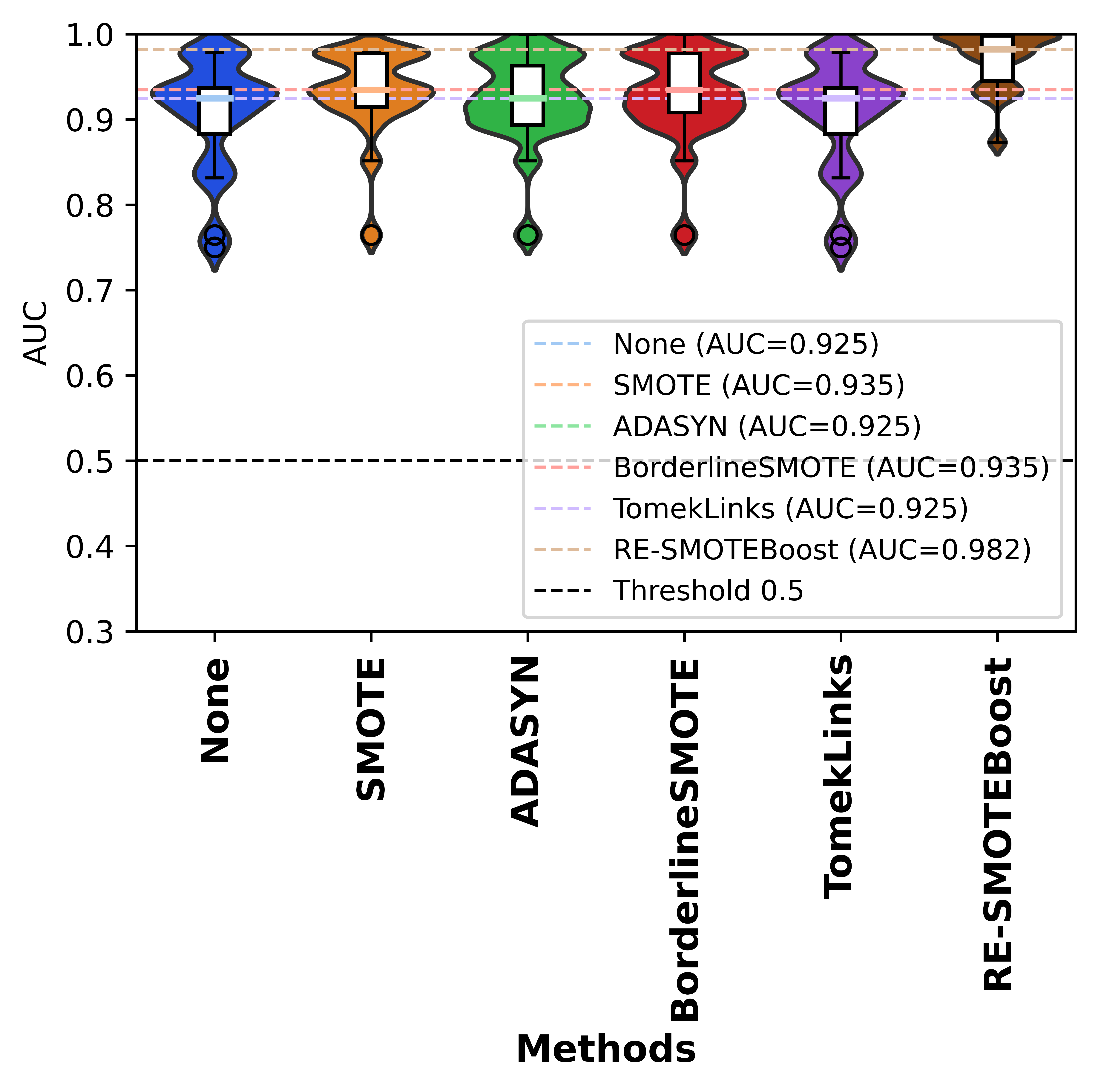} }}
    \subfloat[\centering WDBC dataset]{{\includegraphics[width=5.9cm,height=7cm]{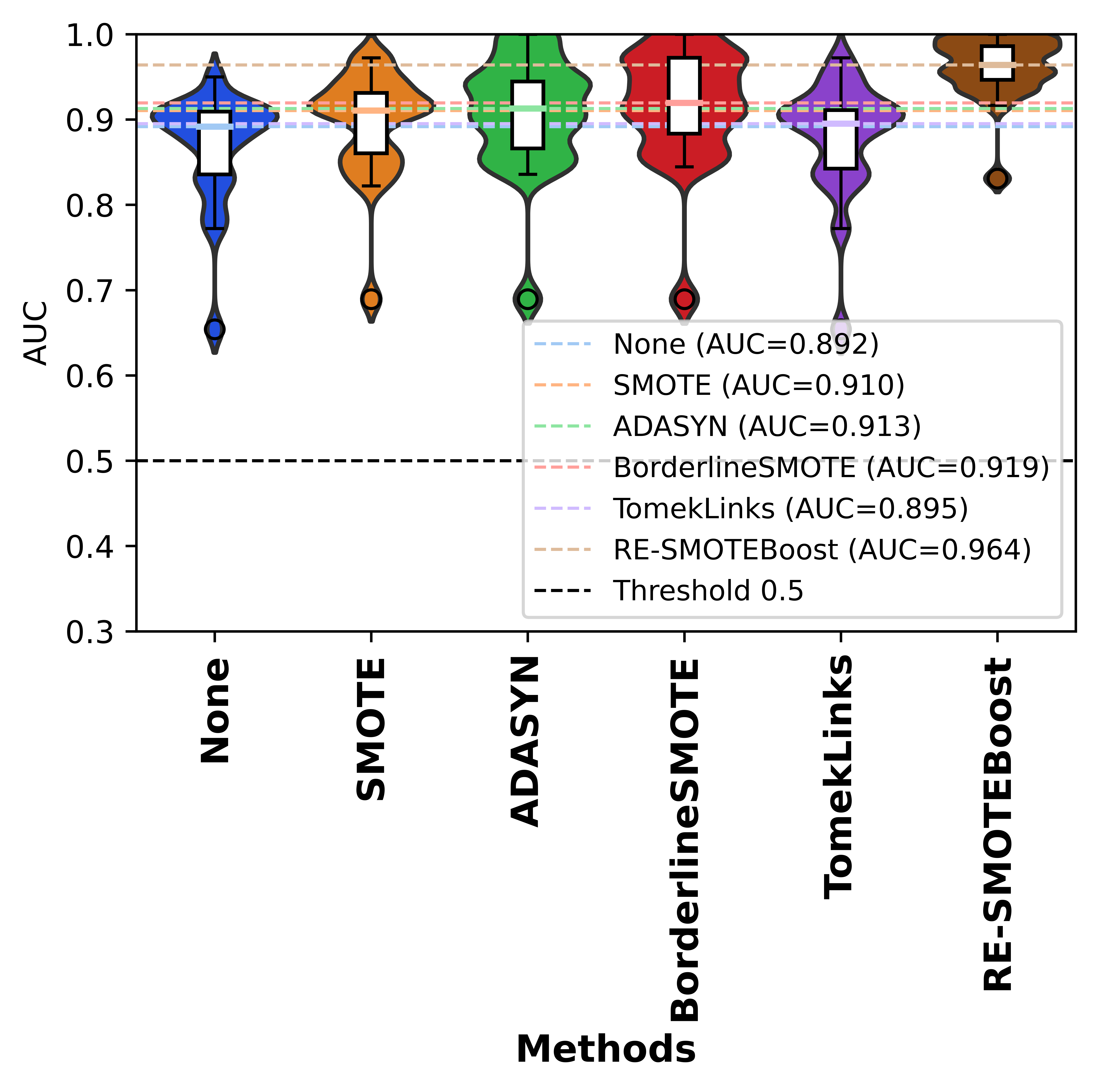} }}
    \subfloat[\centering Mammographic Mass]{{\includegraphics[width=5.9cm,height=7cm]{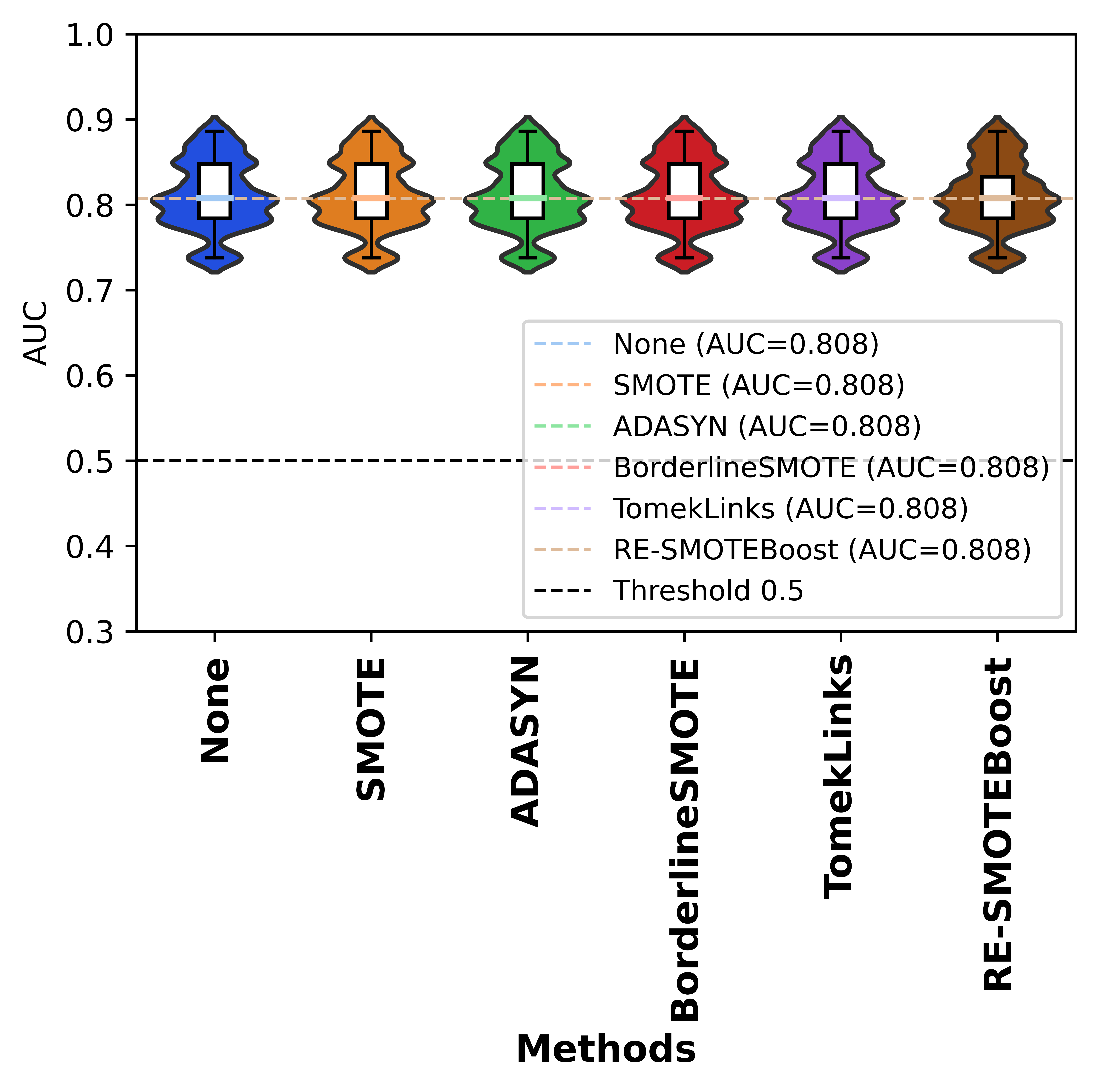} }}
    
    \caption{comparative analysis of  AUC measurement across different Data-level methods using 20CV.}
    \label{fig:fig6}
\end{figure*}

\begin{figure*}
    \centering
    
    \subfloat[\centering WOBC dataset]{{\includegraphics[width=5.9cm,height=7cm]{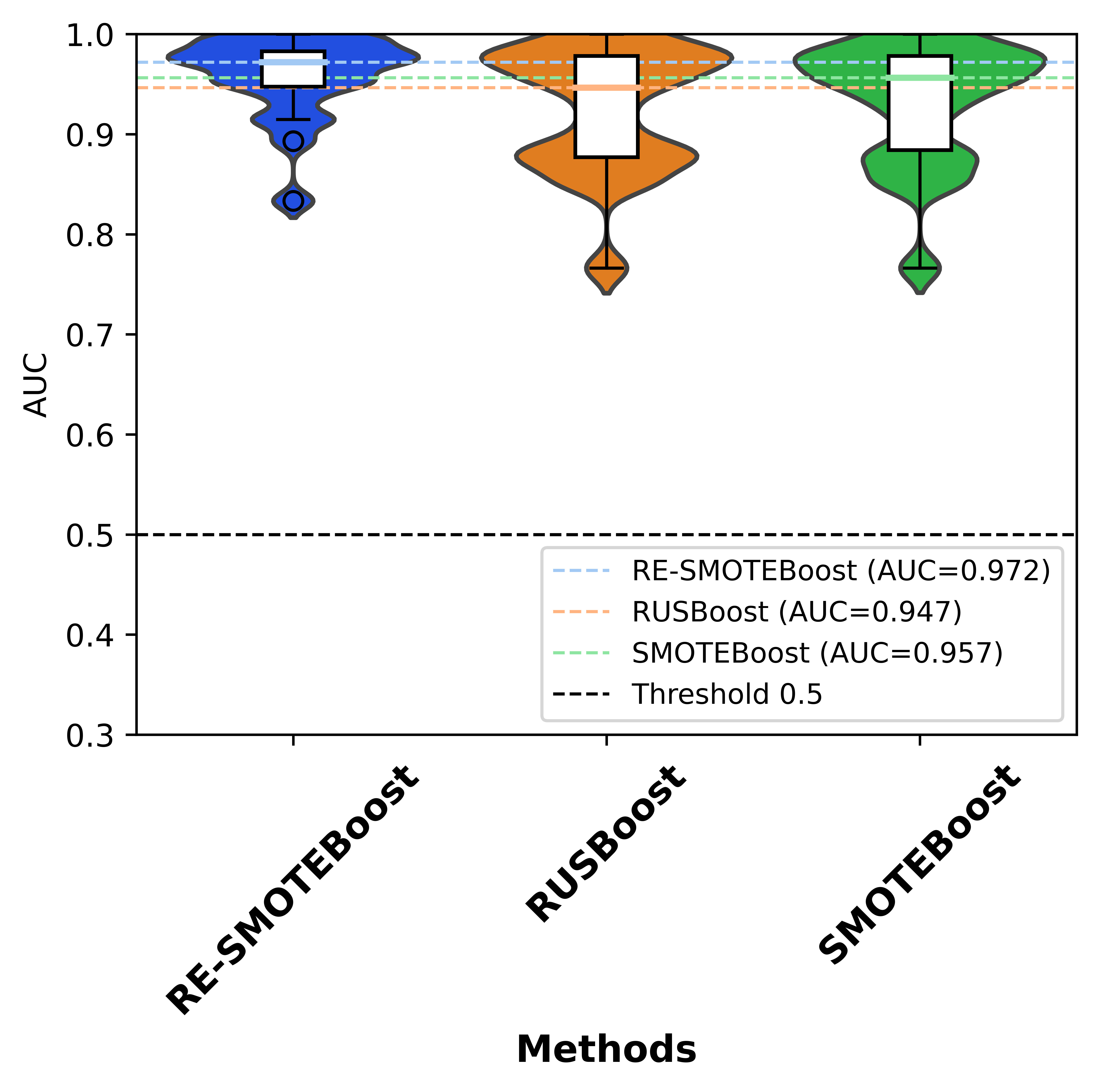} }}
    \subfloat[\centering WDBC dataset]{{\includegraphics[width=5.9cm,height=7cm]{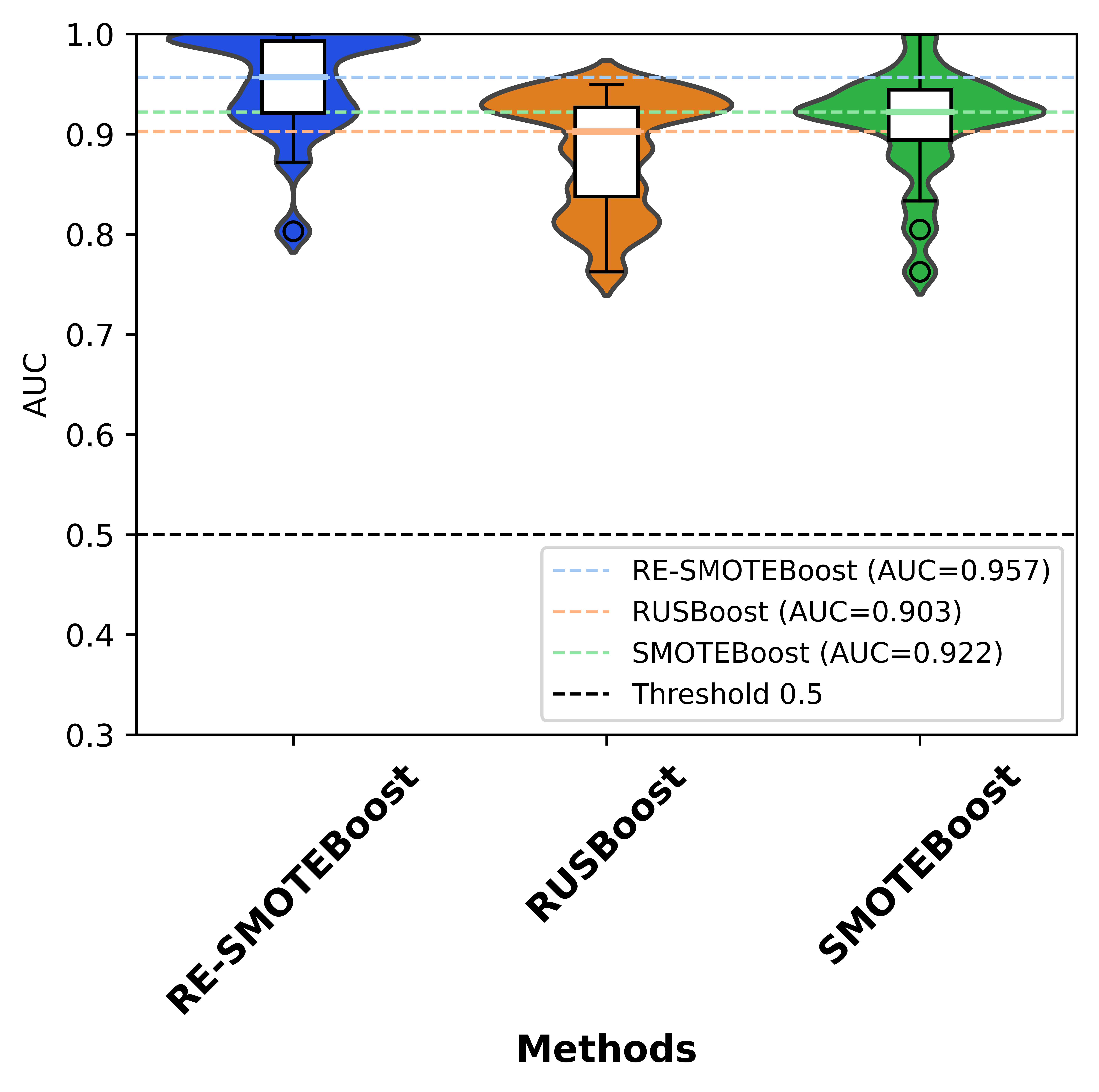} }}
    \subfloat[\centering Mammographic Mass]{{\includegraphics[width=5.9cm,height=7cm]{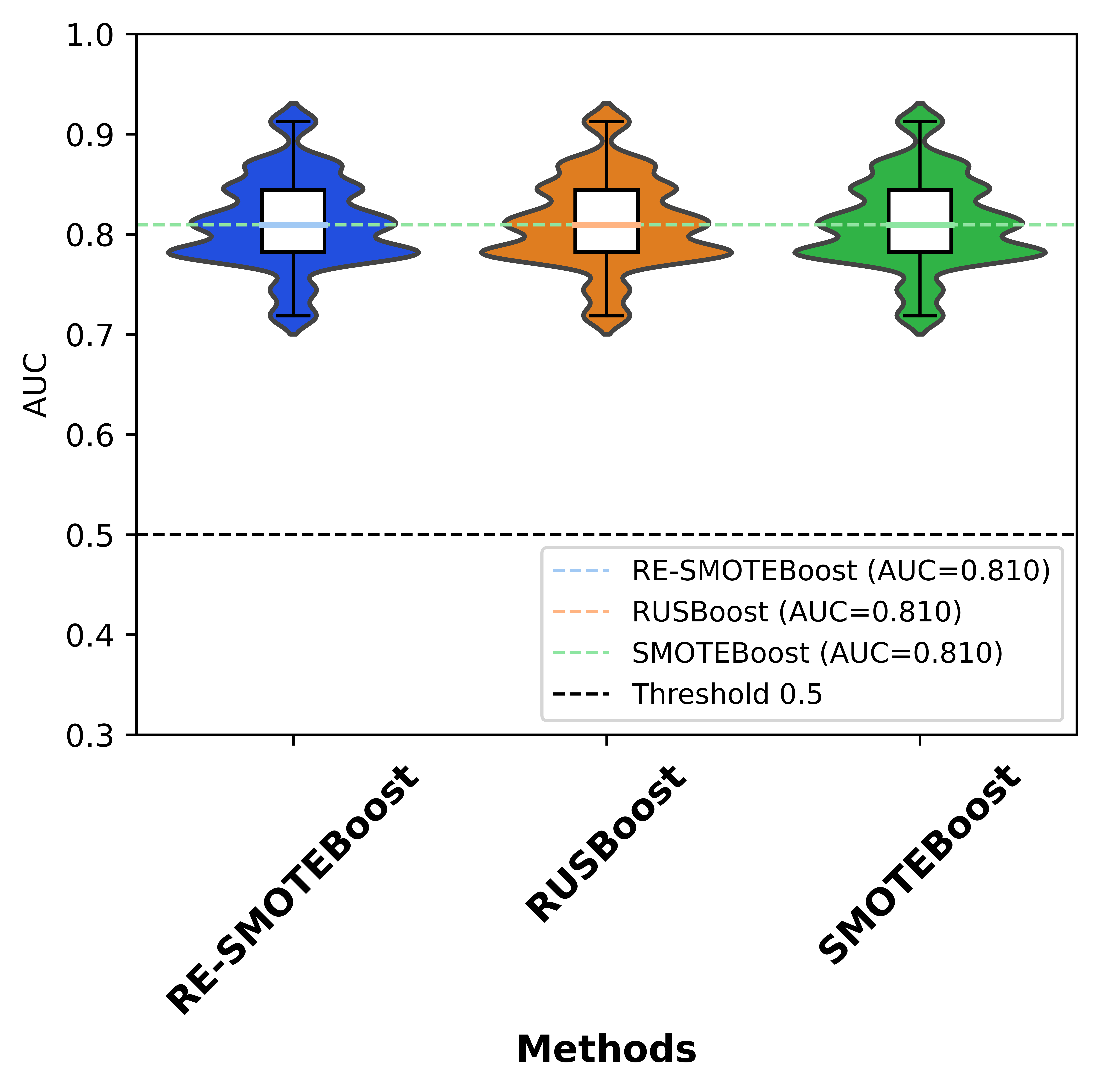} }}
    
    \caption{comparative analysis of  AUC measurement across different boosting methods using 20CV.}
    \label{fig:fig7}
\end{figure*}

Besides, we analyzed and compared the model performance on data processed with different oversampling techniques, employing a Decision Tree as the base classifier, using a violin plot, as shown in Fig. \ref{fig:fig6} and \ref{fig:fig7}. The violin plot is a statistical visualization based on probability density, used to display the dispersion (variance) of the data and prediction results. It includes five components: the minimum, first quartile (Q1 = 25\%), median, third quartile (Q3 = 75\%), and maximum \cite{Methododology85}\cite{Experimentation86}. Based on the box plots for the different models, the median line of the proposed method, DT-RE-SMOTEBoost, achieves the highest value and is positioned above the median lines of all other methods, demonstrating the superiority of the proposed method. Furthermore, the proposed method achieves the highest maximum value, outperforming other methods, while its minimum value remains relatively low. Observing the size of the box, it is worth noting that the proposed method's box is more compact and smaller than those of the other models, which supports our initial objective of this research to improve the reliability of the proposed model.

\begin{figure*}
    \centering
    
    \subfloat[\centering WOBC dataset]{{\includegraphics[width=5.9cm,height=6cm]{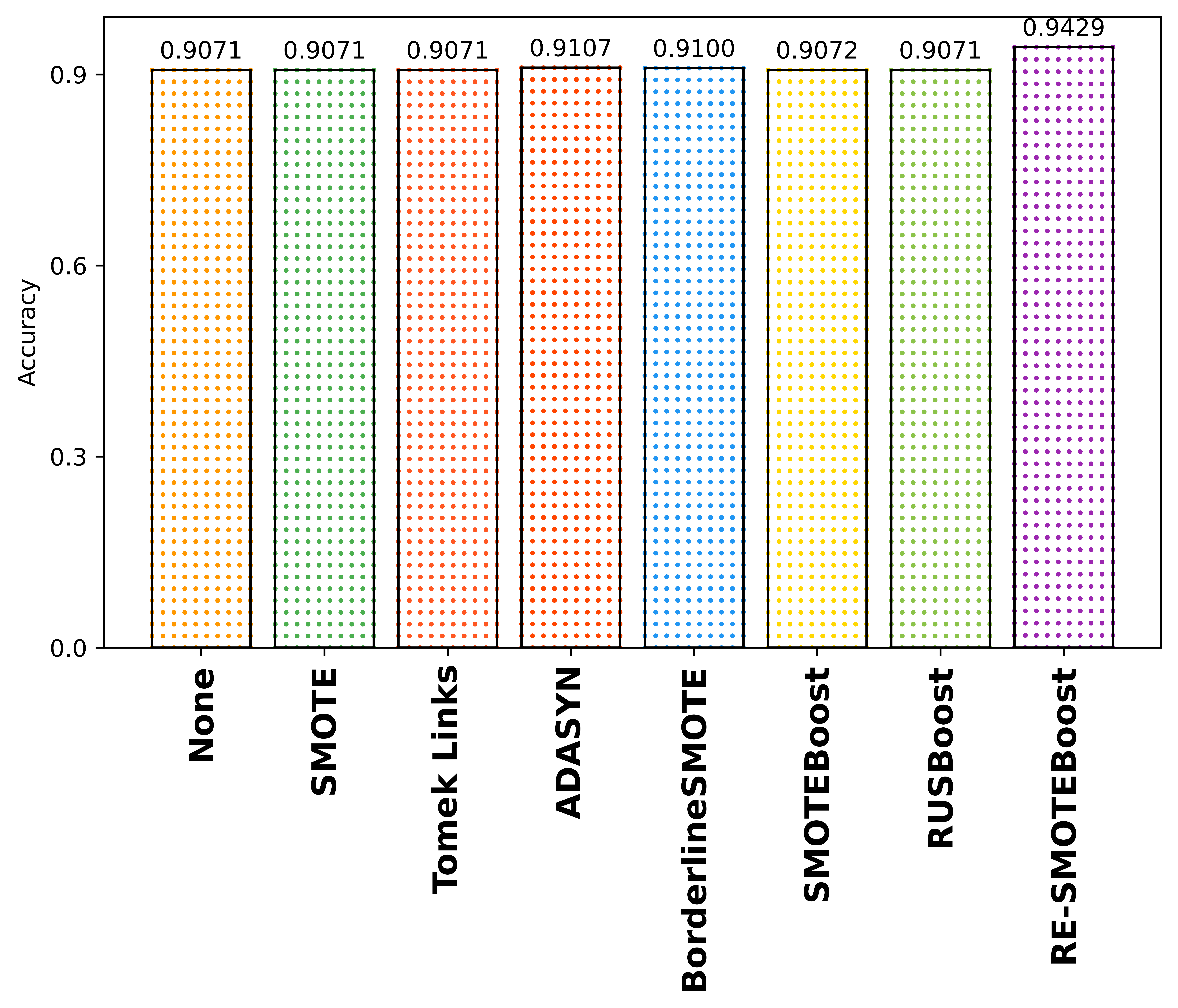} }}
    \subfloat[\centering WDBC dataset]{{\includegraphics[width=5.9cm,height=6cm]{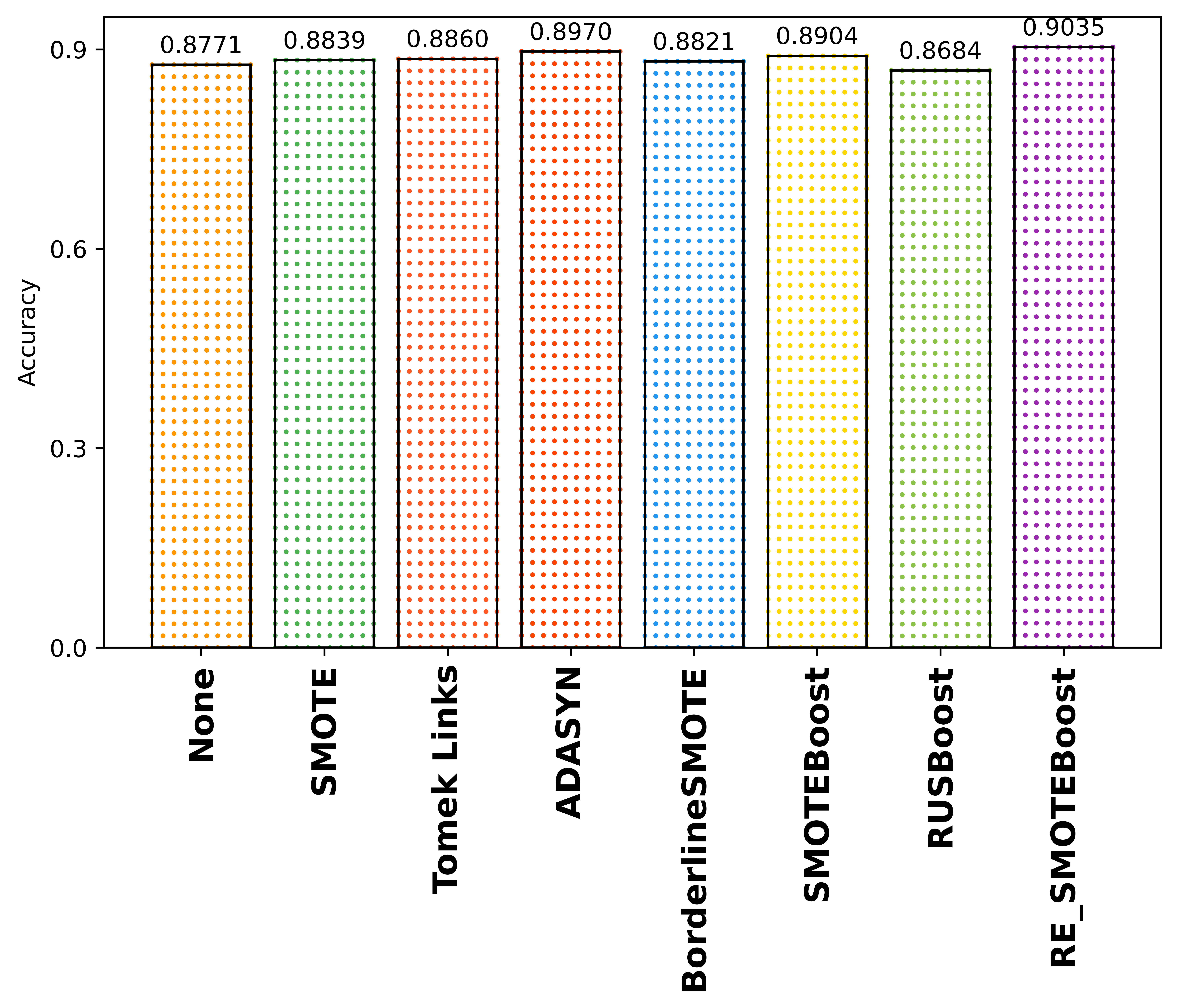} }}
    \subfloat[\centering  Mammographic Mass]{{\includegraphics[width=5.9cm,height=6cm]{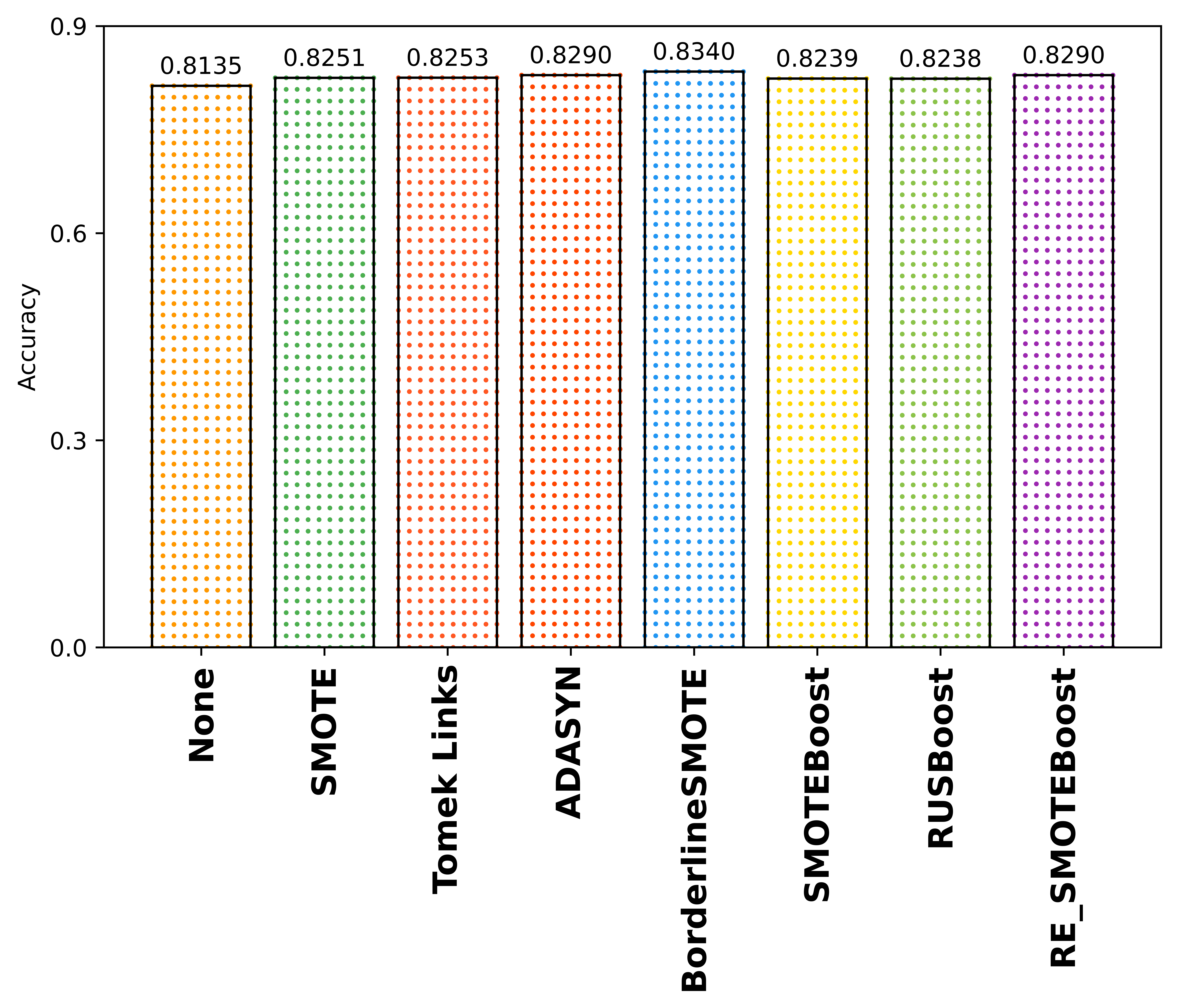} }}
    
    \caption{comparative analysis of accuracy  across different sampling methods using L=100 replications and DT classifer.}
    \label{fig:fig8}
\end{figure*}

\begin{figure*}
    \centering
    
    \subfloat[\centering WOBC dataset]{{\includegraphics[width=5.9cm,height=6cm]{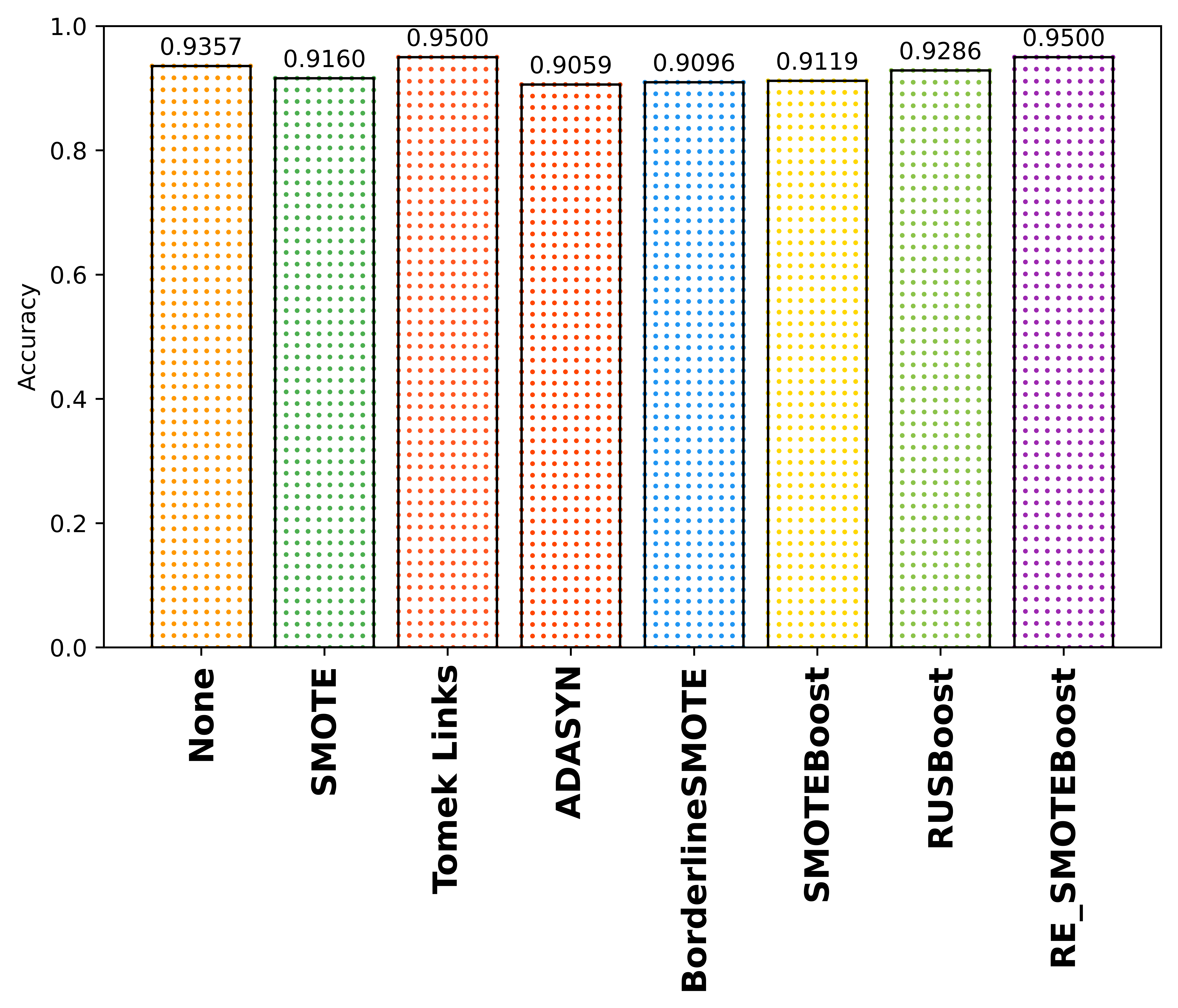} }}
    \subfloat[\centering WDBC dataset]{{\includegraphics[width=5.9cm,height=6cm]{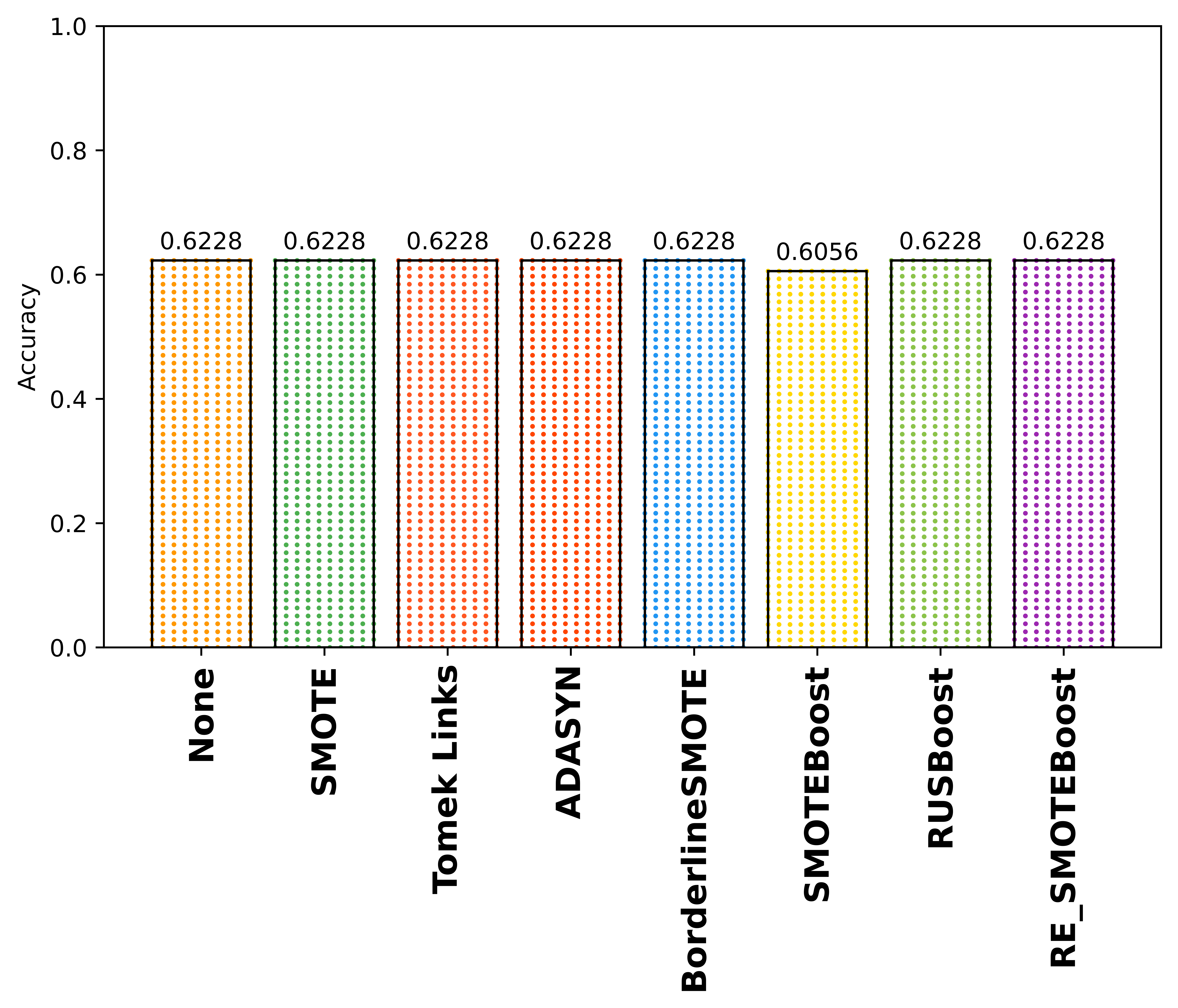} }}
    \subfloat[\centering  Mammographic Mass]{{\includegraphics[width=5.9cm,height=6cm]{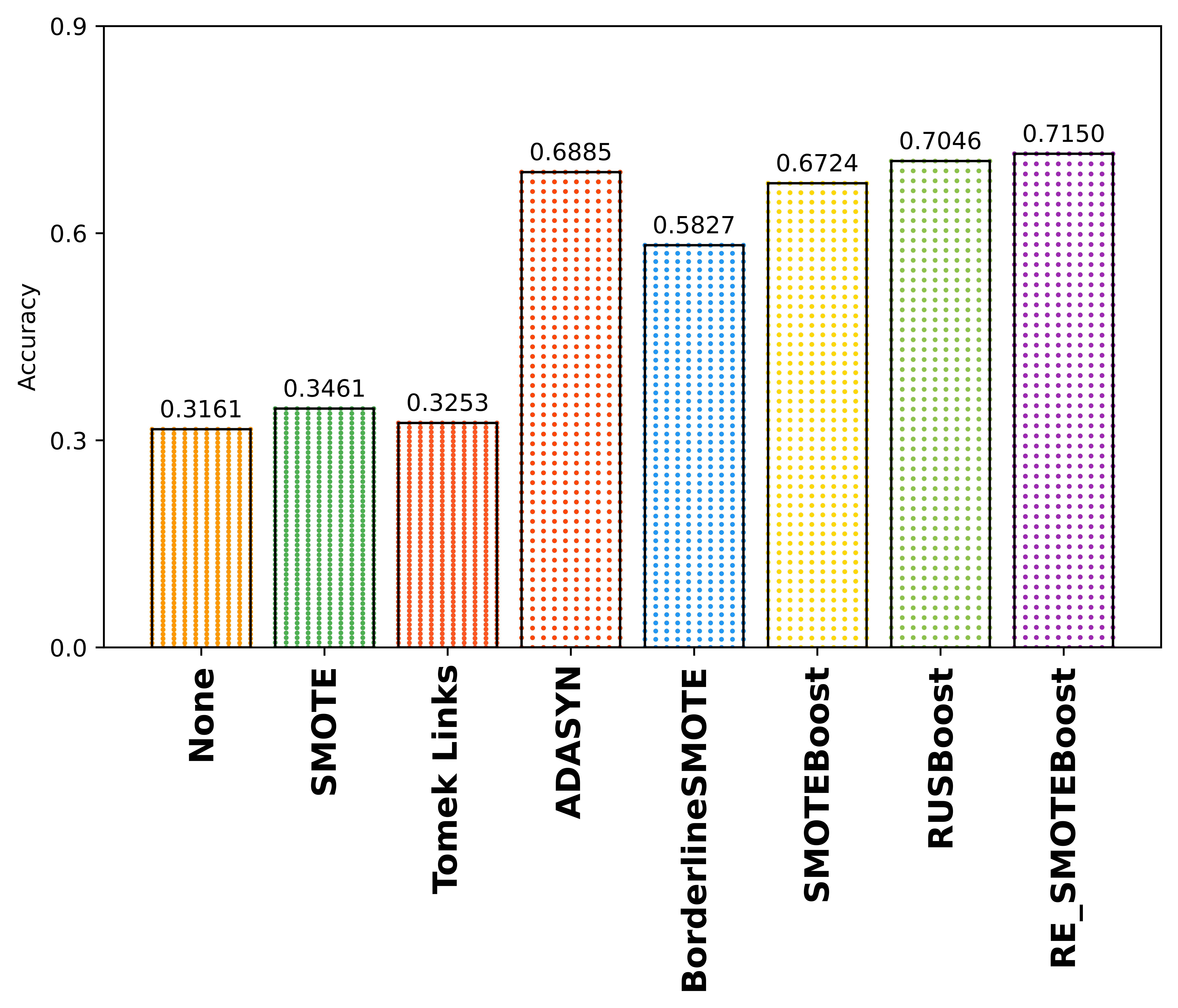} }}
    
    \caption{comparative analysis of accuracy across different sampling methods using L=100 replications and SVM classifier.}
    \label{fig:fig9}
\end{figure*}

\begin{figure*}
    \centering
    \includegraphics[width=17.7cm,height=7cm]{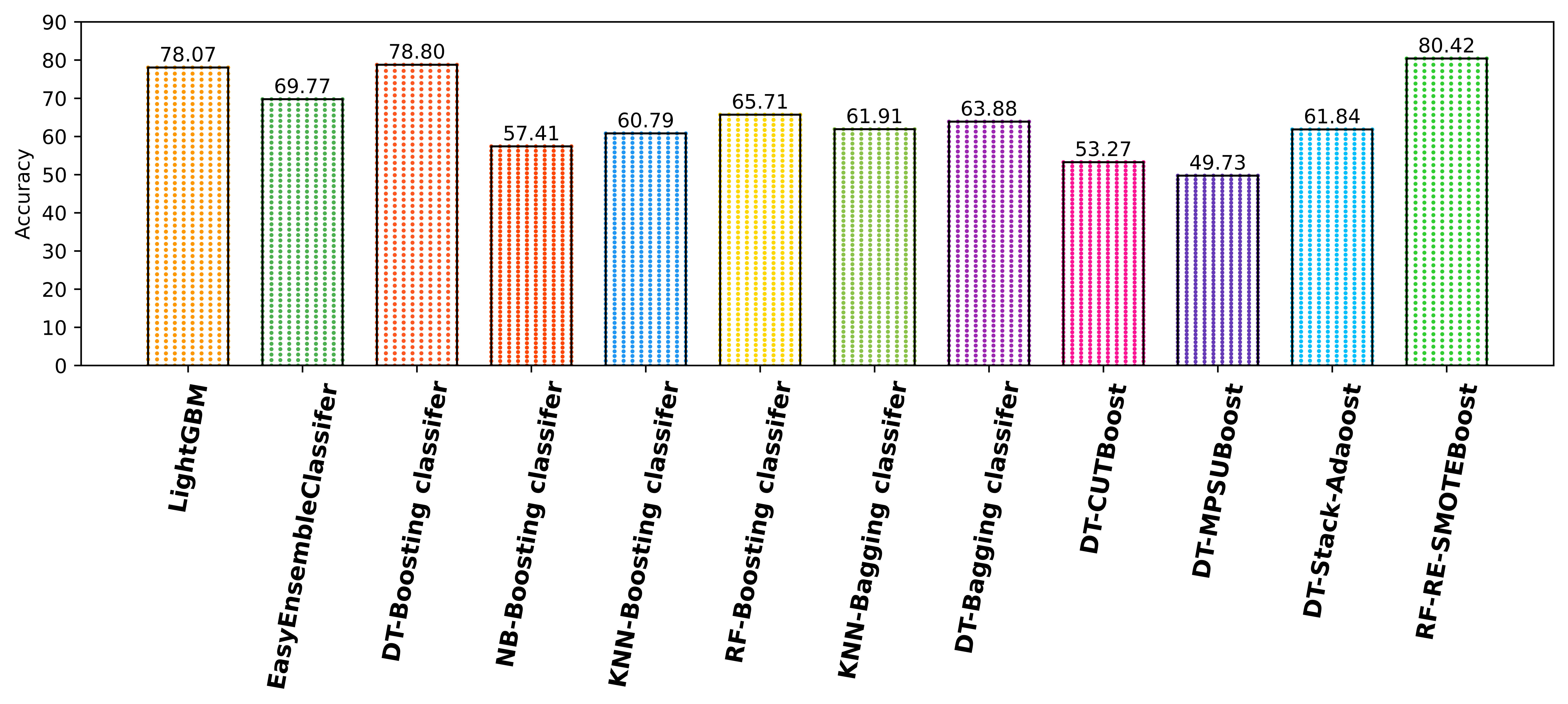} 
    
    \caption{comparative analysis of average  accuracy of nine datasets across all  methods using L=100 replications.}
    \label{fig}
\end{figure*}

Although accuracy is not a recommended metric in imbalanced classes \cite{RelatedWork24}, Fig. \ref{fig:fig8}, \ref{fig:fig9}, and  \ref{fig} provides evidence of the effectiveness of the proposed method, illustrating its superiority over other sampling methods for DT and SVM classifiers. Notably, in the worst-case scenario, the proposed method achieves similar accuracy to other methods and is at least as good as or better than theirs. For example, in a comparative study using SVM on the WDBC dataset, the majority of methods obtained a maximum accuracy of 0.6228\%. However, in one scenario with the DT classifier on the mammographic dataset, BorderSMOTE exhibits a very slight superiority, with a difference of 0.005\%.

\begin{figure*}
    \centering
    
    \subfloat[\centering WOBC dataset]{{\includegraphics[width=5.9cm,height=6cm]{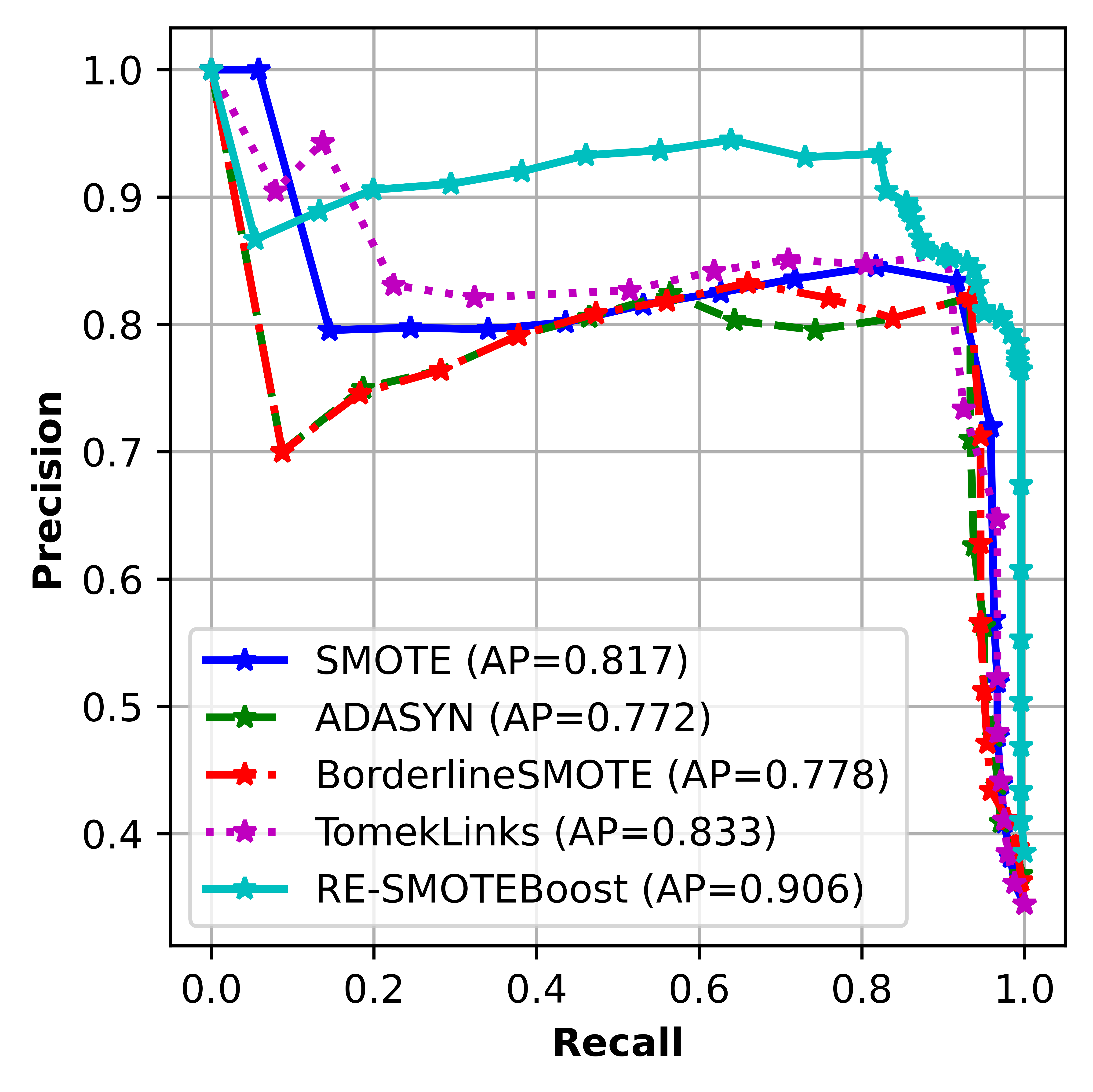} }}
    \subfloat[\centering WDBC dataset]{{\includegraphics[width=5.9cm,height=6cm]{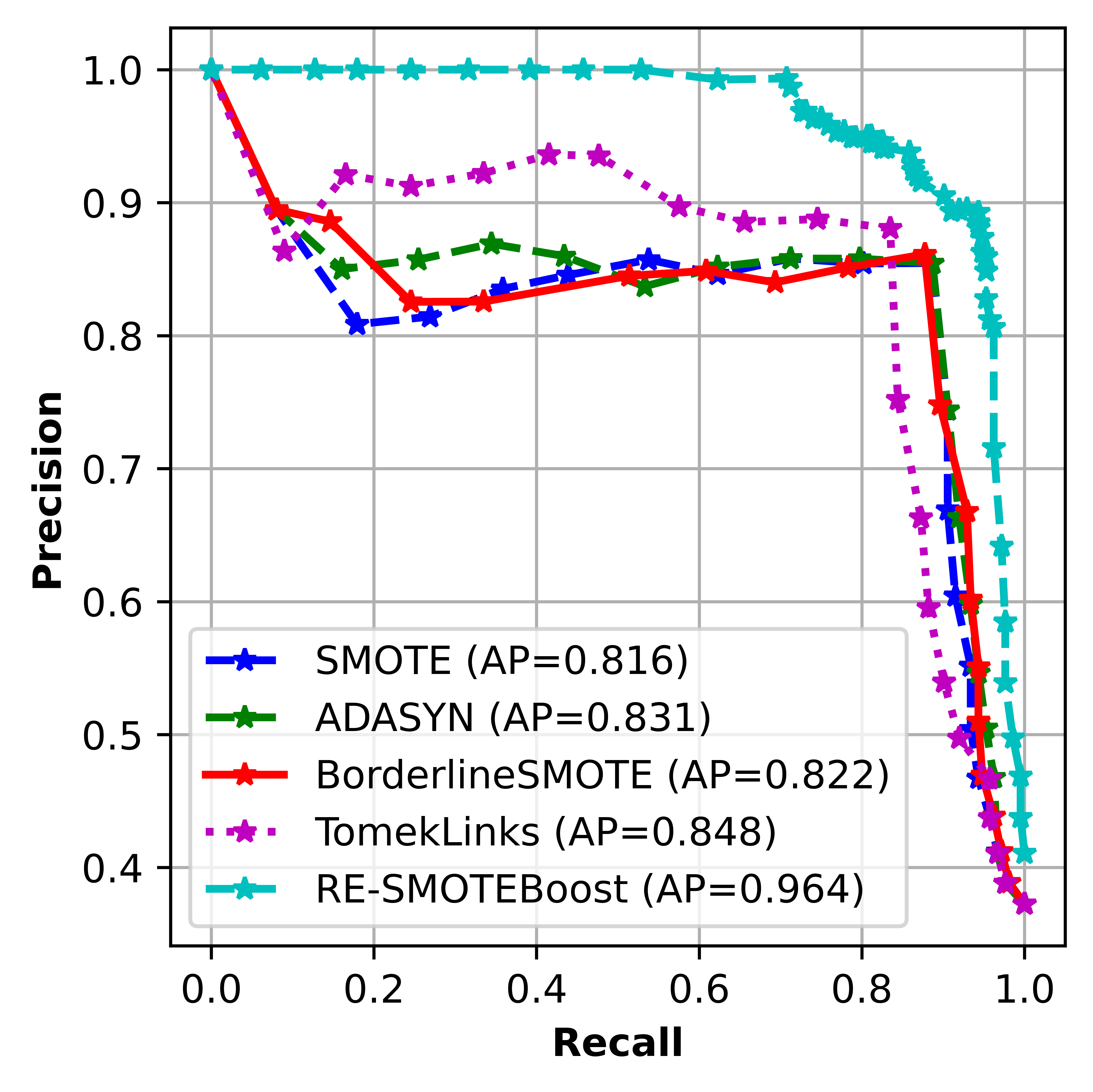} }}
    \subfloat[\centering Mammographic Mass]{{\includegraphics[width=5.9cm,height=6cm]{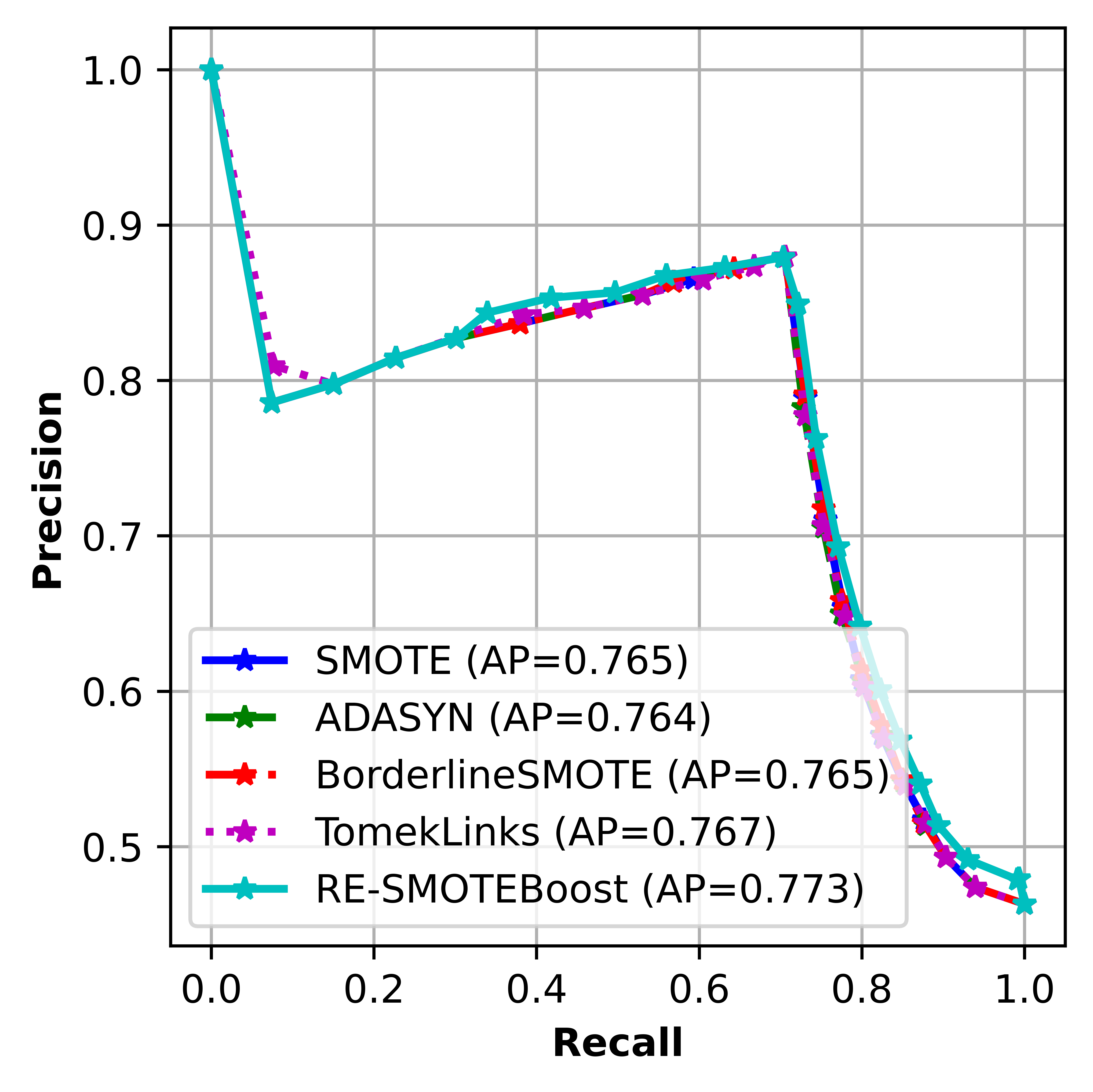} }}
    
    \caption{comparative analysis of precision-recall curves across different data-level methods using 10CV.}
    \label{fig:fig10}
\end{figure*}

\begin{figure*}
    \centering
    
    \subfloat[\centering WOBC dataset]{{\includegraphics[width=5.9cm,height=6cm]{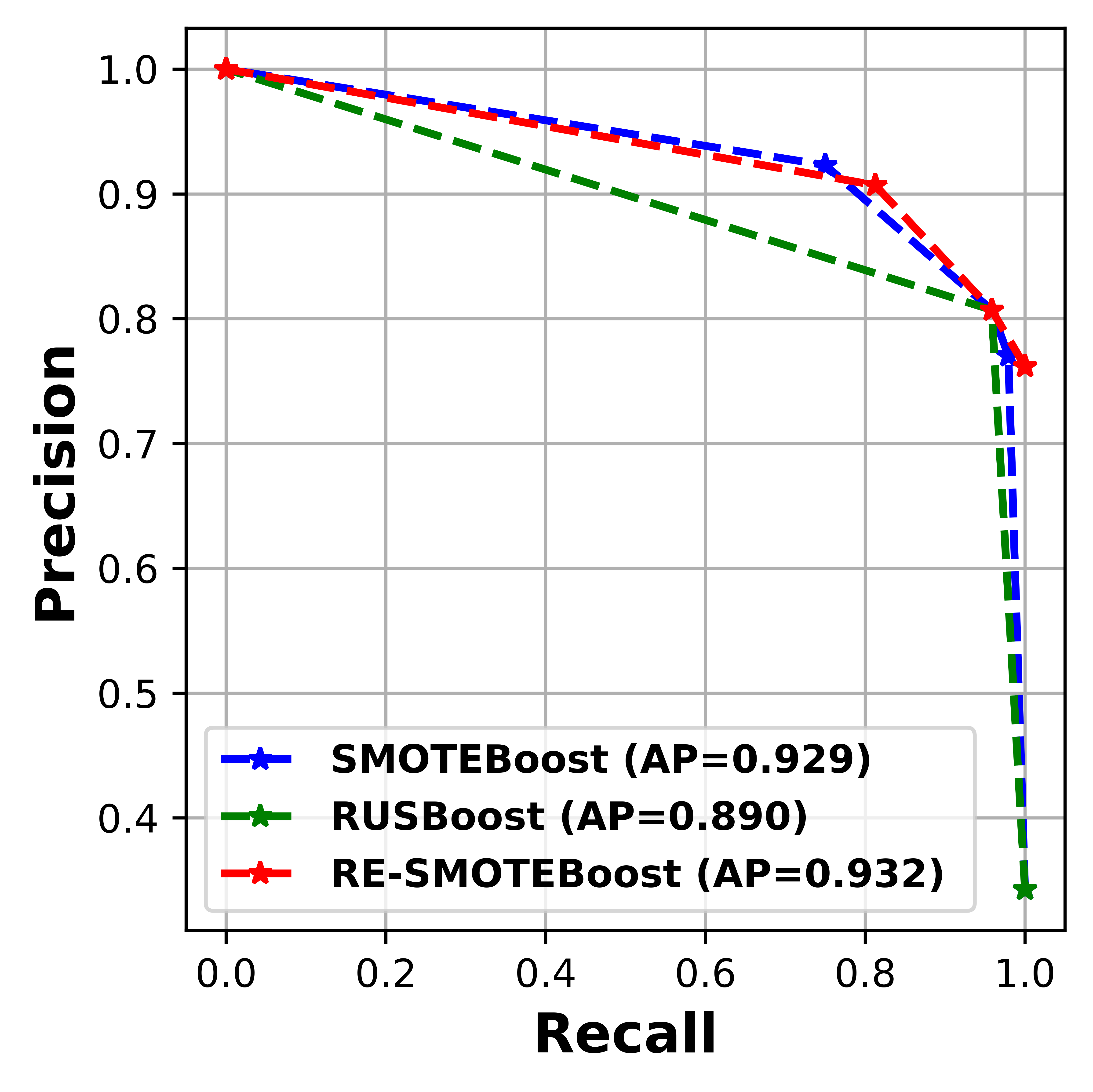} }}
    \subfloat[\centering WDBC dataset]{{\includegraphics[width=5.9cm,height=6cm]{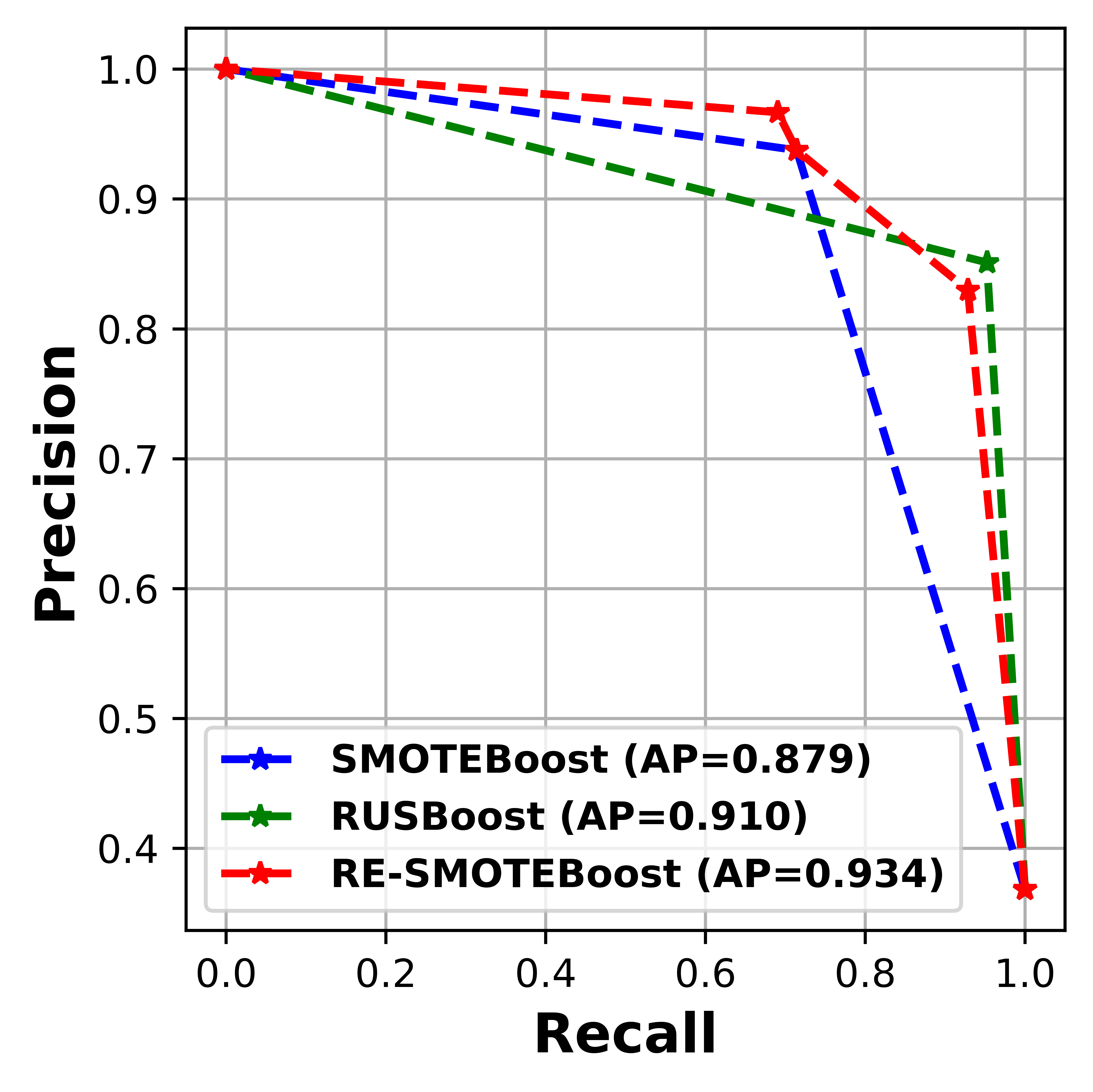} }}
    \subfloat[\centering Mammographic Mass]{{\includegraphics[width=5.9cm,height=6cm]{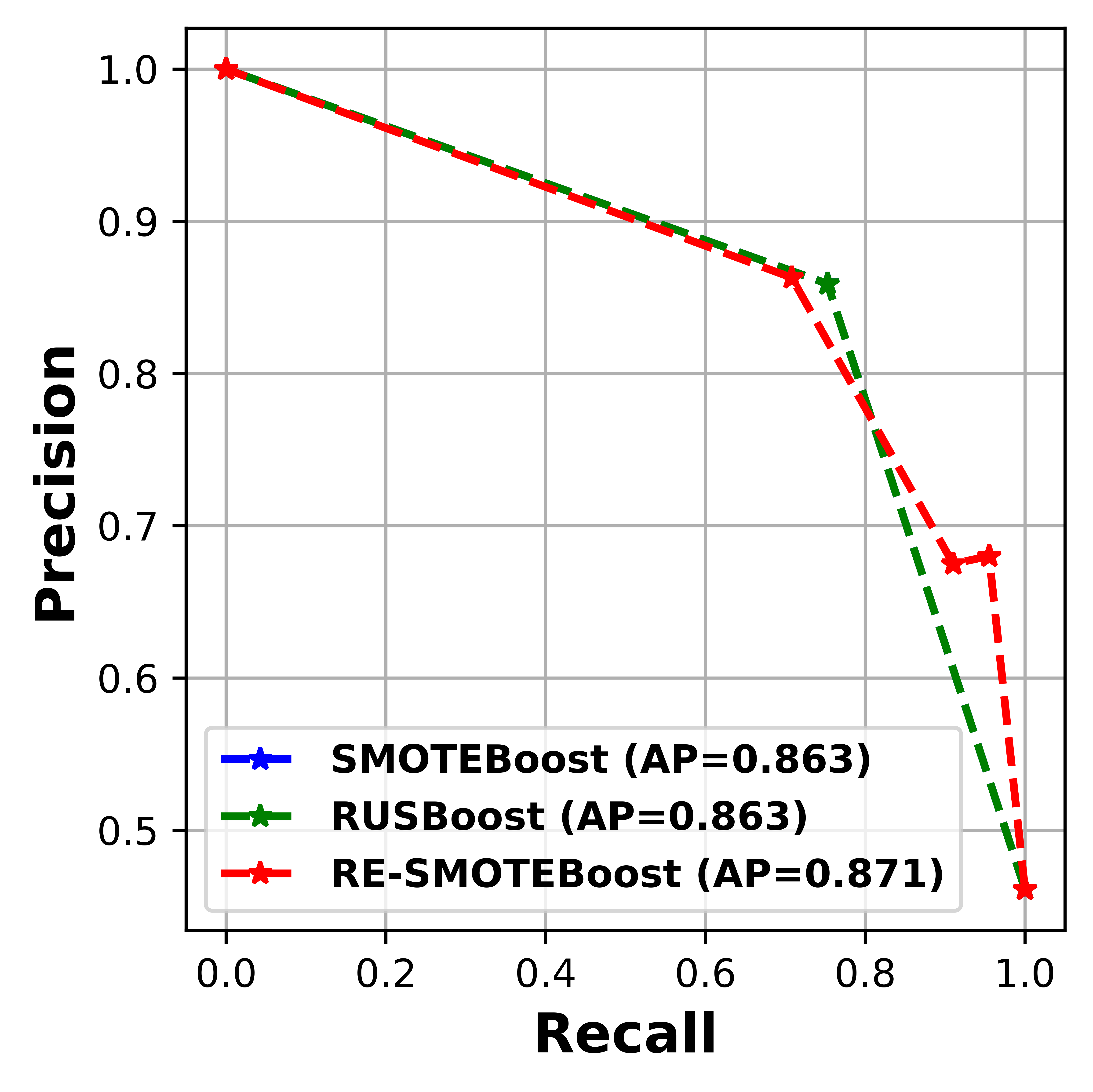} }}
    
    \caption{comparative analysis of precision-recall curves across different boosting methods.}
    \label{fig:fig11}
\end{figure*}

Fig. \ref{fig:fig10} and \ref{fig:fig11} present the analysis of Precision-Recall curves for different sampling methods using the DT classifier. The DT classifier was chosen for its superior performance over the other classifiers used in this experiment. Precision-Recall curves are common metrics for assessing classification performance on imbalanced data. It is considered more suitable than the ROC curve \cite{Experimentation88}. The evaluation was conducted across multiple datasets. Based on the findings, we observe that the proposed method, DT-RE-SMOTEBoost, outperforms other sampling methods and achieves significant improvements in Precision-Recall values. Analyzing Fig. \ref{fig:fig10} and \ref{fig:fig11}, the proposed method reaches its peak performance on the WDBC dataset with a maximum value of 0.964\%. Compared to other methods, the proposed method achieved a greater performance difference. For instance, it significantly enhances Precision-Recall, with improvements ranging from 7.3\% compared to the best-performing model (Tomek Links) to 13.4\% compared to the worst model (ADASYN) on WOBC.

\subsubsection{\textbf{Evaluating the reliability of diagnostic improvements}}

A key challenge in cancer disease diagnosis is ensuring model reliability and stability, which means that the model not only achieves high diagnostic accuracy but also maintains it consistently \cite{Methododology84}. The reliability enhancement of the ensemble technique and proposed method can be analyzed through performance variance using Eq \eqref{eq22}. The variance of each model’s performance is measured and presented in Table \ref{tab12}. Interestingly, DT-RE-SMOTEBoost achieved higher stability in all relevant metrics for both the WOBC and mammographic datasets. DT-RE-SMOTEBoost demonstrated superior stability across all relevant metrics for both the WOBC and mammographic datasets, outperforming other methods, except in the F-score measurement, where DT-SMOTE showed a slight advantage over our proposed method on the WOBC dataset. For instance, DT-RE-SMOTEBoost achieved higher variance reduction with notable differences in precision (88.8\%), recall (88.8\%), and G-means (88.8\%) compared to the best-performing algorithm (DT-RUSBoost) on the mammographic dataset. It is worth noting that DT-RUSBoost achieved the same variance value as DT-RE-SMOTEBoost on the WDBC dataset in some metrics. In particular, the proposed model can achieve a recall variance reduction of approximately (75\%), F-score (75\%), and G-means (75\%) using SVM  on the WOBC dataset. For the WDBC dataset, all metrics obtained similar variance values for all relevant measurements. However, for the mammographic dataset, the proposed method ranked second place after the RUSBoost model, except for precision variance, where the proposed method ranked first with significant improvement. For Coimbra achieved smaller variance value compared to other methods.

\begin{table*}[h]
\renewcommand{\arraystretch}{0.6} 
\setlength{\tabcolsep}{5pt} 

\caption{Diagnostic testing reliability (variance) of the proposed method vs. state-of-the-art methods. .}

\label{tab12}
\setlength{\tabcolsep}{5pt}
\centering
\resizebox{\textwidth}{!}{ 
\begin{tabular}{l c c c c c c c c c}

\hline
\multirow{2}{*}{Cancer dataset} & \multirow{2}{*}{Methods}  & \multicolumn{4}{c}{DT classifier} & \multicolumn{4}{c}{SVM classifier} \\ 
\cmidrule(lr){3-6} \cmidrule(lr){7-10}  
&  & $\sigma $(Precsion) & $\sigma $(Recall) & $\sigma (f_1$-score) & $\sigma (g-means)$ & $\sigma $(Precsion) & $\sigma $(Recall) &  $\sigma (f_1$-score) & $\sigma $(g-means) \\ 

\midrule
        \multirow{4}{*}{WOBC} & SMOTE & 1.10934E-31 & 1.23260E-32 &\textbf{ 4.93038E-32} & 1.23260E-32 & 9.75372E-05 & 0.00026 & 0.00019 & 0.00026 \\
        & SMOTEBoost & 3.53553E-06 & 4.11678E-06 & 3.04096E-07 & 4.11678E-06 & 7.31254E-05 & 0.00016 & 0.00013 & 0.00016 \\
        & RUSBoost & 1.10934E-31 & 1.23260E-32 &\textbf{ 4.93038E-32 }& 1.23260E-32 & 1.10934E-31 & 4.93038E-32 & 4.93038E-32 & 4.93038E-32 \\
        & RE-SMOTEBoost &\textbf{1.23260E-32} &\textbf{ 1.23260E-32} &\textbf{ 1.97215E-31} & \textbf{1.23260E-32} &\textbf{ 0.00000} & \textbf{1.23260E-32} & \textbf{1.23260E-32 }& \textbf{1.23260E-32 }\\\\
    
        \multirow{4}{*}{WDBC} & SMOTE & 8.43047E-05 & 0.00067 & 0.00035 & 0.00067 & 0.00000 & 3.08149E-33 & 0.00000 & 3.08149E-33 \\
        & SMOTEBoost & 9.39828E-05 & 0.00048 & 0.00030 & 0.00048 & 0.01014 & 1.95988E-05 & 0.00074 & 1.95988E-05 \\
        & RUSBoost & \textbf{4.93038E-32 }& 1.23260E-32 & 1.23260E-32 & 1.23260E-32 & 0.00000 & 3.08149E-33 & 0.00000 & 3.08149E-33 \\
        & RE-SMOTEBoost &\textbf{ 1.97215E-31 }& \textbf{1.23260E-32} & \textbf{1.23260E-32} & \textbf{1.23260E-32} & \textbf{0.00000 }&\textbf{ 3.08149E-33 }& \textbf{0.00000} & \textbf{3.08149E-33 }\\\\
    
        \multirow{4}{*}{Mammographic} & SMOTE & 8.91504E-05 & 9.46647E-05 & 8.35583E-05 & 9.46647E-05 & 0.01530 & 0.01542 & 0.01521 & 0.01542 \\
        & SMOTEBoost & 0.00011 & 0.00012 & 0.0001 & 0.00012 & 0.02581 & 0.00600 & 0.01736 & 0.00600 \\
        & RUSBoost & 1.10934E-31 & 1.10934E-31 & 4.93038E-32 & 1.10934E-31 &\textbf{ 4.93038E-32} & \textbf{0.00000 }& 1.23260E-32 &\textbf{ 0.00000} \\
        & RE-SMOTEBoost & \textbf{1.23260E-32 }&\textbf{ 1.23260E-32 }&\textbf{ 4.93038E-32} & \textbf{1.23260E-32} &\textbf{ 1.23260E-32} & \textbf{4.93038E-32} & \textbf{4.93038E-32} & \textbf{1.23260E-32} \\\\
      
        \multirow{4}{*}{Breast Cancer Coimbra} & SMOTE & 0.00153 & 0.00159 & 0.00155 & 0.00159 & 7.70372E-34 & 0.00000 & 1.23260E-32 & 0.00000 \\
        & SMOTEBoost & 0.00127 & 0.00107 & 0.00098 & 0.00107 & 1.71875E-05 & 0.00000 & 1.360123E-05 & 0.00000 \\
        & RUSBoost & 1.23260E-32 & 1.23260E-32 & 0.00000 & 1.23260E-32 & 7.70372E-34 & 0.00000 & 1.23260E-32 & 0.00000 \\
        & RE-SMOTEBoost &\textbf{ 0.00000} &\textbf{ 0.00000 }&\textbf{ 4.93038E-32} &\textbf{ 0.00000 }& \textbf{0.00000} & \textbf{0.00000} & \textbf{2.77334E-32} &\textbf{0.00000} \\\\

        \bottomrule
\end{tabular}

}
\end{table*}

\subsubsection{\textbf{Analysis of class overlapping}}

\begin{figure*}
    \centering
\includegraphics[width=17.7cm,height=12cm]{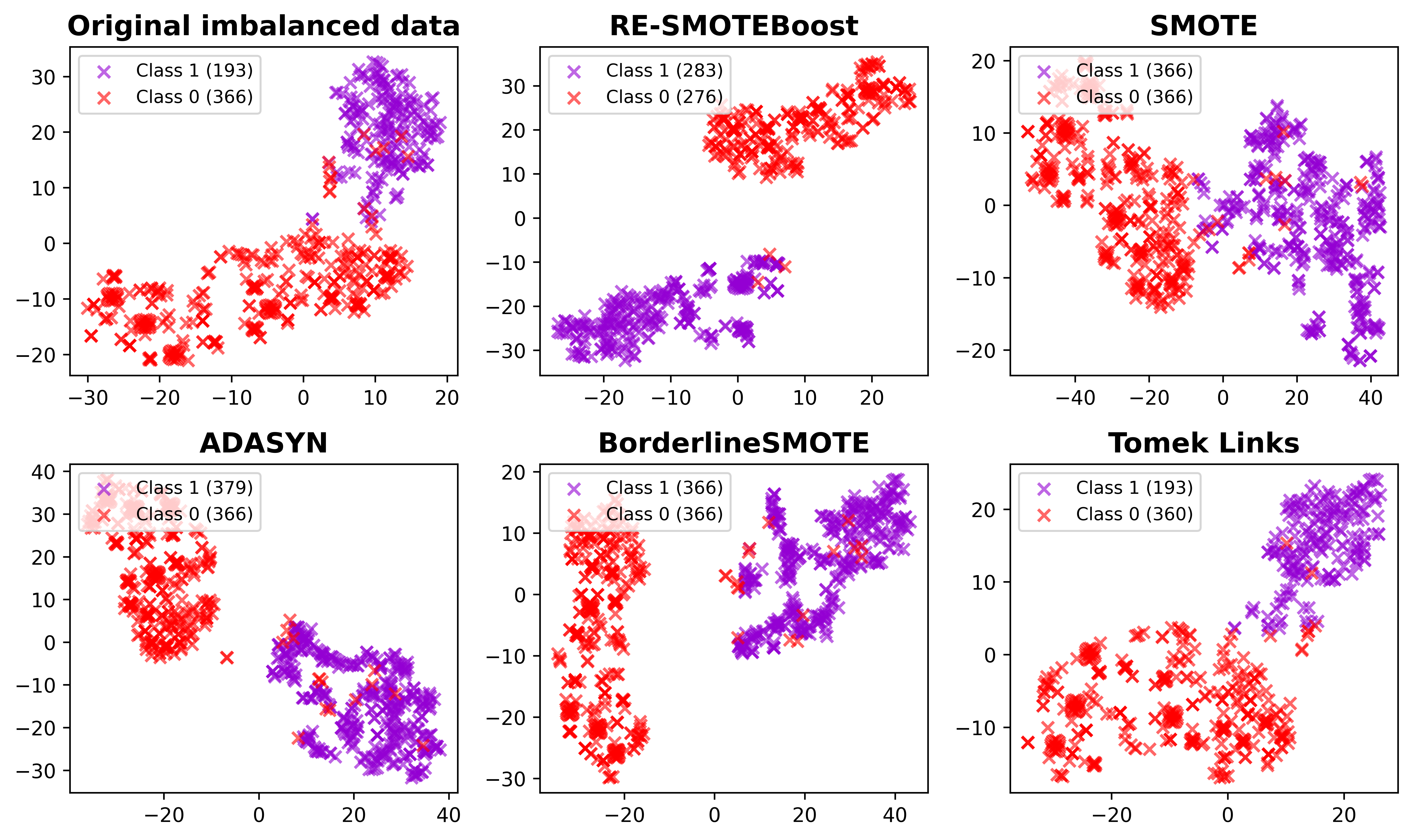}

    \caption{Visualization of the generated data distribution using the T-SNE tool on WOBC dataset.}
    \label{fig:fig12}
\end{figure*}

In this section, we analyze the experimental results to provide deeper insight into the effectiveness of the proposed method, RE-SMOTEBoost, in reducing class overlapping. To justify the effectiveness of RE-SMOTEBoost, we apply the t-SNE tool for data visualization, as shown in Fig \ref{fig:fig12} .Furthermore, we apply Fisher’s discriminant ratio to compare the degree of overlap between each feature in the dataset produced by the proposed method and other sampling techniques using formula  \eqref{eq23} \cite{Experimentation96}. Table \ref{table13}  shows the results of this comparison.

\begin{equation}
 f=\frac{(N_1-N_2)^2}{v_1^2+ v_2^2}. \label{eq23}
\end{equation}

where $N_1, N_2$ indicate the means and $v_1, v_2$ indicate the variances of the samples from the two different classes.

Table \ref{table13} presents the number of features that have smaller Fisher's Discriminant Ratio values after applying different sampling methods. The results indicate that the proposed method, RE-SMOTEBoost, produces a higher number of features with smaller Fisher's Discriminant Ratio values compared to the other methods. For example, RE-SMOTEBoost yields 18 features with smaller Fisher's Discriminant Ratio values, while RUSBoost produces only 13. This result indicates that the proposed method reduces class overlap due to the proposed double regularization penalty.

From Fig. \ref{fig:fig12}, it is evident that the proposed RE-SMOTEBoost method produces a more structured distribution compared to other methods. The majority and minority classes in the proposed method are clearly defined, with no small disjoint anomalies \cite{Experimentation86}, where data points from one class are not misclassified as belonging to the other. Furthermore, there is no class overlap, as the classes are separated. This is due to the double regularization penalty in the proposed method, which pushes synthetic samples away from the majority class and closer to the minority class. Additionally, the proposed noise filter helps the algorithm avoid using noise and outliers in overlapping areas to generate synthetic data, which helps to prevent small disjuncts. Unlike other methods, SMOTE creates synthetic samples in overlapping regions, which can complicate decision boundaries and potentially reduce prediction performance. ADASYN, however, focuses on generating samples near the decision boundary, particularly targeting hard-to-classify instances. Nevertheless, it can still produce misclassified samples and lead to small disjuncts, as some generated points may not fit well within the main body of the minority class. As shown in Fig \ref{fig:fig12}, red points (majority classes) appear within the area of violet points (minority classes). The same issue applies to BorderlineSMOTE: although it generates samples near the decision boundary, it can also result in small disjuncts. In some cases, it may create samples in sparse regions of the minority class or rely on noise and outliers in overlapping areas to generate samples through linear interpolation. Tomek Links, being an under-sampling method, removes samples near the minority class to balance the dataset. However, it may remove important data from the majority class, depending on the distribution shown in \ref{fig:fig12}. It removes some samples from the majority class, but the decision boundary remains unclear, reducing the effectiveness of the predictive task. Additionally, the proposed method differs in how it balances data compared to existing methods. The proposed method balances the data by reducing both majority and minority class samples. For example, it reduces 90 samples (25\%) from the majority class (from 366 to 276) and generates 90 new samples (25\%) for the minority class (from 193 to 283). In contrast, SMOTE, ADASYN, and BorderlineSMOTE focus solely on generating new samples for the minority class, with synthetic sample generation rates of 47.27\%, 49.08\%, and 47.27\%, respectively. For Tomek Links, it removes only 1.64\% of samples from the majority class.

\begin{table*}[h]

\caption{Comparison of the overlapping degree on WDBC dataset.}

\label{table13}
\setlength{\tabcolsep}{5pt}
\centering

\begin{tabular}{p{150pt}|p{100pt}}
\hline
\textbf{Comparison of pairs of methods} & \textbf{Number of features ( f)} \\
\hline

RUSBoost vs. RE-SMOTEBoost  & 13\\

RE-SMOTEBoost vs. RUSBoost&18 \\

SMOTEBoost vs. RE-SMOTEBoost&13 \\

RE-SMOTEBoost vs. SMOTEBoost&18 \\

BorderlineSMOTE vs. RE-SMOTEBoost&8 \\

RE-SMOTEBoost vs. BorderlineSMOTE&23\\

Adasyn vs. RE-SMOTEBoost&11 \\

RE-SMOTEBoost vs. Adasyn&20 \\

SMOTE vs. RE-SMOTEBoost&8 \\

RE-SMOTEBoost vs. SMOTE&23 \\
TomekL\_ink vs.  RE-SMOTEBoost& 8 \\

RE-SMOTEBoost vs. Tomek\_ink&23 \\
\hline
\end{tabular}
\end{table*}

\subsubsection{\textbf{Analysis of Roulette Wheel Selection Against Popular Selection Methods}}
\begin{figure*}
    \centering
    
    \subfloat[\centering WOBC dataset]{{\includegraphics[width=5.9cm,height=7cm]{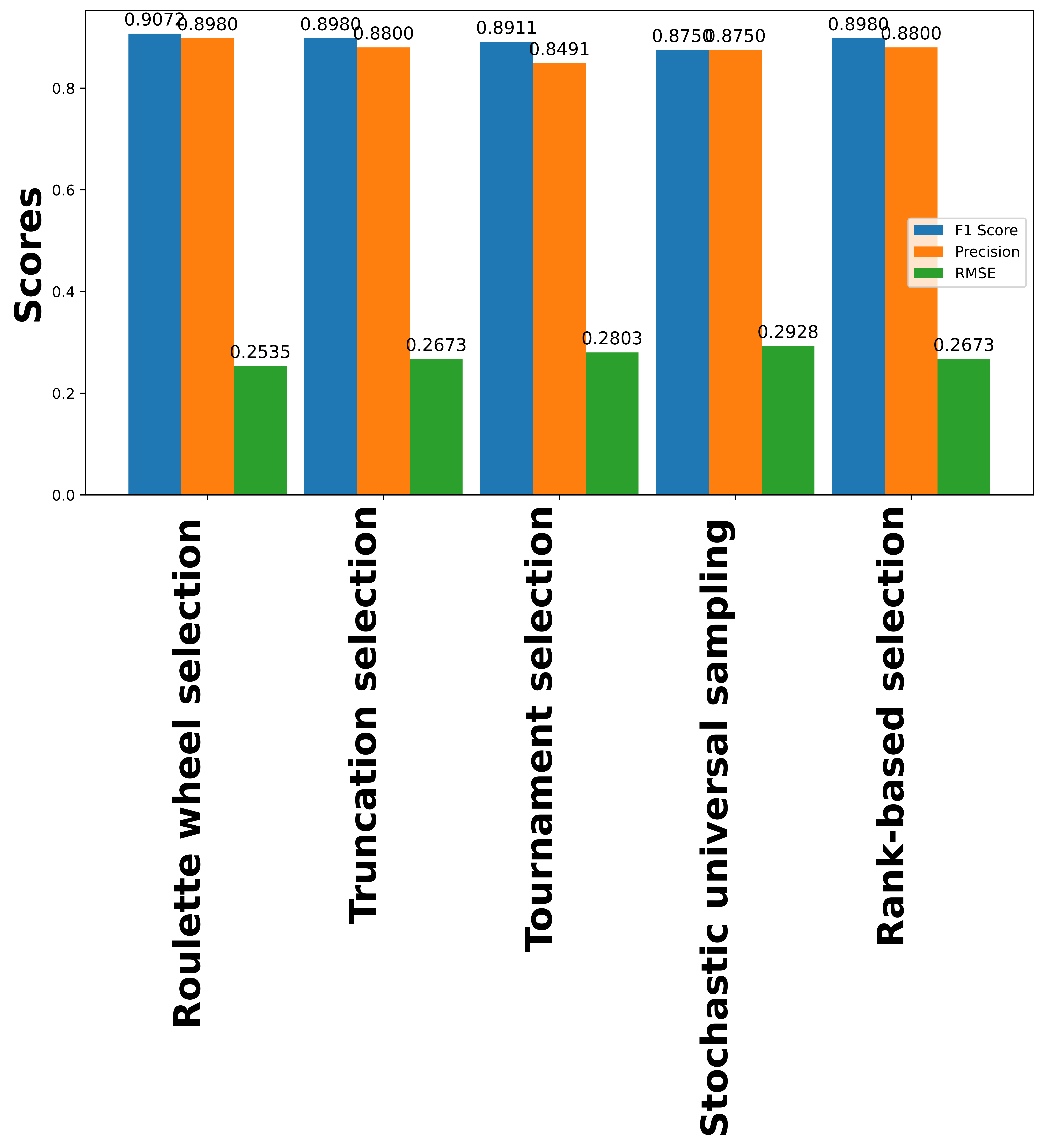} }}
    \subfloat[\centering WDBC dataset]{{\includegraphics[width=5.9cm,height=7cm]{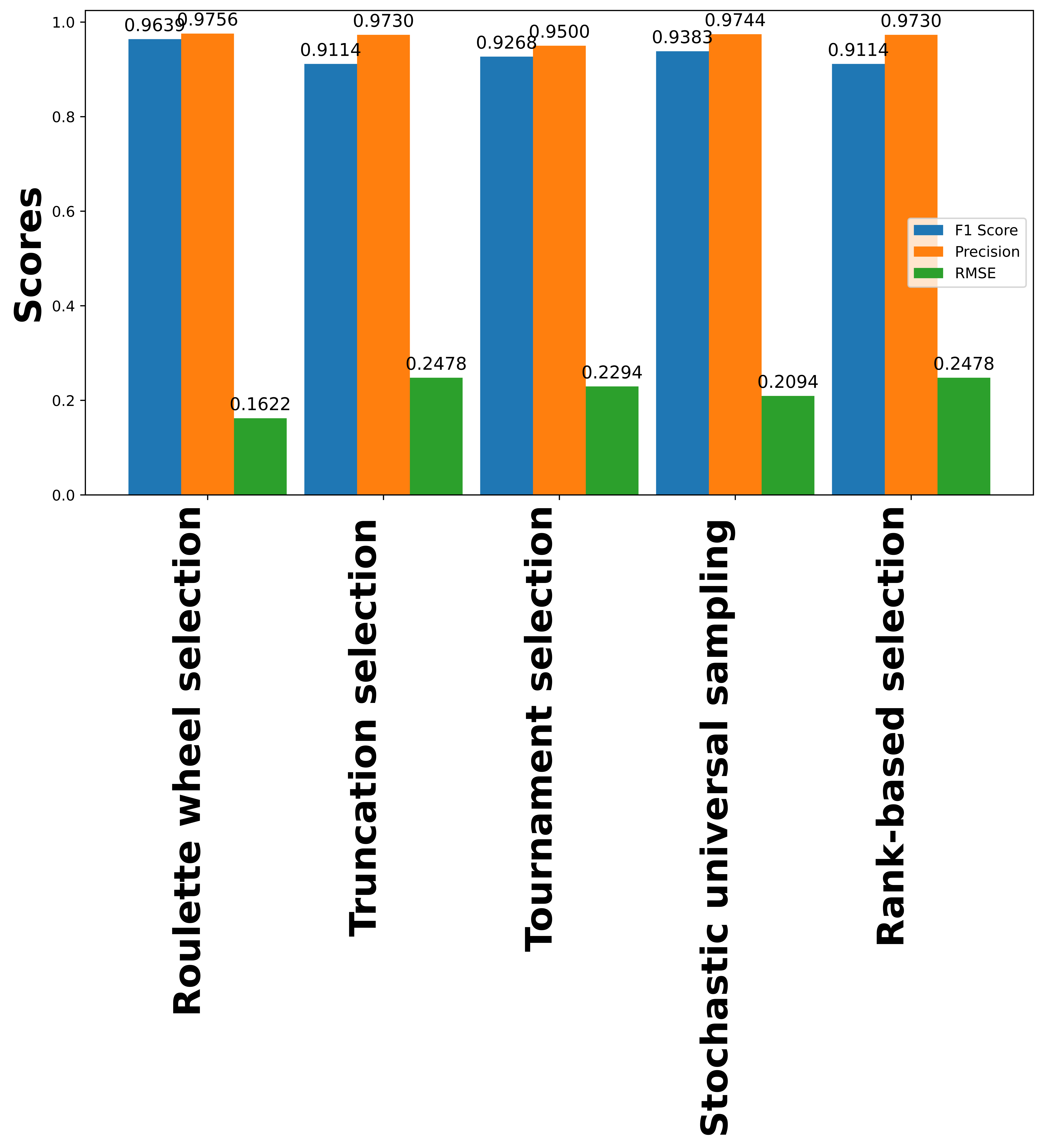} }}
    \subfloat[\centering Mammographic Mass]{{\includegraphics[width=5.9cm,height=7cm]{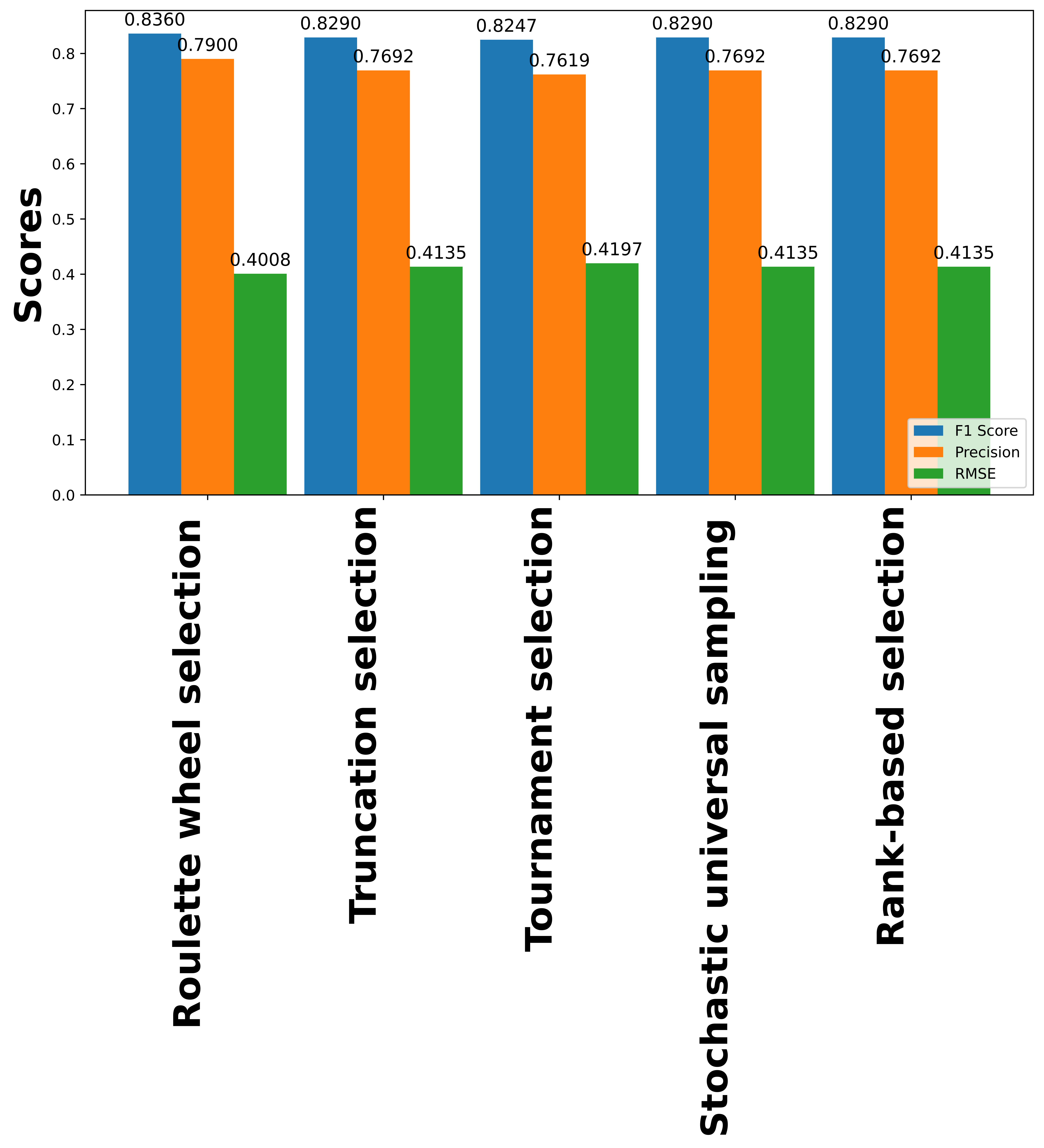} }}
    
    \caption{Performance Comparison of Roulette Wheel Selection vs. Other selection Strategies.}
    \label{fig:fig13}
\end{figure*}

Fig. \ref{fig:fig13}. Comparative results of the proposed method with different selection strategies, including Roulette Wheel Selection (RWS), Tournament, Stochastic universal sampling,  Rank-based, and truncation selection, using various performance metrics (e.g., precision, F1 score, RMSE). The results demonstrate that the proposed RE-SMOTEBoost method with Roulette Wheel Selection outperforms other selection strategies across all three metrics for all datasets. It achieved the lowest Root Mean Square Error (RMSE) of 0.1622, the highest precision of 0.9756, and the highest F1-score of 0.9639 on the WDBC dataset. This practical result supports the integration of RWS in our proposed method for addressing imbalanced data, noise, and overlapping challenges.

\subsection{Experimental discussion }
In cancer diagnosis research, many studies have aimed to improve the algorithms and structures of classification models to enhance prediction accuracy. However, most traditional classification algorithms assume balanced data or minimal misclassification costs, failing to address the challenges posed by data imbalance. The imbalanced data  is a common issue in cancer diagnosis datasets, often due to limitations in sample size, difficulties in data collection, and privacy concerns \cite{Experimentation89}. In this study, we discuss the issue of imbalanced data in cancer diagnosis and its potential impact on patient outcomes, particularly when using machine learning techniques for early diagnosis. Based on the literature, the ensemble-based approach has demonstrated improved performance in cancer diagnosis \cite{Experimentation90}. State-of-the-art ensemble techniques, including boosting methods such as the SMOTEBoost algorithm, use the SMOTE strategy for oversampling the minority class. However, SMOTE generates new synthetic samples by replicating existing samples, which may lead to overfitting, and skewing predictions. Additionally, it suffers from overlapping classes \cite{introduction24}. Furthermore, some noisy samples may contribute to the generation of synthetic samples. As a result, the newly synthesized samples may lack logical consistency, potentially leading to a decline in the classifier's performance \cite{Experimentation98}.

Considering both the superiority of this method for addressing class imbalance and its drawbacks, we have proposed a new method called RE-SMOTEBoost and validated its effectiveness through experiments in the breast cancer diagnosis scenario. To do this, we applied a roulette wheel selection technique to identify minority class samples in the overlapping region, prioritizing those with higher probabilities of generating new samples. This approach encourages the classifier to focus on overlapping regions where the minority class data is extremely sparse, improving its ability to capture the decision boundary. Secondly, we integrated a filter based on information entropy to reduce synthetic noise and retain high-quality samples. To control the position of synthetic data and avoid generating overlapping data, we introduced a double regularization penalty to control the synthetic samples' proximity to the decision boundary, which creates a clear separation between classes. The experimental results showed that the proposed method improved both effectiveness and reliability. Practically, the proposed method achieved the highest accuracy value (95\%) among other sampling methods using the SVM classifier with the WOBC dataset. For instance, it increases the accuracy by 3.58\%-3.22\%, i.e., $\frac{Accuracy_{RE-SMOTEBoost}-Accuracy_{None}}{Accuracy_{RE-SMOTEBoost}}$, compared to the worst model (e.g., the None model without using sampling methods) and the best-performing model (ADaSyn), using the DT classifier on WOBC, respectively, the proposed method shows promising results. Overall, this result indicates that the proposed method performs effectively in imbalanced data scenarios, making it a strong candidate for adoption in analyzing imbalanced medical data. However, accuracy may not fully capture the model's performance [87]. For this reason, metrics such as precision, recall, F-score, and G-means are employed for further validation of the effectiveness of the proposed method. The findings showed that the proposed method outperformed other methods for boosting and data-level approaches, except in a few scenarios where the proposed method achieved the same result as other methods, such as SVM-SMOTEBoost on the WDBC dataset in the boosting methods comparison (Table 6). This could be explained by the structure of the SVM classifier, which affects the overall result more than the boosting approach itself. This explanation is in line with \cite{Methododology84}, which has consistently found that the structure of a single model (SVM) controls the final performance, and the boosting approach focuses on the performance of the base classifier (SVM in this study) to predict the final performance \cite{Experimentation91}. For this reason, researchers have recommended using weak classifiers with boosting methods, such as the DT classifier, as the first option due to the increasing or decreasing performance influenced by the definition of the base classifier \cite{Experimentation92}. However, compared to other methods across all datasets,  the proposed method still achieved superior performance for all these relevant metrics. Notably, the proposed method achieved a higher recall value (94.70\%) compared to other methods using the SVM On the WOBC dataset. There are very few exceptions where the precision of other methods (e.g., DT-SMOTEBoost on the WDBC dataset) achieved slightly superior performance compared to the proposed method. However, looking at the corresponding recall and F-score values for the exceptions, we note that both recall and F-score are lower than the proposed method. This observation confirms the effectiveness of the proposed method from a different perspective. Firstly, increasing the precision value does not serve the main objective of this study, where the main goal is to improve minority class recognition. However, high precision means a high false positive error (many samples from the majority class are classified as the minority class), but there is no indication of minority class errors. In contrast, a high recall value indicates that the proposed method correctly predicted minority class samples (minimal false negatives), which supports the main research question of this study: to improve minority class recognition and reduce overlap. Practically, this result means that cancer patients can be more accurately identified and reduce the loss of patient lives \cite{introduction6}. F-score is the trade-off between recall and precision performance, with emphasis on the minority class, and it is considered more important than other metrics. A high F-score value indicates the effectiveness of the proposed method. This assumption is in line with \cite{introduction6}, which confirms that good models achieve precision and recall trade-offs.Fig  \ref{fig:fig6} and \ref{fig:fig7} show the comparative analysis of AUC measurements across different sampling methods with 20CV, using a violin plot as a visualization tool, while Fig\ref{fig:fig10} and \ref{fig:fig11} illustrates a comparative study of precision-recall curve measurements across different sampling methods using 10CV. Based on the findings from both metrics, the proposed method, DT-RE-SMOTEBoost, achieved the highest AUC and precision-recall values compared to other sampling methods. Since the AUC (ROC curve) is the main robustness metric for model evaluation with imbalanced data \cite{Experimentation93}, the recall-precision curve is more likely to provide insights about the positive class for imbalanced predictions than the ROC curve \cite{Experimentation94}. The achievement of a high AUC value indicates the strong ability of the proposed method to distinguish between the two classes: the majority (negative) and minority (positive) classes. This finding supports the benefit of integrating the proposed double regularization penalty to create clear separation and improve the model's ability to discriminate decision boundaries between classes in overlapping regions. On the other hand, the high precision-recall curve values of the proposed method indicate that it is an effective model for predicting the minority class \cite{Experimentation94}. This supports our research objectives and the assumption that the proposed method enhances the classifier's focus on the minority class. Practically, in real-world situations, these results indicate that the proposed method can achieve great performance in detecting cancer patients (in rare cases), which minimizes mortality and disease incidence rates. Analyzing the reliability results in Table  \ref{tab12} shows that the proposed method achieved the lowest variance values over others in the most relevant metrics. Only in a few cases is it ranked second place. Even in these cases, it ranked second in certain metrics, not all metrics. For instance, it achieved a variable reduction by 100\%-88.8\%, i.e., $\frac{\sigma (precision_{RUSBoost})-\sigma (precision_{RE-SMOTEBoost})}{precision_{RUSBoost}}$, as compared to the worst model (e.g., SMOTE) and the best-performing model (RUSBoost), using a DT classifier on WOBC, respectively, Fig. \ref{fig:fig6} and \ref{fig:fig7} shows that the violin plot of the proposed method is smaller compared to other methods, and the median line is located at the center of the box. The small variance value from Table \ref{tab12} and the results from Fig \ref{fig:fig6} and \ref{fig:fig7} indicate the reliability of the proposed model. Reliability means that the results remain stable across multiple execution runs, ensuring higher consistency and greater confidence in the predicted values. The median line at the center of the AUC box means the structural distribution of training data is maintained, which makes the performance of the model symmetrically stable and supports the idea of using information entropy to select high-quality samples from both classes in our study. This finding reflects a practical benefit, where the model has high stability and better generalization performance (more confidence in testing real cases).

In summary, these findings have significant implications for practical applications in medical settings. They can assist clinicians in the early diagnosis of cancer, particularly in rare cancer cases, by improving the performance of predictions and diagnoses while ensuring the robustness and reliability of decision-making. Additionally, it can contribute to addressing the class imbalance challenges in large dataset scenarios by reducing the size of both classes, resulting in a smaller yet structurally balanced distribution that machine learning models can process effectively. Theoretically, this research adds new ideas to existing literature and opens the door for further improvement of models. However, this study is limited to the scope of a binary classification scenario. Additionally, it deals with numerical data and text rather than other types, such as images. This limited scope opens opportunities for extending the study in future research, such as applying the One-vs-All (OvA) approach to the proposed method for multi-class situations. Additionally, the application can be extended to other types of data.
\section{Conclusion}
In clinical practice and disease analysis, implementing machine learning models as a powerful tool for early diagnosis has become an inevitable reality. Traditional models are built on the assumption that the data is balanced. However, in real-world applications such as the medical field, most of the available data is imbalanced due to limited samples, data collection difficulties, and privacy concerns. A boosting-based approach, including the SMOTEBoost method, is widely used to address the issue of imbalanced data by generating new synthetic samples for minority classes. However, it generates synthetic samples by linearly interpolating between a minority class sample and its neighbors, which produces noise and outlier data. Moreover, it generates samples for the minority class while neglecting the influence of the majority class. Furthermore, it overlooks the overlapping region, where the decision boundary is unclear, which is crucial for accurate prediction. As a result, synthetic samples may be generated within this area, which increases the complexity of the prediction task, or placed too far from it, thereby neglecting minority samples situated in the overlapping region. This research contributes to addressing the aforementioned weaknesses and introduces a new enhancement of the previous method, named RE-SMOTEBoost. The proposed method aims to achieve data balance through a double pruning process applied to both the majority and minority classes. For the majority class, information entropy is utilized to eliminate low-quality samples while retaining high-quality ones. In the case of the minority class, a roulette wheel selection mechanism is employed to assess the samples, prioritizing those in the overlapping and inter-class regions with a high probability of generating new synthetic samples. This process maintains the structural distribution of data after the balancing procedure by returning only high-quality samples that contain important data. Additionally, a noise filter based on information entropy has been introduced to filter the synthetic samples with low quality. In addition, a double regularization penalty is employed to control the synthetic samples' proximity to the decision boundary and avoid class. Based on the experimental findings, the proposed method achieved a 3.22\% improvement in accuracy and an 88.8\% reduction in variance compared to the best-performing sampling methods.

Although the proposed method showed promising results in addressing imbalanced medical data, its application is limited to binary classification scenarios. Additionally, in our application, we focused on numerical and text data rather than other types, such as images, which require the use of new augmentation techniques like GAN-based approaches. 
As a potential direction for future work, we can extend the proposed method to multi-class problems by using the One-vs-All (OvA) approach, which transforms multi-class classification into multiple binary classification tasks, allowing us to apply the proposed method. For future applications, we will extend our work on class imbalance and data augmentation to encompass other data types, such as medical images, where advanced deep-learning techniques can be leveraged.

\bibliographystyle{apalike}
\bibliography{references}


\end{document}